\documentclass[ROB,biber,plain]{nowfnt}

\usepackage[utf8]{inputenc}
\usepackage[T1]{fontenc}

\usepackage{mwe} 

\usepackage[english]{babel} 

\usepackage{titlesec} 
\usepackage{amsmath} 
\usepackage{amsfonts} 
\usepackage{subfig} 
\usepackage{graphicx} 
\usepackage[toc]{glossaries} 
\usepackage{tabulary} 
\usepackage{multirow} 
\usepackage{multicol} 
\usepackage{array} 
\usepackage[absolute,overlay]{textpos}

\usepackage[colorinlistoftodos,prependcaption,textsize=tiny]{todonotes} 

\usepackage{booktabs} 
\makeglossaries

\newcommand*{\clearurl}{%
	\iffieldundef{doi}%
	{}%
	{\iffieldundef{url}%
		{}%
		{\clearfield{url}}%
		\iffieldundef{eprint}%
		{}%
		{\clearfield{eprint}}%
	}%
	\iffieldundef{url}%
	{}%
	{\iffieldundef{eprint}%
		{}%
		{\clearfield{eprint}}%
	}%
}
\AtEveryBibitem{\clearurl}
\AtEveryCitekey{\clearurl}

\titlespacing*{\section}{0pt}{\baselineskip}{\baselineskip}
\titlespacing*{\subsection}{0pt}{\baselineskip}{\baselineskip}
\titlespacing*{\subsubsection}{0pt}{\baselineskip}{\baselineskip}

\newcommand{\red}[1]{{\color{Red}{#1}}\xspace}

\DeclareMathOperator*{\argmax}{arg\,max}
\DeclareMathOperator*{\argmin}{arg\,min}
\graphicspath{ {./images/} }

\newcommand{\REAL}{\mathbb{R}}

\newcommand{\STATES}{\mathcal{S}}
\newcommand{\ACTIONS}{\mathcal{A}}
\newcommand{\TRANSITION}{\mathcal{T}}
\newcommand{\REWARDS}{\mathcal{R}}
\newcommand{\OBSERVATIONS}{O}
\newcommand{\STATE}{s}

\newcommand{\STATENEW}{s'}
\newcommand{\ACTION}{a}
\newcommand{\OBSERVATION}{o}
\newcommand{\POLICY}{\pi}

\newcommand{\AFFORD}{\mathcal{AF}}
\newcommand{\INTENT}{\mathcal{I}}

\newcommand{\decay}{\gamma}

\newcommand{\EXPECT}{\mathbb{E}}
\newcommand{\prob}{\mathbb{P}}

\AtBeginEnvironment{tabulary}{\fontsize{10}{12}\selectfont}
\AtBeginEnvironment{tabular}{\fontsize{10}{12}\selectfont}

\newcommand{\rot}[2]{\multirow{#1}{*}{\rotatebox[origin=c]{90}{#2}}}
\newcommand{\PreserveBackslash}[1]{\let\temp=\\#1\let\\=\temp}
\newcolumntype{Z}[1]{>{\PreserveBackslash\centering}p{#1}}

\newacronym{afd}{AfD}{Abstraction from Demonstration}
\newacronym{actamer}{ACTAMER}{Actor-Critic TAMER}
\newacronym{aeus}{AEUS}{Expected Utility Selection}
\newacronym{aggrevate}{AggreVaTe}{Aggregate Values To Imitate}
\newacronym{ai}{AI}{Artificial Intelligence}
\newacronym{a-opi}{A-OPI}{Advice-Operator Policy Improvement}
\newacronym{al}{AL}{Apprenticeship Learning}
\newacronym{asr}{ASR}{Automatic Speech Recognition}

\newacronym{bc}{BC}{Behavioral Cloning}
\newacronym{bf}{BF}{Basis Function}
\newacronym{bco}{BCO}{Behavioral Cloning from Observations}

\newacronym{ceiling}{CEILing}{Corrective and Evaluative Interactive Learning}
\newacronym{coach}{COACHc}{COrrective Advice Communicated by Humans}
\newacronym{coache}{COACHe}{Convergent Actor-Critic by Humans}
\newacronym{cnn}{CNN}{Convolutional Neural Network}
\newacronym{cycle-of-learning}{CoL}{Cycle-of-Learning}

\newacronym{dagger}{DAgger}{Data Aggregation}
\newacronym{da-rb}{DA-RB}{DAgger with Replay Buffer}
\newacronym{ddpgfd}{DDPGfD}{Deep Deterministic Policy Gradient from Demonstration}
\newacronym{dempref}{DemPref}{Learning Reward Functions by Integrating Human Demonstrations and Preferences}
\newacronym{dmp}{DMP}{Dynamic Movement Primitive}
\newacronym{dp}{DP}{Dynamic Programming}
\newacronym{dpl}{DPL}{Direct Policy Learning}
\newacronym{ds}{DS}{Dynamical System}
\newacronym[longplural={Degrees of Freedom},shortplural={DoF}]{dof}{DoF}{Degree of Freedom}
\newacronym{d-coach}{D-COACH}{Deep COACH}
\newacronym{d-tamer}{D-TAMER}{Deep TAMER}
\newacronym{dstl}{DSTL}{Desired State Transition Learning}
\newacronym{dqfd}{DQfD}{Deep Q-learning from demonstrations}

\newacronym{ensemble dagger}{EnsembleDAgger}{Ensemble Dagger}
\newacronym{eil}{EIL}{Expert Intervention Learning}
\newacronym{eeg}{EEG}{ElectroEncephaloGraph}
\newacronym{emg}{EMG}{ElectroMyoGraph}
\newacronym{efd}{EfD}{Exploration from Demonstration}

\newacronym{ferl}{FERL}{Feature Expansive Reward Learning}
\newacronym{fresh}{FRESH}{Feedback-based REward SHaping}

\newacronym{ga}{GA}{Genetic Algorithm}
\newacronym{gmm}{GMM}{Gaussian Mixture Model}
\newacronym[longplural={Gausian Processes}]{gp}{GP}{Gaussian Process}
\newacronym{gpr}{GPR}{Gaussian Process Regression}
\newacronym{gui}{GUI}{Graphical User Interface}

\newacronym{hcai}{HCAI}{Human Centered Artificial Intelligence}
\newacronym{hcml}{HCML}{Human Centered Machine Learning}
\newacronym{hil}{HIL}{Human in the Loop}
\newacronym{hil-il}{HIL-IL}{Human in the Loop Imitation Learning}
\newacronym{hil-ml}{HIL-ML}{Human in the Loop Machine Learning}
\newacronym{hil-ai}{HIL-AI}{Human in the Loop Artificial Intelligence}
\newacronym{hil-rl}{HIL-RL}{Human in the Loop Reinforcement Learning}
\newacronym{hri}{HRI}{Human-Robot Interaction}
\newacronym{hrt}{HRT}{Human-Robot Teaching}
\newacronym{hci}{HCI}{Human-Computer Interaction}
\newacronym{hat}{HAT}{Human-Agent Transfer}
\newacronym{hg-dagger}{HG-DAgger}{Human Gated DAgger}
\newacronym{hsmm}{HSMM}{Hidden Semi-Markov Model}

\newacronym{ilosa}{ILoSA}{Interactive Learning of Stiffness and Attractors}
\newacronym{iai}{IAI}{Interactive Artificial Intelligence}
\newacronym{il}{IL}{Imitation Learning}
\newacronym{iec}{IEC}{Interactive Evolutionary Computation}
\newacronym{ier}{IER}{Interactive Evolutionary Robotics}
\newacronym{iil}{IIL}{Interactive Imitation Learning}
\newacronym{ils}{ILS}{Interactive Learning Systems}
\newacronym{iml}{IML}{Interactive Machine Learning}
\newacronym{iwr}{IWR}{Intervention Weighted Regression}
\newacronym{irl}{IRL}{Inverse Reinforcement Learning}
\newacronym{isabl}{I-SABL}{Inferring Strategy-Aware Bayesian Learning }
\newacronym{interactive rl}{Interactive RL}{Interactive Reinforcement Learning}

\newacronym{knn}{KNN}{K-Nearest Neighbors}
\newacronym{kmp}{KMP}{Kernelized Movement Primitive}
\newacronym{kl}{KL}{Kullback–Leibler}

\newacronym{lags-ds}{LAGS-DS}{Locally Active Globally Stable Dynamic System}
\newacronym{land}{LaND}{Learning to Navigate from Disengagements}
\newacronym{lazy dagger}{LazyDAgger}{Lazy DAgger}
\newacronym{lbw}{LBW}{Learning By Watching}
\newacronym{lira}{LIRA}{Learning Interactively to Resolve Ambiguity}
\newacronym{lfc}{LfC}{Learning from Critique}
\newacronym{lfd}{LfD}{Learning from Demonstration}
\newacronym{lfo}{LfO}{Learning from Observations}
\newacronym{lfp}{LfP}{Learning from Preferences}
\newacronym{loki}{LOKI}{Locally Optimal search after K-step Imitation}
\newacronym{lstm}{LSTM}{Long Short-Term Memory}
\newacronym{lut}{LUT}{Look Up Table}
\newacronym{lwr}{LWR}{Locally Weighed Regression}
\newacronym{lec}{LEC}{Learning with an External Critic}

\newacronym{mbd}{MbD}{Mapping by Demonstration}
\newacronym{mle}{MLE}{Maximum Likelihood Estimation}
\newacronym{map}{MAP}{Maximum a Posteriori}
\newacronym{ml}{ML}{Machine Learning}
\newacronym{mp}{MP}{Movement Primitive}
\newacronym{mse}{MSE}{Mean Squared Error}
\newacronym{mdp}{MDP}{Markov Decision Process}
\newacronym{mpc}{MPC}{Model Predictive Control}
\newacronym{mc}{MC}{Monte Carlo}

\newacronym{nn}{NN}{Neural Network}
\newacronym{nist}{NIST}{National Institute of Standards and Technology}

\newacronym{pbd}{PbD}{Programming by Demonstrations or Programming by Doing}
\newacronym{pca}{PCA}{Principal Component Analysis}
\newacronym{pdf}{PDF}{Probability Density Function}
\newacronym{pd}{PD}{Proportional Derivative}
\newacronym{pid}{PID}{Proportional Integral Derivative}
\newacronym{pro mp}{ProMP}{Probabilistic Movement Primitive}
\newacronym{pomdp}{POMDP}{Partially Observable Markov Decision Process}
\newacronym{power}{PoWER}{Policy Learning by Weighting Exploration with the Returns}
\newacronym{ppl}{PPL}{Preference-Based Policy Learning}
\newacronym{ps}{PS}{Policy Search}

\newacronym{rbf}{RBF}{Radial Basis Function}
\newacronym{rl}{RL}{Reinforcement Learning}
\newacronym{rlwhil}{RL-HiL}{RL with Human-in-the-Loop}
\newacronym{rnn}{RNN}{Recursive Neural Network}

\newacronym{saferl}{Safe-RL}{Safe Reinforcement Learning}
\newacronym{safe dagger}{SafeDAgger}{Safe DAgger}
\newacronym{sabl}{SABL}{Strategy-Aware Bayesian Learning }
\newacronym{seds}{SEDS}{Stable Estimator of Dynamical System}
\newacronym{shiv}{SHIV}{Svm-based reduction in Human InterVention}
\newacronym{shield}{SHIELD}{Super-Human InsErtion using Learning from Demonstration}
\newacronym{svm}{SVM}{Support Vector Machine}
\newacronym{srl}{SRL}{State Representation Learning}

\newacronym{tamer}{TAMER}{Training an Agent Manually via Evaluative Reinforcement}
\newacronym{td}{TD}{Temporal-Difference}
\newacronym{td-dis}{TD-DIS}{Temporal-Difference per Decision Importance Sampling}
\newacronym{thrifty dagger}{ThriftyDAgger}{Thrifty DAgger}
\newacronym{tics}{TICS}{Task-Instruction-Contingency-Shaping}
\newacronym{tips}{TIPS}{Teaching Imitative Policies in State-space}
\newacronym{tp-gmm}{TP-GMM}{Task-Parameterized Gaussian Mixture Model}
\newacronym{tpc}{TPC}{Tactile Policy Correction}
\newacronym{tpp}{TPP}{Trajectory Preference Perceptron}
\newacronym{trl}{TRL}{Technology Readiness Level}

\newacronym{vr}{VR}{Virtual Reality}

\title{Interactive Imitation Learning in Robotics: A Survey}

\booltrue{authortwocolumn} 
\maintitleauthorlist{
	Carlos Celemin$^{*}$ \\
	Delft University of Technology \\
	c.e.celeminpaez@tudelft.nl
	\and
	Rodrigo Pérez-Dattari$^{*}$ \\
	Delft University of Technology \\
	r.j.perezdattari@tudelft.nl
	\and
	Eugenio Chisari$^{*}$ \\
	University of Freiburg \\
	chisari@cs.uni-freiburg.de
	\and
	Giovanni Franzese$^{*}$ \\
	Delft University of Technology \\
	g.franzese@tudelft.nl
	\and
	Leandro de Souza Rosa \\
	Delft University of Technology \\
	l.desouzarosa@tudelft.nl
	\and
	Ravi Prakash \\
	Delft University of Technology \\
	r.prakash-1@tudelft.nl
	\and
	Zlatan Ajanović\\
	Delft University of Technology \\
	z.ajanovic@tudelft.nl
	\and
	Marta Ferraz \\
	Delft University of Technology \\
	m.ferraz@tudelft.nl
	\and
	Abhinav Valada \\
	University of Freiburg \\
	valada@cs.uni-freiburg.de
	\and
	Jens Kober \\
	Delft University of Technology \\
	j.kober@tudelft.nl
}

\issuesetup
{%
	copyrightowner={xx},
	volume        = xx,
	issue         = xx,
	pubyear       = xx,
	isbn          = xxx-x-xxxxx-xxx-x,
	eisbn         = xxx-x-xxxxx-xxx-x,
	doi           = xx.xxxx/XXXXXXXXX,
	firstpage     = 1, 
	lastpage      = xx
}

\author[1]{Celemin$^{*}$, Carlos}
\author[1]{Pérez-Dattari$^{*}$, Rodrigo}
\author[2]{Chisari$^{*}$, Eugenio}
\author[1]{Franzese$^{*}$, Giovanni}
\author[1]{de Souza Rosa, Leandro}
\author[1]{Prakash, Ravi}
\author[1]{Ajanović, Zlatan}
\author[1]{Ferraz, Marta}
\author[2]{Valada, Abhinav}
\author[1]{Kober, Jens}

\affil[1]{Department of Cognitive Robotics, Delft University of Technology}
\affil[2]{Robot Learning Lab, University of Freiburg}

\articledatabox{\nowfntstandardcitation}

\begin{document}
\makeabstracttitle

\begin{abstract}
Interactive Imitation Learning (IIL) is a branch of Imitation Learning (IL) where human feedback is provided intermittently during robot execution allowing an online improvement of the robot's behavior.

In recent years, IIL has increasingly started to carve out its own space as a promising data-driven alternative for solving complex robotic tasks. The advantages of IIL are twofold, 1) it is data-efficient, as the human feedback guides the robot directly towards an improved behavior (in contrast with Reinforcement Learning (RL), where behaviors must be discovered by trial and error), and 2) it is robust, as the distribution mismatch between the teacher and learner trajectories is minimized by providing feedback directly over the learner's trajectories (as opposed to offline IL methods such as Behavioral Cloning).

Nevertheless, despite the opportunities that IIL presents, its terminology, structure, and applicability are not clear nor unified in the literature, slowing down its development and, therefore, the research of innovative formulations and discoveries.

In this article, we attempt to facilitate research in IIL and lower entry barriers for new practitioners by providing a survey of the field that unifies and structures it. In addition, we aim to raise awareness of its potential, what has been accomplished and what are still open research questions.

We organize the most relevant works in IIL in terms of human-robot interaction (i.e., types of feedback), interfaces (i.e., means of providing feedback), learning (i.e., models learned from feedback and function approximators), user experience (i.e., human perception about the learning process), applications, and benchmarks. Furthermore, we analyze similarities and differences between IIL and RL, providing a discussion on how the concepts \emph{offline}, \emph{online}, \emph{off-policy} and \emph{on-policy} learning should be transferred to IIL from the RL literature.

We particularly focus on robotic applications in the real world and discuss their implications, limitations, and promising future areas of research.

\end{abstract}
\chapter{Introduction}\label{sec:introduction}

\section{Motivation}\label{sec:IntroMotivation}
\begin{figure}
    \centering
    \includegraphics[width=0.90\linewidth]{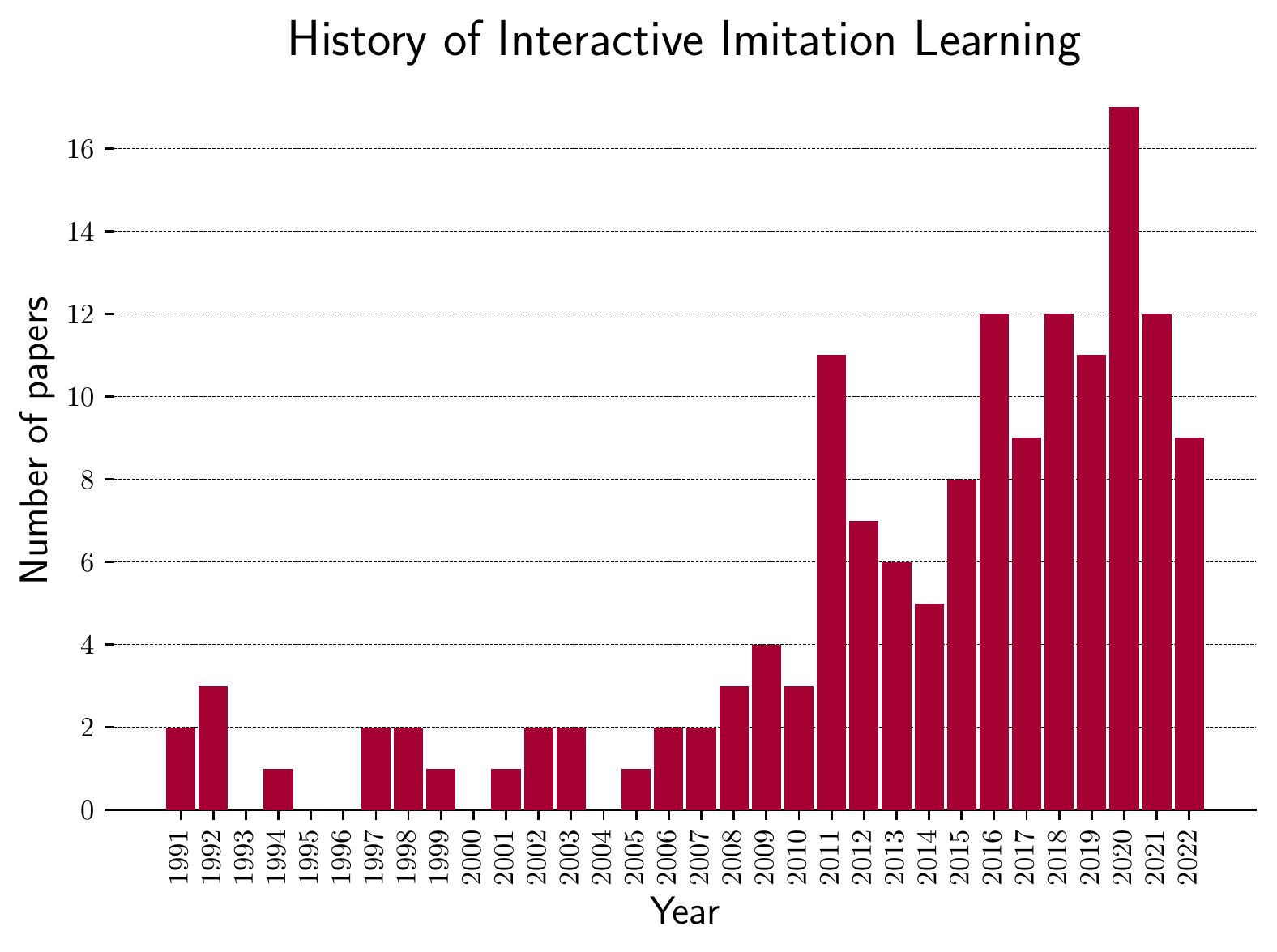}
    \caption{Histogram of \gls{iil} papers (from the group of works surveyed in this paper, till the beginning of 2022) written per year.}
    \label{fig:history_iil}
\end{figure}
Existing robotic technology is still mostly limited to being used by expert programmers who can adapt the systems to new required conditions, but not flexible and adaptable by non-expert workers or end-users. \gls{il} has obtained considerable attention as a potential direction for enabling all kinds of users to easily program the behavior of robots or virtual agents. The teaching process takes place directly in the application context, in a natural way for humans, and does not require engineering effort to adapt the behavior for each different scenario.

In the case teachers (i.e., humans with knowledge about the task) are available and able to transfer their knowledge to the agent, it is preferred to program behaviors from recorded demonstrations rather than tackling the problem with other \gls{ml} techniques such as \gls{rl}, which involve additional design, infrastructure, safety, and data efficiency challenges \citep{sutton2018reinforcement}, and in many cases are not applicable to physical systems due to time and resource limitations.

When considering the advantages of programming robots in a natural way, like we humans do for teaching complex skills (e.g., requiring fast dynamics and dexterity) to others, the possibilities are not limited to recording demonstrations, for later fitting a policy model, as it is done in traditional \gls{il} methods \citep{argall2009survey}.
In practice, an initial set of demonstrations or instructions tend to suffice to teach very simple and easy tasks from human to human, e.g., the instructions for opening a door, plugging a phone charger, or the user guide for most devices we use on a daily basis.
Nevertheless, for complex skills such as playing a sport, a loop of interactions is required for learning, because then the teacher explains/shows the student what to do by directly correcting/evaluating its actions, improving its behavior over past mistakes and successes. Otherwise, considering and explaining all possible scenarios in advance would be intractable for both the teacher and the student.

This kind of teaching is based on different types of teaching feedback, like demonstrations, sporadic corrections, or evaluations (grading) with value judgments or rankings. As an example, when teaching a complex skill like playing tennis, various steps can be involved. The teacher shows full demonstrations of the stroke themselves to the learner. When the student tries to replicate the example, the teacher can show what a better execution would look like. After the student performs the stroke, the teacher could advise with voice instructions to slightly correct the angles, velocities, or forces of the movement. Moreover, the teacher can sporadically congratulate the student or make it clear that some decisions were not so good.
This kind of interactive teaching approach seems to be, for humans, the most natural strategy for teaching to perform more complex skills; therefore, it is desirable to teach robots in the same way.

In recent years, the domains of robotics and \gls{ml} have increasingly adopted and developed these interactive teaching strategies, as can be observed in Fig. \ref{fig:history_iil}.
In this paper, \gls{iil} refers to all the methods that include the teacher in the learning loop for training sequential decision-making systems.
The objective of this work is to survey the literature on these methods and to present the most relevant observations in an organized structure.

The study of \gls{iil} methods has increased and the community has grown because these strategies introduce additional benefits with respect to learning paradigms such as traditional \gls{il}.
Some of those advantages are: 

\begin{itemize}
    \item A more natural or intuitive teaching approach.
    \item Enabling users who are non-experts at demonstrating the task to teach successful policies.
    \item Obtaining richer datasets consisting of data from situations that are not faced when learning from full demonstrations, as the distribution of data collected is induced by the learner instead of the teacher, avoiding data mismatch issues (see Chapter \ref{sub:covariateshift}).
    \item More flexibility to the teachers, who are not constrained to use only demonstrations for transferring their knowledge, but they can use other kinds of feedback, like relative corrections, human reinforcements, or comparisons.
    \item Offers alternatives to solve the correspondence problem that exists between the space where teachers can give demonstrations and the space where the robot executes the actions.
    \item Some methods have more tolerance for the teacher's mistakes or provide a better approach to compensate for them.
\end{itemize}

Nonetheless, there are certain challenges that should be considered when a teacher is in the learning loop. Human teachers can be inconsistent and make mistakes, there is uncertainty in their input that tries to explain their intention, they need to learn to adapt to the changing behavior of the learning agent, and the learning process is open-ended \citep{dudley2018review}.

In this paper, we review the context that defines the domain of \gls{iil} and how it relates to other known learning approaches.
We highlight the most relevant aspects to be considered for teaching an agent interactively and organize the methods according to them. 
This study is based on grouping and surveying the most relevant established papers in the literature, along with more recent follow-up works that have shown promising contributions.
All these papers were gathered in a set of works used as reference for organizing the different classifications proposed throughout the different chapters. This set is also used for generating the tables in the Chapters \ref{sec:modes} and \ref{sec:ModelsLearned}, and the plot of Figure \ref{fig:history_iil}.

One of the reasons such organization of \gls{iil} methods does not exist so far is due to the varied terminology used by different authors to refer to some of these methods, which in many cases, only partially overlap.
Below, we introduce most of the names and keywords used to refer to the approaches that are relevant in this paper.

\section{Terminology Unification}\label{sec:IntroTerminology}

In the literature, there are many terms linked to \gls{ml} approaches that enable teachers to interactively shape learning systems.
As a consequence, many of them are used to describe similar learning problems, which makes it difficult for practitioners (especially beginners) to have a clear outlook of the field when studying the well-spread collection of related papers.
In this section, we introduce some of those terms and discuss how they relate to each other, group them into sets that partially overlap or contain some others, and provide a definition of \gls{iil}. 
Based on this definition and structure, we set the bounds of the topic of interest of this paper. 

\begin{figure}
    \centering
    \includegraphics[width=0.8\linewidth]{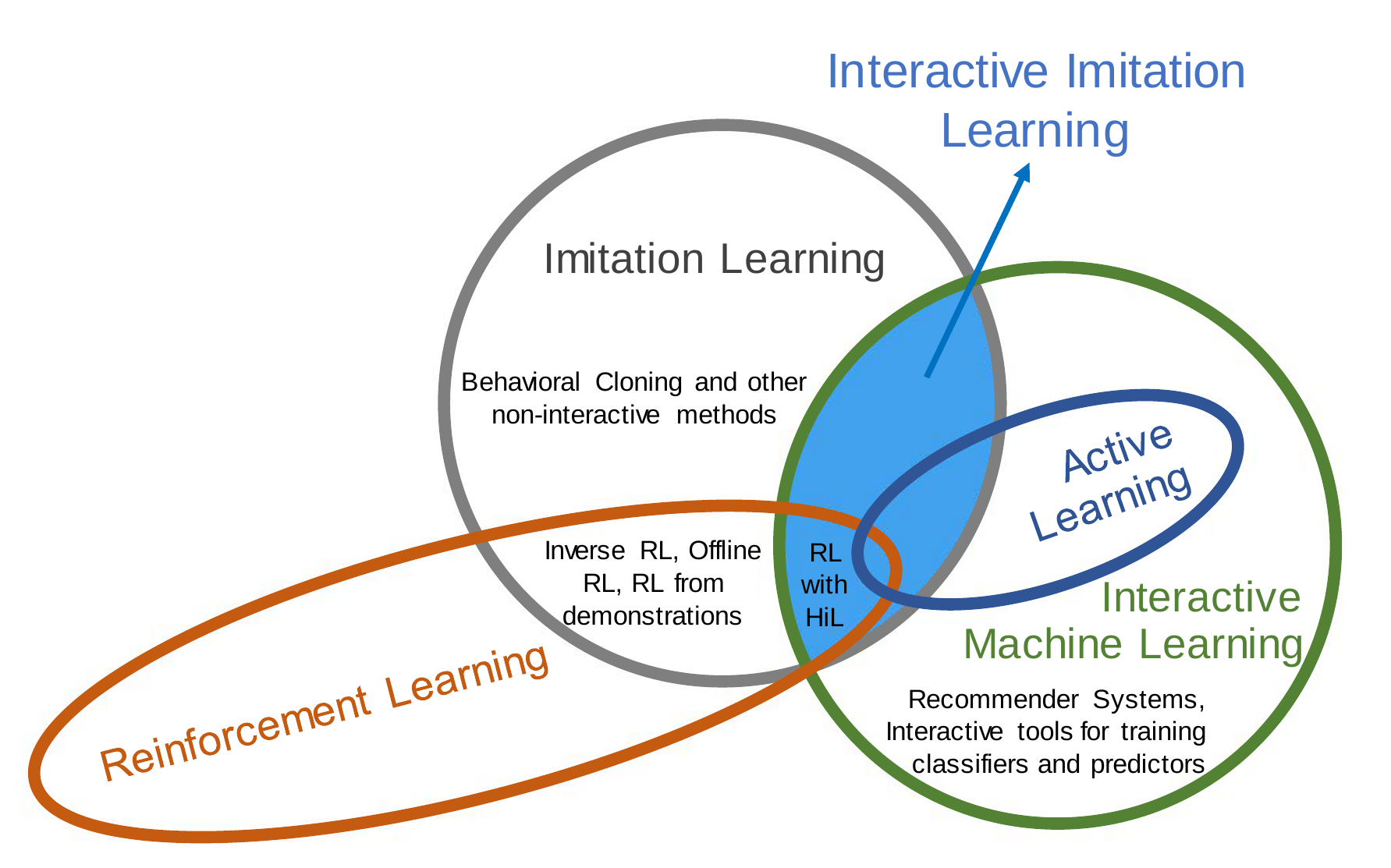}
    \caption{Relationship between different sets of learning paradigms related to the scope of this paper. The intersection of \gls{il} with \gls{iml}(blue area) is what defines the scope of this paper, called here \gls{iil}}
    \label{fig:LearningSets}
\end{figure}

Fig. \ref{fig:LearningSets} depicts with a Venn Diagram the relationship between all learning paradigms discussed below. 

\subsection{Imitation Learning}
In the context of robotics, the terms \gls{lfd}, \gls{pbd}, and \gls{il} are indistinctly used when referring to the paradigm of enabling robots to derive controllers from human demonstrations \citep{billard2013robot}.
Originally, these terms have been used by multiple authors referring to learning approaches that derive policies from datasets of explicit teacher demonstrations of a task.

Some recent methods enable human teachers to train robots through evaluative feedback, like \gls{lfc}, or \gls{interactive rl}, in which the teachers provide feedback that rates the desirability of the exhibited behavior during training time.
Although these approaches do not fully fit the literal meaning of \gls{lfd} or \gls{il}, some authors consider that evaluative feedback is just one of the demonstration modes a teacher could use within a learning process \citep{chernova2014robot}, therefore they also can be considered part of the world of \gls{il}.

Since \gls{il} is used at different levels of robot control and similar problems, we can rephrase the definitions of \gls{lfd}, \gls{pbd}, and \gls{il} as \emph{the set of \gls{ml} methods that leverage teacher's input as the source of knowledge for training sequential decision-making systems}.
Most of the time, the teacher is a human user, while in some cases it could be another decision-making agent (e.g., a computationally expensive policy like an \gls{mpc} or a planner system), and it has an understanding of either what are the objectives of the task, what to do, how good an action/policy is, or how good is the policy with respect to others.

In other words, methods are not considered \gls{il} if they leverage the input of a teacher to train non-sequential decision-making systems, e.g. an image classifier \citep{fails2003interactive}.

In the last two decades, articles have been published reviewing varied perspectives of \gls{il}, proposing categorizations for organizing the types of methods, identifying the benefits and drawbacks of the most known approaches, listing the open challenges, and introducing and structuring the field of study \citep{billard2008survey,argall2009survey,billing2010formalism,billard2013robot,chernova2014robot,amershi2014power,billard2016learning,hussein2017imitation,lee2017survey,calinon2018learning,osa2018algorithmic,li2019human,zhang2019leveraging, ravichandar2020recent}.

\subsection{Interactive Machine Learning}
There exists a considerable amount of learning methods that leverage human teachers within the learning loop for training sequential and non-sequential decision-making systems. 
Through different types of interaction, they make use of the knowledge a human has about the process, without the need to hard-coding it.
Therefore, these methods enable users who are not expert \gls{ml} practitioners to train models according to their insights and intuition.
The set of approaches that cover all the learning loop schemes involving humans transferring knowledge to the agent is known as \gls{iml} \citep{amershi2014power,fails2003interactive,ware2001interactive,holzinger2016interactive,dudley2018review,jiang2019recent}.

\citet{holzinger2016interactive} define \emph{``IML-approaches as algorithms that can interact with both computational agents and human agents and can optimize their learning behavior through these interactions''}.
\citet{dudley2018review} explain the contrast between \gls{iml} and classical \gls{ml} as \emph{``Interactive Machine Learning is distinct from classical machine learning in that human intelligence is applied through iterative teaching and model refinement in a relatively tight loop of set-and-check. In other words, the user provides additional information to the system to update the model, and the change in the model is reviewed against the user’s design objective''}.

Some other authors refer to the same domain with a more explicit name like \gls{hil-ml} \citep{xin2018accelerating,wu2021survey}.
Other authors refer to it in a more general way, combining the term \gls{ai}, e.g.,  with \gls{hil-ai} \citep{zanzotto2019human}, or \gls{iai} \citep{wenskovitch2020interactive}.
\gls{hcml} or \gls{hcai} is a larger domain that contains all the mentioned approaches with a human in the learning loop, additionally, it also includes the approaches based on \gls{ml}/\gls{ai} that have humans in the execution loop, i.e., systems that interact with humans as in \gls{ml}/\gls{ai}-based \gls{hci} or \gls{hri} systems.

Methods of \gls{iml} serve a wide domain of applications, including classification, regression, image processing, information retrieval, anomaly detection, among other systems \citep{ware2001interactive,fails2003interactive, amershi2012regroup, ngo2014efficient, amershi2014power,dudley2018review,jiang2019recent}. 
It is important to clarify that although \gls{iml} methods always include a human in the learning process, in some applications the human does not always perform as a \emph{teacher}, but rather is a user about whom the system learns through the interactions without explicit signals, as it is the case for Recommender Systems \citep{burke2002hybrid,bobadilla2013recommender,beel2016paper}.

Active Learning is one of the most traditional approaches of \gls{iml}, which consists of endowing the learner with capabilities for querying the teacher for more data in specific situations.
The learner is able to choose from which data samples it learns, allowing it to learn with higher accuracy from fewer samples \citep{cohn1996active,settles2009active}.

\subsection{Interactive Imitation Learning}
The set of \gls{iml} covers a broad spectrum of problems it can be applied to, including sequential and non-sequential decision-making. 
\gls{il} is narrower and specific to sequential problems. 
Unlike \gls{iml}, \gls{il} also involves methods that learn from teachers in a sequential manner, without the need for continuous interaction in the learning loop, as is the case of \gls{bc},  \gls{irl}, offline \gls{rl}, or \gls{rl} from demonstrations, which learn from a set of demonstrations that have been recorded before the learning process starts.

Also known as direct \gls{il}, \gls{bc} \citep{bain1995framework} applies supervised learning to a set of previously recorded expert demonstrations, in order to obtain a model that imitates the demonstrations.
In contrast, \gls{irl} is known as indirect \gls{il} because it uses recorded demonstrations to obtain an objective function or reward function that explains the goal of the task, so it can be used in an \gls{rl} process for obtaining a policy that imitates the demonstrator \citep{ng2000algorithms,zhifei2012review}.
In offline \gls{rl} the principles of classical online \gls{rl} are extended to be applied over datasets of demonstrations, without collecting any new sample during training time \citep{levine2020offline}.
We refer to \gls{rl} from demonstrations to the domain of all methods of classical online \gls{rl} that leverage recorded demonstrations to initialize the policy, or that keep that data in a buffer that is continuously used for updating the policy along with the new samples that are collected with the interactions \citep{kober2008policy,hester2018deep}.

The previous methods are not interactive, even though they learn from data demonstrated by teachers.
We hereby, take the term \gls{iil} that has been previously used in the literature and redefine it as the set of methods resulting from the intersection of the \gls{il} and \gls{iml} sets.
Therefore, we can say that \emph{\gls{iil} methods involve the approaches that learn from the knowledge provided by a teacher in the learning loop of a sequential decision-making system}.
Human teachers can transfer their knowledge to the learning agent through different modalities of interaction, and they are able to observe the effect of their feedback throughout the incremental learning process.

Methods of \gls{ml} that actively choose or query training samples are known as Active Learning \citep{settles2009active} methods, and they aim to increase the sampling efficiency of the learning process.
It is a subset of \gls{iml} that also overlaps with the \gls{iil} domain. 


It is important to make a distinction between \gls{iil}, \gls{iml}, and \gls{ils}, which is also used in the literature and sometimes referred to as learning from interactions, or interactive learning.
\gls{ils} are real/virtual entities that learn from the interaction with the world, a human, or another entity. This definition is complemented in \citet{cuayahuitl2013machine} with the description: \emph{"A machine can therefore be said to learn from interactions in a particular class of tasks if its performance improves with the given interactions over time”}.
The \gls{ils} that learn from the interaction with the world/environment enclose \gls{rl} methods \citep{sutton2018reinforcement}, wherein the agent learns from its own experience and not from a teacher.
The subset of \gls{ils} that learn from the interaction with other agents acting as teachers results in the same set of \gls{iil} methods, which are the focus of this paper.

\gls{rl} systems that obtain data from human teachers in the form of either demonstrations or evaluations (human reinforcements) during the learning process are known as \gls{hil-rl} and are also a type of \gls{iil}. 

\section{Others Surveys and Outline}\label{sec:other_surveys}
In recent years, there has been an explosion in the adoption of \gls{il} methods. 
There exist a large body of surveys discussing \gls{il} from different points of view. In particular \citet{chernova2014robot} provides a general overview of the methodology of learning from demonstration where different topics are analyzed, such as how the human teacher interacts with the robot to provide demonstrations, which modeling technique to choose (low/high level), how the human can refine an existing task and how to incorporate interactive and active learning components. Given the big spectrum of the paper of~\citet{chernova2014robot}, interactive methods are mentioned as one possible evolution of \gls{il}, but they are not the main focus of the work, and, therefore, not analyzed in depth.

A similar collection and analysis of the literature were conducted recently by \citet{ravichandar2020recent}. Here, topics such as non-expert robot programming, data efficiency, safe learning, and performance guarantees are discussed. The authors highlight the importance of learning from social cues, reasoning about the availability of human demonstrators, how to behave in their absence and how to ask for help. However, \gls{iil} is only marginally analyzed.

Similarly, \citet{hussein2017imitation} propose a survey on different learning methods for \gls{il}. The survey underlines how \gls{bc} has limitations due to errors in the demonstration and poor generalization. As a possible solution, it is proposed to combine \gls{il} with \gls{rl}, refine the policy with \gls{rl}, or use active learning. However, marginal attention is given specifically to interactive methods. 

In a recent survey, \citet{osa2018algorithmic} provide a structural analysis on \gls{il}, focusing on \gls{bc} and \gls{irl} methods. The authors mention that incremental and interactive learning methods can be employed to alleviate the \emph{covariate shift} problem (Chapter \ref{sub:covariateshift}) that exist in \gls{bc} methods. While they highlight the necessity of such methods from an algorithmic and mathematical perspective on machine learning, the authors do not provide an extensive treatment of the topic, as it is outside the scope of their work.

The topic of Human-Centered \gls{rl} is investigated by \citep{li2019human} as well as \citet{zhang2019leveraging}, where human evaluative feedback is used to teach behaviors to learning agents. They divide the field into three categories: learning from human reward, from interpreted human reward, and from action-translated human reward. Although these works are surveying the concept of human feedback from a \gls{rl} perspective, a broader discussion of other \gls{iil} methods is not covered. 

In our paper, we provide a survey of the Interactive Imitation Learning literature, ranging from seminal early work to the most recent advances. We investigate the role of \gls{iil} in the broader picture of sequential decision-making problems, with a focus on robotics applications. Besides providing an organized view of the state-of-the-art of the field, we aim to distill the most important takeaways and contribute a useful perspective on the topic. Our goal is for this manuscript to be a helpful reference for future work as well as a starting point for newcomers to the field. Our discussion spans multiple dimensions, ranging from the type of feedback a human teacher can provide to the agents, to the models that are learned through this interaction, to the existing benchmarks and applications proposed in recent years. In particular, we structure the analysis over multiple chapters as follows:
\begin{itemize}
\item Chapter \ref{sec:background} provides an overview of the sequential decision-making problem and its different formulations, formalizes the \gls{iil} problem and defines core concepts such as Feedback and Covariate Shift.
\item Chapter \ref{sec:modes} discusses the different modalities of feedback that a human teacher can provide to the robot, ranging from evaluative to preference to corrective feedback or interventions. We examine their strengths and weaknesses, with a focus on the trade-off between richness of information and human effort required.
\item Chapter \ref{sec:ModelsLearned} considers the various types of models that the robot is able to learn from the provided feedback, including policies, transition models, and objective functions. We discuss how certain models are best learned by specific types of feedback, and how they are used to achieve the main objective of solving sequential decision problems.
\item Chapter \ref{sec:AuxiliaryModels} reviews auxiliary models that the robot could learn in addition to the main objective, such as uncertainty and risk estimation models, environment dynamics, task features and models of the human teacher. We analyze the advantages that such models provide and the settings in which they can be adopted.
\item Chapter \ref{sec:model representations} discusses the different types of function approximation and model representation strategies common in the literature, including motion-conditioned models and deep neural networks. We consider their advantages and disadvantages and provide recommendations on their usage.
\item Chapter \ref{sec:on off policy learning} provides a comparison between on-policy and off-policy methods with a focus on the \gls{iil} setting.
\item Chapter \ref{sec:reinforcement learning with HIL} analyzes the special case of \gls{iil} methods used in gls{rl} framework, called RL with Human in the Loop.
\item Chapter \ref{section:interfaces} presents an overview of the interfaces used for enabling the communication between the robot/computer and the teacher, examining their role and importance in the learning pipeline. They range from physical contact with the robot embodiment to external devices such as remote controllers to contact-free approaches such as video and voice.
\item Chapter \ref{sec: user experience} provides an overview of the human factors to consider in \gls{iil}, such as available human-robot interfaces, user experience, and performance metrics, as well as guidelines on how to design user studies in \gls{iil}.
\item Chapter \ref{sec:BenchmarksAndApplications} surveys the principal benchmarks and datasets used in the literature to evaluate the proposed methods as well as the different fields of application of these algorithms, such as assistive, household, medical or industrial robots;
\item Chapter \ref{serc:research challenges} provides a discussion of the current challenges and opportunities in the field of \gls{iil}, as well as directions for future work.
\item Chapter \ref{sec:conclusions} completes the survey with a summary of the main concepts discussed as well as the most relevant takeaways and contributions to the field.
\end{itemize}
\newpage
\chapter{Theoretical Background}\label{sec:background}

In this chapter, first we formalize the sequential decision-making problem using the \gls{mdp} framework~\citep{bellman1957markovian}, and then formalize the \gls{iil} problem.

Many problems like solving Rubik's cube with a robotic hand, controlling the propulsion of a rocket, swinging up a pendulum or finding the best strategy in a chess game share the necessary idea of finding the best set of actions that would successfully accomplish the task. 
These problems share many properties and therefore they can be modeled using a common framework (i.e \gls{mdp}).

\section{Decision Theory} \label{sec:background-dmp}
A wide variety of problems can be formalized as a sequential decision-making process, where the decision-making authority is an \emph{agent}, operating in a certain \emph{environment}. At each time instance $t$ (also known as time step), the agent receives information describing the situation of the environment with the \emph{state} vector $s_t$, and executes an \emph{action} $a_t$, aiming to change the environment towards a desired state according to the goal of the task. 
The environment transitions to a new state $s_{t+1}$, and provides a reward $r_{t}$, which is a signal that explains the objective of the task.

When a decision-making problem has well-defined initial and terminal conditions, it is known as a \emph{finite horizon} problem, and the period of time between its start and end is called an  \emph{episode}.
The collection of states and actions experienced by the agent throughout an episode is known as a \emph{trajectory} $\tau=(s_0, a_0, ..., s_T, a_T)$, where $T$ corresponds to the number of time steps visited by the agent.

Decision theory provides a formal and complete framework for decision-making by combining probability and utility theory \citep{russell2016artificial}.

\subsection{Markov Decision Process (MDP)}

Initial foundations for \gls{mdp} are set by \citet{bellman1957markovian} and further extended by \citet{howard1960dynamic}. An \gls{mdp} models a stochastic, sequential decision-making process in a fully observable environment as a tuple $<\STATES, \ACTIONS, \TRANSITION, \REWARDS >$ with four components:

\begin{itemize}
	 \item $\STATES$: A set of all possible \emph{states} $\STATE$.
	 \item $\ACTIONS$: A set of all possible \emph{actions} $\ACTION$. Some problems may have a state-dependent set of actions ($\ACTIONS(\STATE)$). 
	 \item $\TRANSITION(\STATENEW \mid \ACTION,\STATE)$: A \emph{transition model} that defines $\prob(\STATE_{t+1} = \STATENEW \mid \STATE_{t} = \STATE, \ACTION_{t}= \ACTION )$, probability of reaching state $\STATENEW$ if action $\ACTION$ is applied in state $\STATE$ ($\TRANSITION: \STATES \times \ACTIONS \times \STATES \mapsto [0,1]$).
	 \item $\REWARDS(\STATE, \ACTION)$: A \emph{reward function}, determining the \emph{reward} $r$ received for applying action $\ACTION$ in state $\STATE$ ($\REWARDS: \STATES \times \ACTIONS \mapsto \REAL$).
\end{itemize}
Decisions are modeled as state-action pairs $(\STATE, \ACTION)$. The next state $\STATENEW$ is determined by a probability distribution, which is defined by the transition function $\TRANSITION$ and it is based on the current state $\STATE$ and the applied action $\ACTION$. The Markov property defines that the next state $\STATENEW$ is dependent only on the current state and action, while the previous states and actions do not have any influence on the current transition.
A deterministic policy $\POLICY$ is a mapping from states to actions, defining which action should be chosen in that state in $\STATES$ ($\POLICY: \STATES \mapsto \ACTIONS$). 
The policy can also have a stochastic representation, with a distribution over state-action pairs ($\POLICY: \STATES \times \ACTIONS \mapsto [0,1]$).

In certain problems, the agent cannot directly observe the underlying state. Instead, the state can only be inferred indirectly, using an observation model (the probability distribution of different observations given the underlying state). For such problems, it is appropriate to consider the \gls{pomdp} \citep{lesliepackkaelbling1998planning}, where the dynamics are still described using \glspl{mdp}, and the additional observation model $\OBSERVATIONS(\STATE, \OBSERVATION)$ specifies the probability of perceiving an \emph{observation} $\OBSERVATION$ in a state $\STATE$.

\subsection{Sequential Decision-Making Problem}

The objective of solving decision-making problems modeled with \glspl{mdp} or \glspl{pomdp} is to find the policy $\POLICY^*$, that maximizes the expected accumulated reward (also known as \emph{utility} or \emph{value}), i.e., 
\begin{equation}
    \POLICY^* = \argmax_{\POLICY \in \Pi} \EXPECT_{\tau \sim p_{\POLICY}(\tau)} \left[\sum_{t=0}^{T}\decay^{t}\REWARDS(s_{t}, a_{t})\right],
    \label{eq:rl_objective}
\end{equation}

where $p_{\pi}(\tau)$ corresponds to the trajectory distribution induced by $\pi$, $\Pi$ to the set of possible policies, and $\decay$ is a discount factor. Since problems, in general, can have \emph{infinite horizon} ($T \to \infty$), this sum could diverge. Therefore, the sum can be discounted with the discount factor $\decay$, where $0 \leq \decay < 1$.
This problem formulation assumes the reward hypothesis, which claims that \emph{``all of what we mean by goals and purposes can be well thought of as maximization of the expected value of the cumulative sum of a received scalar signal (reward)''} \citep{sutton2018reinforcement}. 

There exists a plethora of methods for solving sequential decision-making problems, each with unique assumptions, strengths and weaknesses. We make rough distinctions between three different approaches: \emph{Control}, \emph{Planning} and \emph{Learning}.
%
%
The boundaries of these approaches are not always clear as many practical solutions commonly lie in between or combine these approaches. In this work, we focus on Interactive Robot Learning approaches, where the goal is to devise a policy by learning from interactions with humans.

\section{Interactive Imitation Learning}\label{sec:background-iil}

\subsection{MDPs in IIL}\label{sec:mdp_iil}
\begin{figure}[t]
	\centering
  	\includegraphics[width=0.5\linewidth]{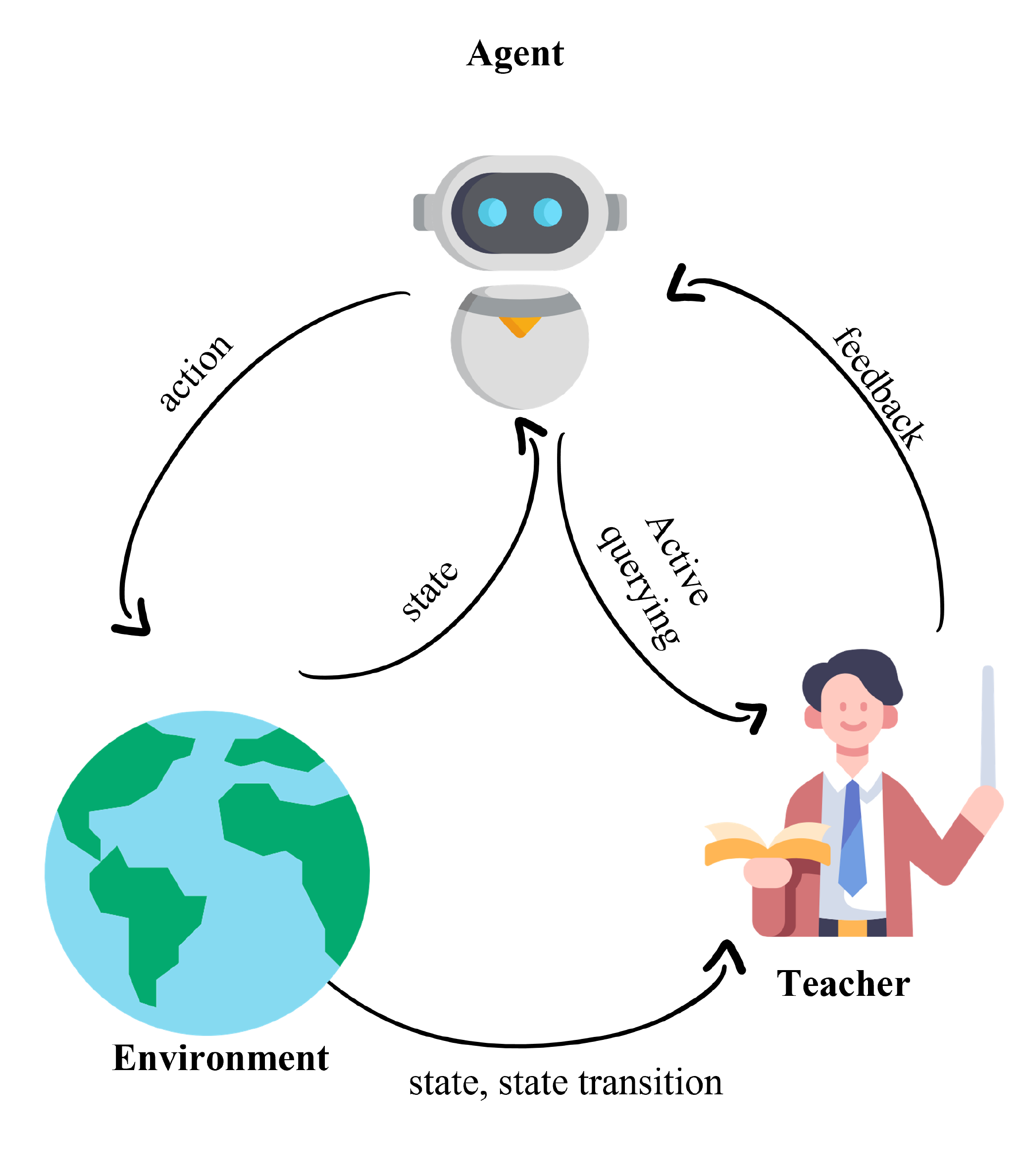}
	\caption{\gls{iil} learning loop.}
	\label{fig:IIL}
\end{figure}

In \gls{iil}, a human teacher, which we will refer to as \emph{the teacher}, aims to improve the behavior of a learning agent, which we will refer to as \emph{the learner}, by occasionally providing feedback to it as a function of the observed behavior (see Figure \ref{fig:IIL}). The period of time when the teacher provides feedback to the learner is known as the \emph{learning process}, which finishes whenever the human considers the learner's behavior appropriate or when no more improvement is observed. The human feedback can be modeled with the \emph{feedback function} $\mathcal{H}$. Although $\mathcal{H}$ can evolve throughout a learning process (i.e., a human may modify its understanding of a task when teaching), for simplicity, the following of this chapter assumes this function does not change.

\begin{figure}[t]
	\centering
  	\includegraphics[width=0.45\linewidth]{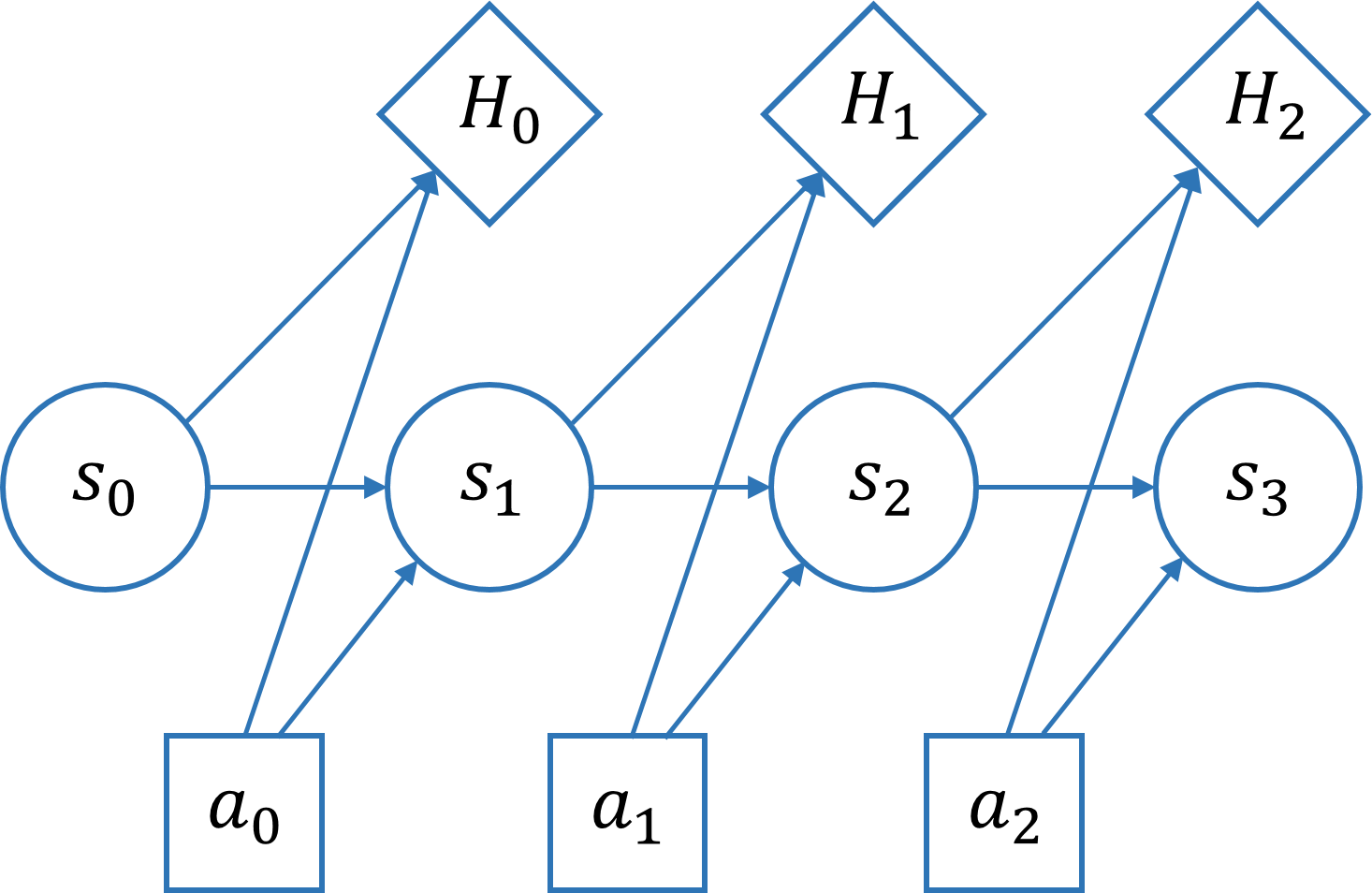}
	\caption{MDP with the feedback signal $H_{t}$.} 
	\label{fig:interactiveMDP}
\end{figure}

$\mathcal{H}$ is presented as a more general alternative to the reward function employed in the \gls{mdp} framework. At every time step, as a consequence of the agent's behavior, $\mathcal{H}$ outputs a \emph{feedback signal} $H_{t}$, which is defined as any type of information that can be used to improve the agent's policy (see Figure \ref{fig:interactiveMDP}. In \gls{iil}, feedback can be occasional; therefore, $H_{t}$ consists of two values: $h_{t}\in \REAL^n$ and $g_{t}\in \{0, 1\}$. $h_{t}$ provides the information employed to improve the agent's performance and $g_{t}$ indicates the instances where feedback was given, i.e., $h_{t}$ exists whenever $g_{t} = 1$. 
Furthermore, and differently from the reward function in \gls{rl}, $\mathcal{H}$ may depend on previously visited states $s_{\leq t} \equiv (s_{0}, s_{1}, ..., s_{t})$ and actions $a_{\leq t} \equiv (a_{0}, a_{1}, ..., a_{t})$. Hence, in the deterministic case, $\mathcal{H}(s_{\leq t}, a_{\leq t}, s'_{t}): \STATES^{t+1} \times \ACTIONS^{t+1} \times\STATES \mapsto \REAL^{n} \times \{0, 1\}$ (note that the domain can be a subset of $\STATES^{t+1} \times \ACTIONS^{t+1} \times \STATES$, where $X^{t}$ represents $X$ to the power of $t$). Alternatively, $\mathcal{H}$ can be modeled as a probability distribution $\mathcal{H}: \STATES^{t+1} \times \ACTIONS^{t+1} \times \STATES \times \REAL^{n} \times \{0, 1\} \mapsto [0, 1]$. Finally, note that this formulation can be extended to cases where the agent generates active queries, where $\mathcal{H}$ would also depend on them.

Given that $h_{t}$ does not necessarily represent a reward, the problem formulation of the MDP needs to be modified accordingly. The next subsection discusses how to approach this problem.

\subsection{Interactive Imitation Learning Objective}\label{sec:iil-objective}

The goal of sequential decision-making problems is to find a policy $\pi$ that generates trajectories $\tau \sim p_{\pi}(\tau)$ such that an objective function $J(\pi)$ is minimized. In \gls{rl}, for instance, the objective function is defined by the policy's (negative) expected return. In \gls{iil}, however, there is not always direct access to this function, as it is commonly represented implicitly inside the teacher's mind and, therefore, it is not always possible to minimize it directly. Consequently, more generally, it is possible to formulate the problem in terms of an observable \emph{surrogate loss} $L(\pi, \mathcal{H})$ computed as a function of the feedback function $\mathcal{H}$. We assume that the minimization of $L(\pi, \mathcal{H})$ indirectly minimizes $J(\pi)$ (or at least leads to near-optimal solutions). Note that when the true objective function of the problem is available, these two functions are the same (i.e., $L = J$). Hence, \gls{iil} aims to find a \emph{learner's policy} $\pi^{l}$ by solving the following optimization problem:
\begin{equation}
    \pi^{l*} = \argmin_{\pi \in \Pi} L(\pi, \mathcal{H}).
    \label{eq:gen_objective}
\end{equation}

One key aspect of this equation is the approach employed to search through the space of solutions $\Pi$. In practice, when learning this sequential decision-making problem, the data used to optimize eq. \ref{eq:gen_objective} comes from a policy that interacts with the environment, which biases the optimization problem. Hence, depending on this policy, different solutions will be obtained. 

To make this idea evident, the problem can be formulated in terms of the expected immediate cost $C(s,a)$ of performing an action $a$ for a state $s$ \citep{ross2011reduction}. Then, we can express this cost in terms of a policy $\pi$ with $C_{\pi}(s) = \EXPECT_{a\sim \pi(s)}\left[C(s,a)\right]$. Consequently, the objective function becomes the accumulated expected immediate cost $J(\pi) = \sum^{T}_{t=0}\EXPECT_{s\sim p_{\pi}^{t}(s)}\left[C_{\pi}(s)\right]$, where $T$ corresponds to the task horizon and $p_{\pi}^{t}(s)$ is the state distribution at time step $t$ induced by $\pi$. Once again, given that we might not have access to $C_{\pi}(s)$, the problem can be formulated in terms of the immediate expected surrogate loss $\ell_{\pi}(s) = \EXPECT_{a\sim \pi(s)}\left[\ell_{\pi}(s, a, \mathcal{H}(s, a))\right]$, yielding the following \gls{iil} optimization problem:
\begin{equation}
    \pi^{l*} = \argmin_{\pi \in \Pi} \sum^{T}_{t=0}\EXPECT_{s\sim p_{\pi}^{t}(s)}\left[\ell_{\pi}(s)\right].
    \label{eq:gen_objective_acc}
\end{equation}

In practice, the expected value of the surrogate loss in Eq. \ref{eq:gen_objective_acc} is estimated from the data collected by a policy that interacts with the environment (i.e., $p^{t}_{\pi}$ is induced by this policy). For instance, \gls{bc} methods use the \emph{teacher's policy} $\pi^{h}$ to collect training data (i.e., $s\sim p_{\pi^{h}}^{t}(s)$), or, in other words, the data comes from executions of the task performed by the teacher.

It turns out that methods like \gls{bc} that learn from data gathered with a policy different from the one that is later evaluated (i.e., $\pi^{l}$) suffer from \emph{covariate shift}. In this context, covariate shift means that the distribution of states visited at evaluation time by the learner (i.e., $s\sim p_{\pi^{l}}^{t}(s)$) differs from the one from where the training data was sampled from (e.g., $s\sim p_{\pi^{h}}^{t}(s)$). As a consequence, the learner visits states that were not well represented in the training data, leading, possibly, to catastrophic mistakes. For more details regarding covariate shift, we refer the reader to Section \ref{sub:covariateshift}.

Therefore, in \gls{iil}, the training data distribution depends on the learner's policy. In this way, these methods aim to minimize the state distribution mismatch between the data sampled at training and test time. Nevertheless, this poses a \emph{chicken-or-the-egg} problem, since without knowing the learner's policy in advance, it is not possible to generate data from the trajectories that this policy would generate \citep{ross2010efficient}. \gls{iil} methods address this by solving the problem iteratively, i.e, the learner's policy is used to collect data, improve its behavior from the data and repeat this process $N$ times until a well-performing policy is obtained. Hence, by noting that $\sum^{T}_{t=0}\EXPECT_{s\sim p_{\pi}^{t}(s)}\left[\ell_{\pi}(s)\right]=T\EXPECT_{s\sim p_{\pi}(s))}\left[\ell_{\pi}(s)\right]$, where $p_{\pi}(s)=\frac{1}{T}\sum_{t=0}^{T}p^{t}_{\pi}$ corresponds to the average distribution of states \citep{ross2011reduction}, the general \gls{iil} problem can be formulated as:  
\begin{equation}
    \textbf{IIL problem: }\pi^{l*} = \argmin_{\pi \in \Pi} \sum_{i=1}^{N}\EXPECT_{s\sim p_{\pi_{i}^{l}}(s))}\left[\ell_{\pi}(s)\right].
    \label{eq:gen_objective_inter}
\end{equation}
Note that in this equation there is an abuse of notation, as $p_{\pi_{i}^{l}}(s)$ represents a distribution of states that \emph{depends} on $\pi_{i}^{l}$, but the actions taken for collecting training data do not always necessarily have to distribute exactly as $\pi_{i}^{l}$. 

From Eq. \ref{eq:gen_objective_inter} it can be observed that every \gls{iil} method has the following properties:

\begin{enumerate}
    \item A surrogate loss $\ell_{\pi}$ is computed as a function of the feedback function $\mathcal{H}$.
    \item The problem is formulated over state distributions that depend on the learner's policy.
    \item The problem is solved iteratively by sampling, at each training iteration, from state distributions that depend on the current learner's policy.
\end{enumerate}

\subsection{Episodic Feedback}\label{sec:episodicfeedback}
A family of \gls{iil} methods solves Eq. \ref{eq:gen_objective} by solving the inverse problem, i.e., $L(\pi,\mathcal{H})$ is unknown and human feedback is employed to estimate $\hat{L}(\pi,\mathcal{H})$. Then, this estimation is minimized $\hat{L}(\pi,\mathcal{H})$ with some optimization method (e.g., path planning or \gls{rl}). Commonly, several trajectories are sampled from the learner's policy $\pi^{l}$ to get Monte Carlo estimates of $L(\pi,\mathcal{H})$ and feedback is provided at the end of them, i.e, $g_{t}=0$ the rest of the time. This feedback consists of an evaluation over the complete trajectory that has the form of a choice/preference \citep{wilde2022learning,wilson2012bayesian,christiano2017deep}, i.e., at each iteration, given the execution of two or more trajectories from the learner, the teacher provides a ranking of them. Then, the feedback is employed to gradually shift the trajectories generated by the learner in a direction where their performance will increase.   

\subsection{Per Step Feedback}\label{sec:perstep}
Alternatively, many \gls{iil} methods directly solve Eq. \eqref{eq:gen_objective_acc} following approaches that were derived either from \gls{rl} (value maximization) or the classical \gls{il} (divergence minimization) literature. Therefore, in these cases, feedback is provided in a \emph{per-step basis}, i.e., the teacher observes the behavior of the learner at each time step and provides feedback if necessary.

\subsection{Value Maximization}
Value Maximization methods correspond to \gls{iil} approaches that employ human feedback to solve problems formulated using the \gls{rl} approach (see Eq. \ref{eq:rl_objective}). In other words, some part of the \gls{rl} problem is modified through $\mathcal{H}$. 

The most direct way of doing this is by \emph{naively} replacing the reward function of an existing \gls{rl} approach with $\mathcal{H}$ and executing the learning process as if nothing changed. However, prior research has shown that such methods may induce \emph{positive reward cycles}, which could lead to unintended behaviors \citep{ho2015teaching}. This shortcoming lead to the development of approaches that built upon the \gls{rl} literature but take into account this and other limitations in the method design. For more information regarding these methods, the reader is referred to Sections \ref{sec:modes} and \ref{sec:reinforcement learning with HIL}.

\subsubsection{Divergence Minimization}\label{sec:divergencemin}
The \gls{iil} methods that are derived from the literature of classical \gls{il} can be modeled as a divergence minimization problem where we assume that we have access to expert trajectories from $\pi^{h}$. Then, the problem is modelled as minimizing the distance between the trajectory distribution of the expert/human $p_{\pi^{h}}(\tau)$ and the learner $p_{\pi^{h}}(\tau)$. The $f-divergence$ family \citep{liese2006divergences} is a class of divergences that measure distances between probability distributions. Hence, the \gls{il} problem can be seen as an \emph{f-divergence minimization problem} \citep{ghasemipour2020divergence,ke2020imitation}. By denoting the f-divergence between two distributions as $D_{f}(\cdot,\cdot)$, \gls{il} can be formalized as:

\begin{equation}
    \pi^{l*} = \argmin_{\pi \in \Pi} D_{f}\left(p_{\pi^{h}}(\tau), p_{\pi}(\tau)\right).
    \label{eq:iil_distance}
\end{equation}

\gls{bc} methods solve Eq. \ref{eq:iil_distance} by using the \emph{forward Kullback–Leibler divergence} (KL), which reduces the problem to the \gls{mle} of the teacher's policy from samples drawn from the trajectory distribution induced by the teacher's policy \citep{bishop2006pattern}, i.e., 
\begin{equation}
    \pi^{l*} = \argmax_{\pi \in \Pi} \EXPECT_{\tau \sim p_{\pi^{h}}(\tau)}\left[\sum_{t=0}^{T} \ln \pi(a_{t} | s_{t})\right].
    \label{eq:bc}
\end{equation}
 
Interestingly, if the \emph{Total Variation} (TV) distance between these distributions is minimized instead, the problem reduces to the minimization of the forward KL divergence between the teacher and the learner policies from a state distribution that follows the learner's policy \citep{ke2020imitation}
\begin{equation}
    \pi^{l*} = \argmin_{\pi \in \Pi} \EXPECT_{s \sim p_{\pi}(s)}\left[ D_{KL}\left(\pi^{h}(a|s),\pi(a|s)\right)\right].
    \label{eq:dagger_form}
\end{equation}
Notably, if we define the surrogate loss of an \gls{iil} problem as $\ell_{\pi}= D_{KL}\left(\pi^{h}(a|s),\pi(a|s)\right)$, then we would have an \gls{iil} method that minimizes the TV divergence between the teacher's and the learner's policy. The method \gls{dagger} \citep{ross2011reduction} minimizes this objective function, which inspired a broad family of \gls{iil} methods. 

In this case, the feedback function directly outputs a desired action for a given state, i.e., $h_{t}$ corresponds to a sample from $\pi^{h}(a|s)$. The samples $h_{t}$ can be employed to estimate $\pi^{h}$ by solving the \gls{mle} problem. Alternatively, $\pi^{l}$ can be modeled as a deterministic policy. In such cases, the samples are approximated by the minimization of a distance between $\pi^{l}$ and $h_{t}$ (e.g., \gls{mse} minimization). This approach can indirectly solve Eq. \eqref{eq:dagger_form} if some assumptions are made; for instance, if it is assumed that $\pi^{l}$ follows a Gaussian distribution with fixed variance, solving Eq. \eqref{eq:dagger_form} is equivalent to finding a discrete policy that models the mean of this distribution through \gls{mse} minimization \citep{osa2018algorithmic}.

\subsection{Covariate Shift}\label{sub:covariateshift}
One of the main advantages of using \gls{iil} over offline \gls{il} methods is its data efficiency. Here, data efficiency is evaluated as the amount of human data (i.e., feedback) that is required to obtain well-performing policies (according to the human's judgment). Offline \gls{il} methods require large amounts of data because of covariate shift. Covariate shift is defined as the prediction problem where the source and target domain probability densities are different \citep{sugiyama2015introduction, osa2018algorithmic} and, therefore, the learned model is required to make predictions with inputs that diverge from the distribution of the training data.

In the context of offline \gls{il}, the source domain corresponds to $p_{\pi^{h}}(\tau)$ and the target domain corresponds to $p_{\pi^{l}}(\tau)$, i.e, the learner learns from samples that are drawn from the teacher's policy. However, given that the learning problem will always be subject to errors, at prediction time, the states that the learner visit will gradually diverge away from the states presented in the training data, leading the agent to visit unknown regions and, therefore, to possibly make catastrophic mistakes. To solve this problem, it is necessary to collect enough data such that the state space is covered as much as possible, which in complex problems can become intractable or require large amounts of resources.

Alternatively, in \gls{iil}, the learner collects data following its own distribution $p_{\pi^{l}}(\tau)$. Hence, by learning iteratively following this strategy, the agent learns to correct its mistakes and avoid regions that may be dangerous or that will not lead to completing the task. In this case, it is not necessary to collect data over the complete state space, but only in the regions that $\pi^{l}$ visits, and, therefore, fewer data is required to obtain well-performing policies.

In the next chapter, different methods that solve either the Value Maximization or the Divergence Minimization problem, along with the type of feedback (i.e., \emph{feedback modality}) that they employ, are introduced. 
\newpage
\chapter{Modalities of Interaction}\label{sec:modes}

In the \gls{iil} literature, there exist various modalities of interaction that a human teacher can adopt to communicate with the learning agent. In this Chapter, we aim to provide a classification of these methods by answering the question \emph{what kinds of feedback could a teacher use to train an agent interactively?} The feedback is the signal containing the information that human teachers explicitly communicate to the learning agent through a Human-Robot (or Human-Computer) interface.
Different kinds of feedback are useful for transferring knowledge to the agent depending on factors like the task complexity, the teacher's understanding or expertise about it, the potential of the teacher to learn through the training process, or the available interface for providing feedback.

The short answer to that question provides two main categories that group the learning methods. They are based on the domain of the feedback provided by the teacher, which could be either in the \emph{evaluative space} or in the \emph{transition (state-action) space}. 
The former covers the methods in which the teacher provides a signal of assessment or evaluation about \emph{how well} the agent performs, while the latter category gathers the methods that require the teacher to provide feedback that let the agent learn \emph{how to do} the task.

In both categories, there are two ways for the user to transmit the assessment or guidance to the agent. 
The teacher could provide feedback that is either relative or absolute.
In the relative feedback case, the teacher provides a signal that contains information about the direction the agent behavior should shift to, with respect to current or other policy executions, e.g., how good a policy/transition is with respect to others, or how a transition should be modified with respect to the current one.
However, since it is only a relative direction, it does not specify explicitly what the exact input-output mapping is that the model should fit to, and it might be required to gather many feedback samples to tune the final mapping, even for a specific state or state-action pair. 
On the other hand, the absolute feedback contains information about the current execution regarding the optimal behavior, implicitly known by the teacher.
The relative feedback requires a lower cognitive load (i.e., less mental effort) for teachers because it is less informative than the absolute counterpart, which makes it in some cases less data efficient.
In other words, the use of relative and absolute feedback can represent a trade-off between data efficiency and cognitive load of the teachers during the interaction.

Table \ref{tab:tableModalities} presents the four modalities a teacher can use for interacting with a learning agent, depending on the kind of information provided and the way it is represented (absolute or relative). Methods corresponding to each row will be discussed below in the Sections \ref{sub:EvaluativeDomain} and \ref{sub:state-actionDomain} respectively.



\begin{table}
    \caption{Modalities of interaction according to the kind of feedback.}
    \label{tab:tableModalities}
    \begin{tabulary}{1\linewidth}{|C|C|C|}     
        \toprule
        \textbf{Learning From Human} & \textbf{Absolute Feedback} & \textbf{Relative Feedback}  \\\midrule
        \textbf{Feedback in Evaluative Space} &  Reinforcements &  Preferences \\\hline
        \textbf{Feedback in Transition (State-Action) Space} &  Absolute Corrections &  Relative Corrections  \\\bottomrule
  \end{tabulary} 
\end{table}

\section{Human Feedback in Evaluative Space}\label{sub:EvaluativeDomain}

We first review different approaches that provide the agent with evaluative feedback, which consists of a scalar value indicating the quality of the agent's behavior. This family of methods tends to be confused with the set of \gls{rlwhil} methods since they partially overlap, however, the latter group is the result of different classification criteria.
In Chapter \ref{sec:reinforcement learning with HIL}, we focus to review the \gls{rlwhil} methods, which we define as the interactive methods that combine \gls{rl} and human input, and learn from both the reward function of the environment and the human feedback that can be of any kind of the modalities discussed in this Chapter, i.e., they are not restricted to learn only from human evaluations.

The earliest works in interactive learning belong to the evaluative feedback category and were inspired by animal clicker training, a common strategy used to train dogs and other domestic pets~\citep{pryor1999clicker}. 
Animal training for purposes like assistance or detection dogs is in fact proof that humans can transfer knowledge to other agents through simple signals specifying whether a behavior is acceptable or not, without explicit demonstrations of \emph{how to do} the task, unlike in traditional methods of \gls{lfd}.
Depending on the method, these feedback signals could be provided to evaluate a transition in a specific time step or a complete roll-out. 
They could be absolute evaluations of performance, or relative as in the cases where the teacher describes how some executions are better or worse than others, using either pair-wise comparisons or rankings.
	
Some of the early approaches that explored the use of evaluative feedback from human supervisors to interactively train agents are \gls{iec} \citep{takagi1998interactive,takagi2001interactive,smith1991designing}, which run a \gls{ga} in which the fitness function is given by the human after observing the performance of the individuals of each generation, i.e., the teacher evaluates each roll-out, and those evaluations are used along with the genetic operators in the search of population (solution) improvement. 
Already in the '90s and the beginning of this century, these methods were applied in a very wide spectrum of applications including graphic art and animation, music, database retrieval, industrial design, face image generation, control, and robotics, among many others \citep{takagi2001interactive}. 
		
The use of \gls{iec} to problems in robotics \citep{lewis1992genetic,ed1996cellular,kamohara1997control,lund1998evolutionary,nojima2003trajectory}, known as \gls{ier}, also attracted the interest of the researchers in those years, however, that domain of study has not been very active in the last decade.
Nevertheless, the community has been inclined to develop methods inspired by \gls{rl}, with the difference of receiving interactive reinforcements from the human in the learning loop, instead of receiving them from a reward function.
		
Using evaluative feedback provided by a human teacher simplifies two problems compared to an autonomous learning process like \gls{rl}, or an Evolutionary Strategy.
It helps to bypass the difficult problem of designing an objective function used for providing feedback in the autonomous case, and additionally, the implementation of the system is simpler since the infrastructure for computing the reward is not required.
However, the human evaluations can be inconsistent due to multiple external factors, or even not compliant with the \gls{mdp} framework, therefore, the algorithms should take into account these considerations in order to avoid convergence issues.
	
\subsection{Learning from Human Reinforcements}\label{subsub:HumanReinforcements}
\begin{figure}
	\centering
	\includegraphics[width=0.750\linewidth]{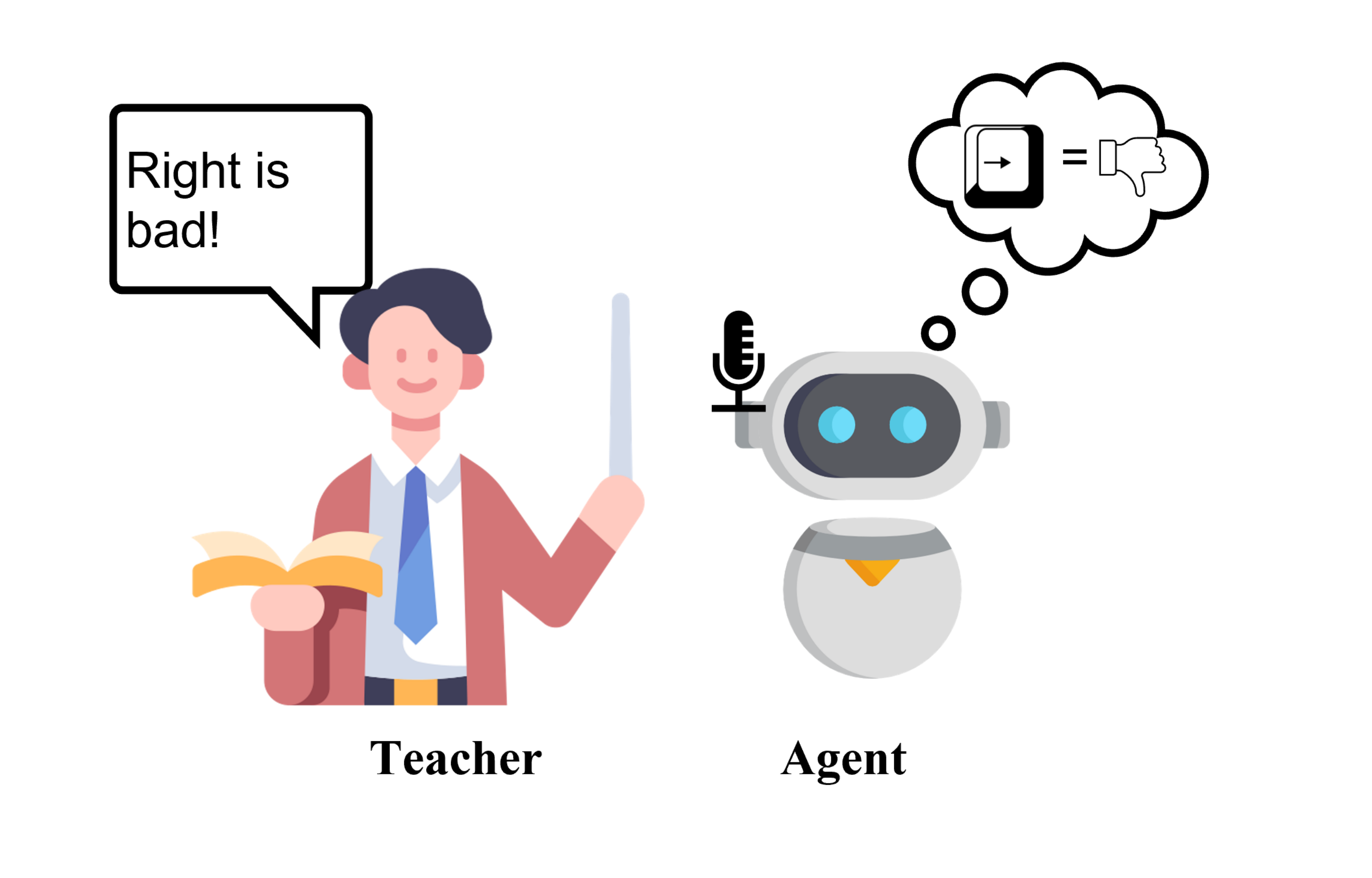}
	\caption{Learning from human reinforcements loop: the teacher is teaching the robot to go to the left and he gives bad rewards when it goes right. }
	\label{fig:mod_h_reinforcements}
\end{figure}

The methods based on absolute feedback in the evaluative space are approaches that take the human signal as a punishment or reward of the current policy execution with respect to the optimal policy implicitly known by the teacher, rather than an evaluation resulting from the comparison with other actions or policy executions (see Fig. \ref{fig:mod_h_reinforcements}). With respect to evolutionary strategies, using human reinforcements can handle the credit assignment problem better, which is ``the problem of assigning \emph{credit} or \emph{blame} for overall outcomes to each of the internal decisions made by a learning machine and which contributed to those outcomes'' \citep{haykin2001neural}.
The feedback is closely linked to the decisions that lead to positive or negative rewards, unlike using a scalar measure that represents the fitness of an entire roll-out of a \gls{ga}.

\subsubsection{Reinforcement Learning only with human rewards}
A simple way of introducing human feedback in the learning process of the agent is to adopt standard \gls{rl} algorithms and let a human teacher provide the reward signal.
Under this paradigm, non-expert users can teach the decision-making systems, even online, by delivering their feedback interactively as an evaluative (approval or disapproval) signal in a \gls{rl} framework where the reward is completely given by the human \citep{kaplan2002robotic,blumberg2002integrated,mitsunaga2008adapting, tenorio2010dynamic, leon2011teaching, suay2011effect, pilarski2011online, yanik2014gesture, hurtado2021from, londoo2022doing}. 

\citet{thomaz2006adding} show how \gls{interactive rl} enables a human user to provide positive and negative rewards in real-time in response to robot actions and to advise anticipatory guidance input that constrains action selection choice and guides the learner towards the desired behavior.
Since a human reward may have a different meaning with respect to an encoded environment reward function, which is the basic reinforcement used in the conventional \gls{rl} approaches, a series of works have analyzed how to model the human reinforcement \citep{thomaz2006reinforcement, thomaz2007asymmetric}.

\subsubsection{The TAMER Framework}\label{sec:tamer}
An important consideration to take into account for agents learning from human rewards, in contrast to learning from environment rewards is that ``\gls{mdp} reward is informationally poor yet flawless and human reinforcement is rich yet flawed'' \citep{knox2010combining}.
In other words, interpreting the human feedback in the same way as an \gls{mdp} reward signal does not take advantage of the prior knowledge of the teacher regarding the long-term consequences of actions.
The \emph{shaping} approach allows interactively training of an agent through signals of positive and negative reinforcement, which take into account the long-term consequences and optimality of the robot's action \citep{knox2009interactively}.
One of the seminal works based on \emph{shaping} is the \gls{tamer} framework \citep{knox2008tamer, knox2012learning}, which addresses how to use delayed human rewards in \gls{rl} problems with discrete action spaces. Other works combine human rewards and \gls{mdp} reward functions by applying transfer learning strategies, where both rewards were combined first sequentially \citep{knox2010combining}, and in a simultaneous scheme \citep{knox2012reinforcement}, in what they refer to as TAMER+RL.

The authors of \gls{tamer} studied the use of the discount factor used with the rewards in \gls{rl} \citep{knox2012reinforcement, knox2012learning, knox2013learning}. They firstly concluded that high discount (low discount factor) performs better for human reward functions used as \gls{mdp} reward. However, they presented a successful case of learning with a low discount from human reward \citep{knox2015framing}. \citet{knox2013training} present the first implementation of the \gls{tamer} algorithm on a real robot. 
	
Later on, \gls{actamer} proposes an Actor-Critic approach that addresses \gls{rl} problems with continuous action spaces and uses the same kind of feedback \citep{vien2012reinforcement, vien2013learning}.

Finally, \gls{d-tamer} \citep{warnell2018deep} leverages the representational capabilities of deep learning in order to deal with high-dimensional inputs such as images. The authors show that the method is able to solve an Atari game environment in just 15 minutes of human-provided feedback.

\subsubsection{Policy Shaping}
An additional line of work is the Policy Shaping framework, initially introduced by \citet{griffith2013policy}. In this work, evaluative human feedback is used to directly update the policy, instead of being considered as a reward or value. The stochastic human policy is trained by increasing or decreasing the probability of an action in a certain state, depending on the feedback provided. 
Building upon the Policy Shaping framework, \citet{loftin2014strategy, loftin2016learning} propose a method to take into account the user feedback strategy, in particular taking into consideration different interpretations of lack of feedback from the teacher. 
They derive two Bayesian policy learning algorithms called \gls{sabl} and \gls{isabl}, which are able to infer the trainer's strategy directly from the received feedback. 
This method is further extended by~\citet{macglashan2014training}, who demonstrate how grounding of natural language commands can be learned from a human trainer providing online reward and punishment. 
The combination of natural language commands and human feedback for training makes the training procedure simple and intuitive for non-experts users.
	
\subsubsection{Convergent Actor-Critic by Humans } \label{subsub:COACHevaluative}
Deriving from the ideas of Policy Shaping, \gls{coache}\footnote{The original acronym of this method is COACH, however, we here call it \gls{coache} (wherein the ``e'' refers to the use of evaluative feedback). This is in order to avoid ambiguities with \gls{coach} (that in this work we add the ``c'' referring to the use of corrective feedback in the action space), published earlier by \citet{celemin2015coach}, and mentioned in section \ref{subsub:RelativeCorrections}} is presented~\citep{macglashan2017interactive}. 
Contrary to prior work, this method assumes that the human feedback does not follow a static rule, but tends to evolve over time, depending on the current policy of the agent. 
In order to take into account the dependency of the provided feedback from the policy, they consider the human feedback to be a label on the advantage function. 
This idea is based on the insight that the TD-error used by actor-critic algorithms is an unbiased estimate of the advantage function, which is a policy-dependent value roughly corresponding to how much better or worse an action is compared to the current policy. 
Moreover, instead of using the feedback to learn an approximated advantage function, it is directly applied to the policy gradient update rule.
An extension of the aforementioned method is presented in~\citet{arumugam2019deep}, where the approach is enhanced with deep neural networks in order to cope with high-dimensional observations such as images. 

\subsubsection{Conclusion}
The use of human reinforcements allows teachers to transfer to the agent the insights of what is right or wrong in the respective time step. It requires a good level of understanding of the task but not necessarily being an expert or knowing what are the exact actions that should be taken in any state.

Since this feedback does not explicitly convey what other action should be executed in case the executed one receives a punishment, one mistaken punishment from the teacher requires many new instances of feedback to revert the wrong feedback effect, which means that these approaches are not very robust to imperfect teachers.

\subsection{Learning from Human Preference}\label{subsec:learning from human preference}
\begin{figure}
	\centering
	\includegraphics[width=0.750\linewidth]{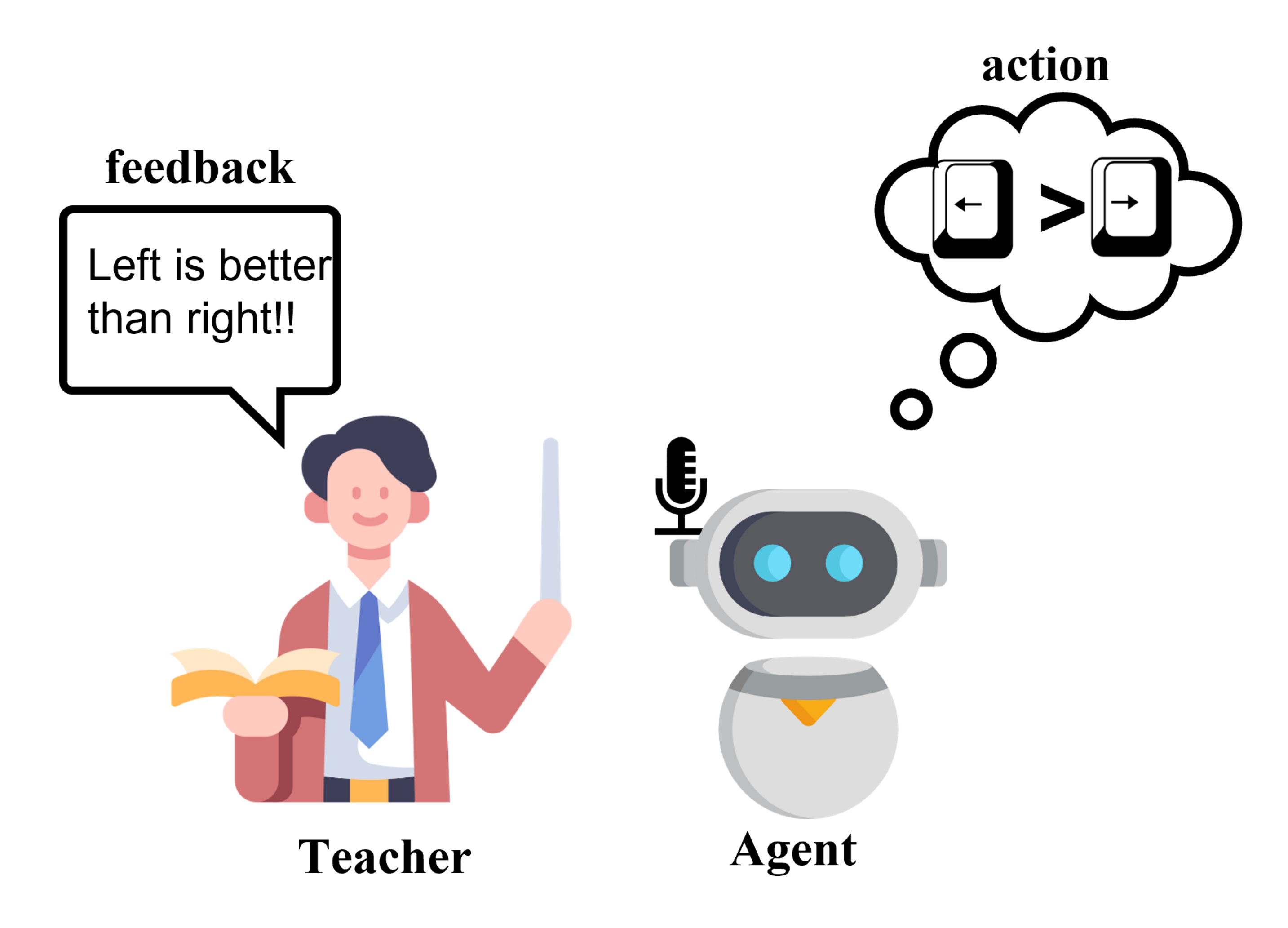}
	\caption{Learning from human preferences: the teacher is teaching the robot to take the left turn and it is specifically saying that going left is better than going right. }
	\label{fig:mod_h_preferences}
    \end{figure}

Learning from absolute evaluative feedback has shown great success, but it requires the human teacher to provide evaluations with respect to an absolute scale. 
On the other hand, methods for learning from human preference consist of comparing two or more sequences of actions and providing a preference score to the agent (see Fig. \ref{fig:mod_h_preferences}), and they do not require the teacher to identify and evaluate what is the credit of the decision at each time step with respect to the success or failure of the task execution, i.e., potentially reducing teacher workload. 
In other words, they use relative evaluative feedback that implicitly indicates the direction in which the solution in the policy space should be shifted, such that it matches the preferences of the teacher. However, since this feedback is relative to other trajectories, policies, or roll-outs, it does not describe how good an execution is in general, and a policy that is preferred over another or ranked as the best out of a set of policies might be ranked low later on with respect to some different executions.
	
	\subsubsection{Preference-Based Policy Learning}
	\gls{ppl}~\citep{akrour2011preference} is one of the first methods that investigate the integration between preference learning and reinforcement learning. In this work, a set of policies is shown to the human teacher who provides feedback as pairwise preferences between policies. With this information, the value of parametrized policies is estimated (here called Policy Return Estimate), the agent selects another set of policies, and the process is repeated. 
	Similarly, \citet{furnkranz2012preference} consider a preference-based reinforcement learning strategy, in which a policy is leaned through an approximate policy iteration setting, instead of learning a value function of the parametrized policies. Moreover, while the former work focuses on learning a ranking function of different policies at the trajectory level, the latter considers the preference feedback on actions in a given state.
	These ideas are extended by \citet{akrour2014programming}, where the agent estimates a utility function in order to account for unavoidable human mistakes. The method is evaluated on various testbeds, including a physical Nao robot, which was able to learn simple tasks such as raising a hand. 
	
	A line of work that focuses on manipulation tasks is the one presented by~\citet{jain2013learning, jain2015learning}. 
	The authors present a co-active online learning framework where the human teacher iteratively provides small adjustments to the trajectory currently proposed by the system. 
	Although these methods learn from the teachers applying preference learning, the interaction modality is also with relative corrections in the transition space, as explained in Section \ref{subsub:RelativeCorrections}, however, after recording the modified trajectory, it is compared and assumed to be preferred with respect to the one intended to be executed by the agent, hence, the data is used following the mechanisms of preference learning.
	
	The methods presented so far show how to train robot policies by preference-based feedback, but they are limited to low dimensional states as inputs. For approaching this issue, deep \glspl{nn} are incorporated into the methods, for instance, \citet{ibarz2018reward} proposes a method where the reward function is modeled as a deep \gls{nn} and is trained with a combination of demonstrations and human preferences. 
	The key idea is that the reward function used in the \gls{rl} step is not hand-crafted, but rather learned via preference feedback. The combination of demonstrations and preference feedback is evaluated on multiple Atari games and shown to outperform both demonstrations or preferences used in isolation.
	\citet{christiano2017deep} model both the policy and the reward function as a deep \gls{nn} to deal with high dimensional observations such as images. The reward function is updated from human preference feedback, whereas the policy is updated by a traditional \gls{rl} algorithm. A drawback of this approach is that the trajectories need to be sampled and replayed on screen for the user to provide feedback. Additionally, millions of steps are required for the policy to converge.
	Lastly, \citet{brown2020safe} propose Bayesian REX, a fast Bayesian Reward inference algorithm from preferences. Here, ranked trajectories are used to train low-dimensional feature embeddings via a self-supervised loss. The reward function is then constructed as a linear combination of such features.
	This group of approaches has been tested mainly in simulated environments, although their benefits could be extended to real-world robotic problems.
	
	\subsubsection{Information Maximization via Active Queries}
	An important component of preference-based learning methods is the choice of trajectories to compare. Some methods sample trajectories randomly from a dataset, other consider two consecutive trajectories generated by the current policy. 
	The goal of active preference-based methods is to improve the convergence of these algorithms by generating at each step the most informative query, as measured by information theoretic metrics such as expected volume removed~\citep{palan2019learning}.
	One of the early approaches in this direction is proposed by~\citet{akrour2012april}, who combine preference-based policy learning with an active ranking mechanism. 
	Another approach for policy learning from trajectory preferences is proposed by~\citet{wilson2012bayesian}. 
	Here, a Bayesian model is employed in order to actively query the human teacher, and two different query selection mechanisms are investigated. 
	The first one is called Query by Disagreement, where the main idea is to generate a sequence of unlabeled samples and evaluate them with two different classifiers. If the two models disagree on the class, then the expert is queried for a label.
	The second one is called Expected Belief Change, which aims to generate a set of candidate preference queries and heuristically select the best among those.
	
	A different approach is to use the provided preferences to learn a reward function, which can then be employed for training on the downstream task, for example in a standard reinforcement learning setting. One example of this approach is presented by~\citet{daniel2015active}, who propose a framework to actively learn a reward function in a bayesian optimization setting.
	\citet{sadigh2017active} present an active learning approach, where the agent decides on the trajectories to compare by maximizing the expected information gained from the query. This information gain is modeled as the volume removed from the hypothesis space by each query. The optimization problem is solved via an adaptive Metropolis algorithm~\citep{haario2001adaptive}. A novel aspect of this work is the complex nature of the queries since it deals with continuous trajectories of a dynamical system. The authors show that this method yields faster convergence to the desired reward compared to non-active approaches. An extension of this framework is batch active preference-based learning~\citep{biyik2018batch}, which aims to balance the number of queries to the human teacher and the number of total interactions. 
	This is achieved by batching multiple queries together in one request to the user. The advantage is faster iterations, and the procedure can be also parallelized when working with multiple users. 
	
	\citet{palan2019learning} introduce the \gls{dempref} framework, where demonstrations and preference feedback are combined to learn the weights of a linear reward function. The demonstrations are used to learn an initial prior over the space of reward functions as well as to ground the query generation process. The method is tested on different robotic manipulation tasks on a physical robot, and the additional use of initial demonstrations is shown to improve the sample efficiency of prior work.
	A further improvement on the active query process is provided by~\citet{biyik2020asking}, which explores an information gain formulation where the ability of the human teacher to respond to a certain query is included in the optimization process. For example, if two trajectories are very similar, it might be difficult for the teacher to provide preference feedback. This approach considers the trade-off between the robot and the human uncertainty and avoids questions that would become redundant.
	This idea is later extended by~\citet{biyik2020active}, where Gaussian Processes are used to model the reward function, as well as by~\citet{myers2021learning}, where multimodal reward functions are learned.

\subsubsection{Conclusion}
Learning from preferences demands very low prior knowledge from the teachers since the feedback is a general performance comparison of different roll-outs, e.g., even one bit of information is enough to state the preference out of two policies, which reduces the effort of the teacher, and widens the spectrum of people who could teach a robot.
Nevertheless, this feature comes with the credit assignment problem that evolutionary-based methods have, as it was mentioned at the beginning of the section. 
Preferences are a relative measure of performance that evaluate a sequence of transitions, therefore the feedback does not specify what decisions make one roll-out better than the other, and the algorithm has to identify them while compromising data efficiency.

Learning from preferences methods are also sensitive to mistakes in the teachers' assessments. 
In both Learning from human reinforcements and preferences, the mistakes in the feedback have a negative impact on the convergence of the process, reaching lower policy performances in a longer time.

\section{Human Feedback in Transition (State-Action) Space}\label{sub:state-actionDomain}

Human feedback in the transition space contains information about \emph{how to do} the task, i.e., explicit feedback that explains how a transition should be done, being it in the space of the actions, or the states.
Unlike in learning from evaluative feedback, with feedback in the transition space, there is no explicit quality assessment of the policy, the feedback signals represent the teacher's insights or understanding of the task execution.
This kind of feedback can be absolute, in which case the teacher is expected to demonstrate the optimal transition for the state the agent is currently visiting. 
Relative feedback, on the other hand, is used in cases where the teacher corrects the policy execution towards the considered right direction with respect to what the robot is executing in that time step. 
However, it does not assume that the correction is the optimal action, but rather a hint in that direction.
The correct action is reached after some iterations that accumulate the incremental progress of many relative corrections.

\subsection{Learning from Human Absolute Corrections}\label{subsec:AbsoluteCorrections}
\begin{figure}
	\centering
	\includegraphics[width=0.750\linewidth]{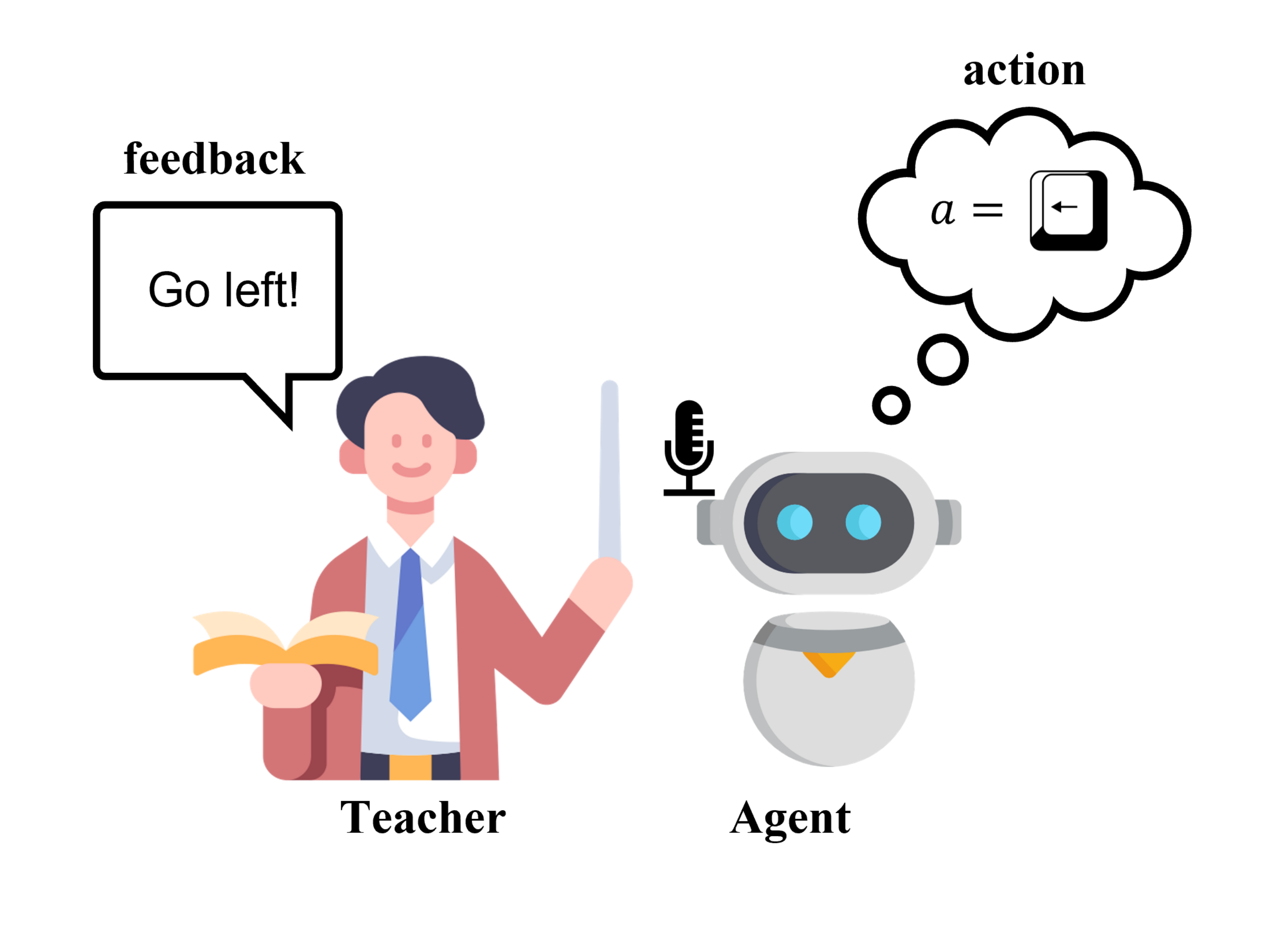}
	\caption{Learning from human absolute corrections: the teacher is explicitly telling the robot to go left.}
	\label{fig:mod_h_corrections}
    \end{figure}

In this kind of interaction, the agents are expected to receive explicit demonstrations of the task execution by the teacher, while the learning policy is controlling the agent, as shown in Fig. \ref{fig:mod_h_corrections}.
Depending on the method, the teacher can provide corrective demonstrations every time step, occasionally according to the teacher's own decision, or because the learner queries them.
Moreover, those demonstrations could be either only recorded, or recorded and executed. 
In the former case the agent executes the action from the current policy, whereas, in the latter, the agent replaces those actions with the ones demonstrated by the teacher, as in teleoperation mode.
These methods are the closest to standard \gls{lfd} methods like \acrlong{bc}, and some of them could even be considered a generalization of \gls{bc}.
	
\subsubsection{Corrective Demonstrations}\label{sec:corrective_demos}

One of the first approaches in this category is the Confidence-Based Autonomy framework presented by~\citet{chernova2009interactive}, which has two components, Confident Execution and Corrective Demonstrations. The first is a strategy that uses various thresholds to evaluate the confidence of the agent in a certain state, and in case it is too low, it stops the autonomous execution and queries the human teacher for additional demonstrations. 
Corrective Demonstrations are the second mechanism, which allows the teacher to provide corrections to any mistake of the agent. 
	
The idea of Corrective Demonstrations is further investigated by~\citet{mericcli2010complementary, mericcli2011task, mericli2011multi}, wherein they propose to leverage both prior hand-coded policies and corrective demonstrations.
Instead of only obtaining directly a policy from the demonstrations, it keeps the hand-coded policy as the primary behavior, which is only replaced by the demonstrations when the robot visits states that are similar to the ones in which the corrective demonstrations were recorded.
This approach was used to train the humanoid robot Aldebaran Nao to solve a complex ball dribbling task, and it shows improvement compared to a hand-coded controller.

Some of the most important methods for learning from corrective demonstrations are inspired or belong to a family of approaches based on \gls{dagger} \citep{ross2011reduction}, which interactively records the correct action demonstrations while a novice policy is controlling the agent.
\gls{dagger} was not specifically intended for human users, the teacher could be another expert policy, like a model-based controller or a planner system.
Indeed, many methods have been proposed after \gls{dagger}, since the requirement of human teacher input every time step is not the most user-friendly approach. 

The idea behind \gls{dagger} is to iteratively generate roll-out trajectories with the current policy, query the expert for corrections on the visited state-action pairs, and finally add the corrected actions to the dataset used for training the policy.
	
As with other methods in this section, this approach enables the expert to provide corrections on the states visited by the current policy, meaning that the data distribution is induced by the agent itself, drastically reducing compounding errors and distribution shift issues common in standard learning from demonstrations settings~\citep{ross2011reduction}.

\gls{dagger} requires a corrective demonstration from the teacher in every state, however, it uses a gating function based on a $\beta$ probability in order to control what action is actually executed, whether the action of the learner $\pi_R(s)$ or the one of the teacher $\pi_T(s)$.

At the beginning of the learning process, $\beta$ is set high for the robot to execute most of the expert teacher actions because the initial robot policies could make many mistakes that lead to dangerous or irrelevant states.
Through the iterations of the algorithm, $\beta$ is decreased to zero in order to give full control to the learning agent.
If $\beta=1$ all the time, \gls{dagger} performs exactly as behavioral cloning because the data distribution is completely induced by the teacher.

If the expert is a human, this is often unfeasible and prone to incorrect labels for robotic tasks, which usually operate at high control frequency, generating a large number of actions for each trajectory.
Most of the variants of \gls{dagger} (mentioned later) \citep{zhang2016query,menda2019ensembledagger,kelly2019hg,hoque2021lazydagger,hoque2021thriftydagger} differ from the original in i) the implementation of the gating function; ii) the way data is recorded, all aiming to improve workload, query efficiency, or safety.
	
The \gls{shiv} algorithm \citep{laskey2016shiv} is very similar to \gls{dagger}, however, it actively requests labels in states considered risky, instead of requiring labels every time step, reducing the human burden. 
The risk is defined when previously unseen states are visited, or when the policy model has a high surrogate loss in the area of the visited state. 
The method was validated in grasping tasks, outperforming the original \gls{dagger}.

A possible alternative is to monitor the policy execution and intervene when necessary, taking over control from the agent completely. 
This is a more natural and intuitive approach for a human teacher compared to labeling individual state-action pairs. 
This setting can be defined as learning from human intervention, and numerous studies have been presented to investigate such methods. 
	
There exist two main types of human intervention approaches: Human-Gated and Robot-Gated~\citep{laskey2017comparing}. 
Both types change the stochastic gating function based on the probability $\beta$ for executing either the action of the learner or the action of the expert, with a different strategy.

\subsubsection{Human-Gated Interventions}

Human-gated interventions allow the expert users to decide themselves when to intervene (control the agent). Its advantage is that safety is ensured by the expert, who is always ready to intervene in case of dangerous behavior.
	
\gls{hg-dagger}~\citep{kelly2019hg} is a direct extension of the \gls{dagger} algorithm, where the human teacher is in charge of intervening when the agent drifts away from the desired behavior. 
Every time an intervention occurs, the expert trajectory is recorded and stored in the training data set used to optimize the policy. 
Additionally, \gls{hg-dagger} learns a safety threshold of a risk metric, which could be used as a policy confidence metric for different regions of the state space.
The method is evaluated on both a simulated and a real-world autonomous driving task, showing better performance compared to behavior cloning and standard \gls{dagger}.
	
The assumption of the method is that the teacher does not intervene in the portions of the trajectory that are well executed.
A different approach is used in the \gls{iwr} framework \citep{mandlekar2020human}, where the robot's own experience is stored together with the teacher's interventions in the replay buffer. 
The authors show that storing such data has the advantage of reinforcing the already good behavior and improving the robustness of the policy, because more data is stored overall, and the data itself is distributed covering wider areas of the state space. 
Nevertheless, since the amount of intervention and non-intervention data is usually imbalanced, the authors propose a weighting parameter to prioritize the intervention samples. 
The method is evaluated on two challenging simulated manipulation tasks with low-dimensional observations, demonstrating better performance compared to \gls{hg-dagger} and to behavior cloning with complete demonstration.
	
\gls{iwr} works under the assumption that the teacher is always able to correct bad behaviors, which might not be true in general, since non-expert users might be in charge of training the robot. 
In~\citet{chisari2022correct} the \gls{ceiling} framework combines human interventions with evaluative feedback.
The use of evaluative feedback on non-corrected portions of the trajectory gives the human teacher the option to decide which part of the trajectory to use for training and which to discard. 
The method is shown to be able to train manipulation tasks from high-dimensional image observations directly in the real world in less than one hour of training.
	
Another related method is the \gls{eil} framework~\citep{spencer2020learning}. 
\gls{eil} aims to learn from the interventions as well as from the timing of the interventions since non-intervention constitutes useful information as well. 
They formalize a constraint on the learner's value function, which is used to differentiate \emph{good enough}, \emph{bad} and \emph{intervention} state-action pairs. 
The method is evaluated on a physical miniature car with a discrete action space, consisting of a library of 64 driving primitives. \gls{eil} is benchmarked against behavior Cloning and \gls{hg-dagger}, showing safe and more desirable trajectories.
Another recent work in the same category is \gls{shield}~\citep{luo2021robust}, which focuses on the problem of industrial insertion. It extends the \gls{ddpgfd}~\citep{vecerik2017leveraging} algorithm with a collection of different design choices, including on-policy corrections, i.e., the human can intervene to guide the agent back into the optimal region in case of deviations.

In Cycle-of-Learning \citep{goecks2019efficiently}, human-gated interventions are used for improving a policy obtained from demonstrations pre-recorded in the first stage.
The experiments with a perching task using a simulated drone showed that this approach has better performance than using either only demonstrations or only human interventions.
	
Corrective demonstrations are not only used for learning an explicit policy, but also for learning objective functions.
In \gls{land} \citep{kahn2021land}, the teacher takes over the control of autonomous navigation robots during failure situations. 
However, the data gathered during the interventions is not used for updating the policy, but for training a disengagement predictive model that is used as part of the cost function of the task, which is optimized during the decision-making with a model predictive control-based planner.

\subsubsection{Robot-Gated Interventions}

Robot-gated interventions require the agent to estimate when an intervention is necessary, which does not require constant attention from the teacher, since the robot is the one deciding when the intervention should be performed, allowing the human to supervise multiple robots at once~\citep{hoque2021thriftydagger}.
These methods generally require the agent to estimate a measure of performance, safety, or uncertainty about the currently observed state, which is then used to determine when to query or enable the human teacher control.
However, these kinds of approaches have to deal with the disengagement of the users, who do not react immediately when requested and require some time to be able to take over the system again.

One example of this approach is presented in~\citet{delpreto2020helping}, where the policy outputs a discrete vector of confidence scores for four different gripper orientations, and the one with the highest confidence is picked. An apprenticeship model is developed, which queries the teacher intervention in case of too many failures in a row or if the output confidence is lower than a certain threshold.

A variation of \gls{dagger} called \gls{safe dagger} \citep{zhang2016query} trains a classifier that predicts whether the learning policy deviates from the expert and, if it is the case, it switches the control to the expert in order to prevent executing unsafe actions. 
The authors mention that the metric used for comparing the expert and learning policy should depend on the task.
Experiments with a driving simulator showed that \gls{safe dagger} is safer and more efficient than \gls{dagger}.
\gls{ensemble dagger} \citep{menda2019ensembledagger}---a method that extends \gls{safe dagger}---uses the deviation classifier as a discrepancy rule, along with a doubt rule that also switches control from the learning policy to the expert teacher.
The doubt rule is computed based on the novelty/uncertainty of the policy, which is measured with the variance of an ensemble of neural networks.
The doubt rule lets the agent prevent executing dangerous actions in unseen states, in addition to the actions of the learning policy that tend to deviate from the expert teacher.

The \gls{lazy dagger}~\citep{hoque2021lazydagger} framework also extends \gls{safe dagger}, in particular, it aims to reduce context switching by adding noise to the actions provided by the supervisor to improve the data distribution as well as by adopting an asymmetric switching criteria, modeled as a hysteresis function. Finally, \gls{thrifty dagger}~\citep{hoque2021thriftydagger} is proposed, where the switching policy is learned instead. 
Interventions are queried in case the encountered state is sufficiently novel or risky. Similar to \gls{ensemble dagger}, novelty is estimated by computing the variance of each output of a set of policies, whereas the ``risk'' of a state is estimated by learning a Q-function to evaluate the discounted probability of success from that given state and the action proposed by the policy.

\subsubsection{Conclusion}
Learning from absolute corrective demonstrations is the interactive approach most similar to standard learning from demonstration since the teacher should explicitly show what the robot has to do, i.e., she/he is required to be an expert at solving the task.
However, these interactive methods have the advantage of i) reducing the compound errors, because the demonstrations are given to correct the current learning policy deviations; ii) reducing the cognitive load of the teachers since they are not required to give full demonstrations in most of the methods, but rather occasional corrections; and iii) dealing better with the mistaken demonstrations, which are not normally considered by imitation learning methods intended for non-human teachers.

Mistaken demonstrations have a direct effect that the teacher is able to predict, allowing the teachers to be aware of how to fix their mistakes.
Although in most methods the mistaken feedback remains in the database used for training the policy, it is possible to compensate for them with correct labels outnumbering the mistakes, something relatively simple to do given the explicit nature of this kind of feedback (unlike with evaluative feedback).

\subsection{Learning from Human Relative Corrections} \label{subsub:RelativeCorrections}
\begin{figure}
\centering
\includegraphics[width=0.750\linewidth]{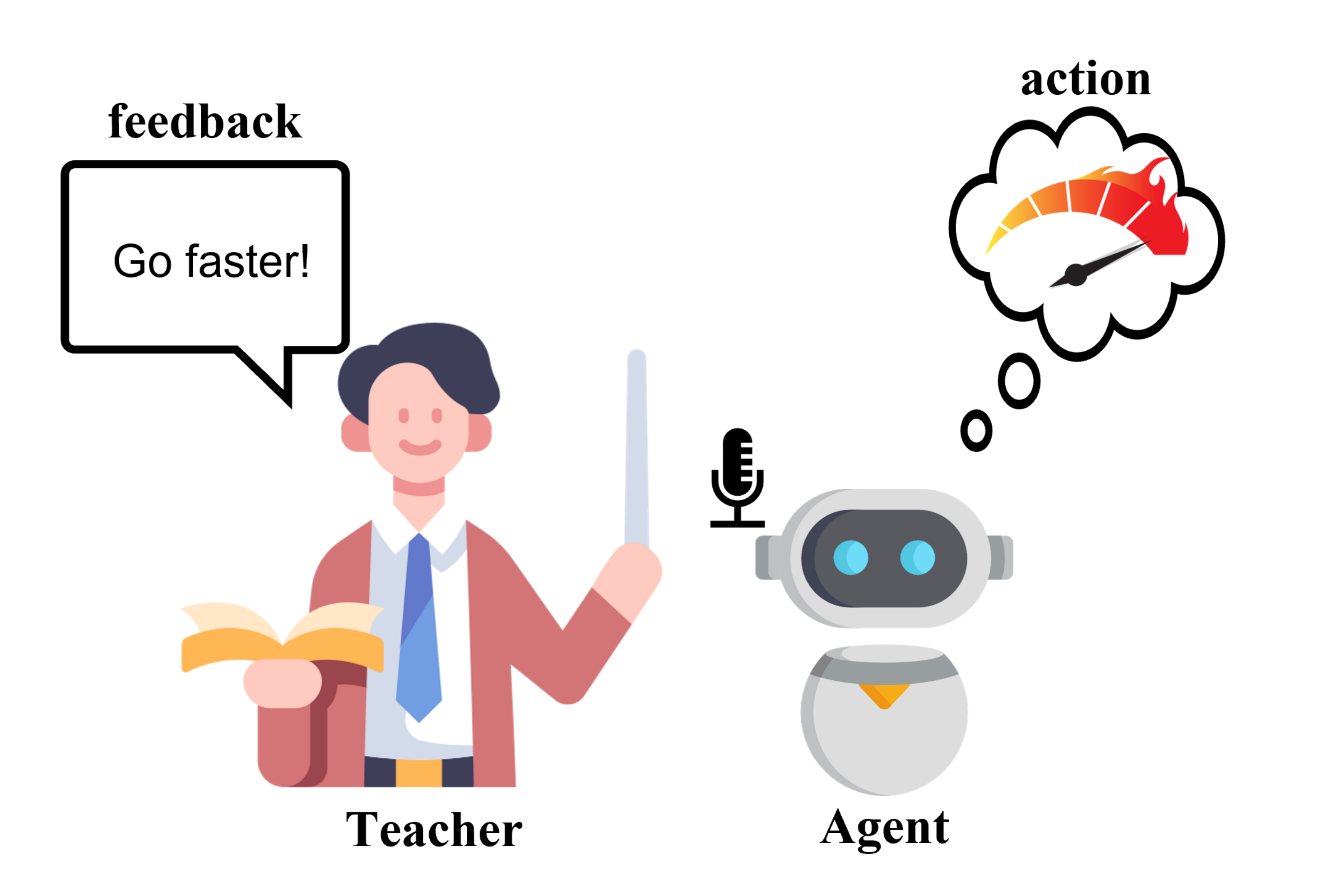}
\caption{Learning from human relative corrections: the teacher is correcting the velocity of the robot telling it that it can increase the value with respect to the current one.}
\label{fig:mod_h_relcorrections}
\end{figure}

Methods in this category do not require the teacher to know what the exact action or state transition should be applied by the agent in every state. 
However, they need to understand how a change of the action/state-transition magnitude would impact the execution of the task.
In other words, the teacher should be able to roughly estimate how a transition would change if the policy is slightly modified.
For instance, knowing that less power in a propeller decreases the acceleration of an aircraft or boat, or more force in the pedal brake slows down a car. 
With these insights, teachers could suggest how to modify a continuous policy in a more natural way (see Fig. \ref{fig:mod_h_relcorrections}), as it happens when a coach is instructing a student for learning a physical skill, e.g., in football training: \emph{kick the ball a bit harder and more to the right side}, in a singing lesson: \emph{slightly increase the volume of your voice in this part of the song}, during a dance move \emph{bend the knees less}.
This correction could be discrete (increase/decrease the action) as well as continuous-valued, depending on the interface used (see Chapter~\ref{section:interfaces})

\paragraph{Advice Operators } One of the first interactive methods using relative corrections is \gls{a-opi}, where at each iteration, it rolls out the current policy while recording the state-action pair's trajectory. Then, in an offline phase, the teacher selects the parts of the trajectory considered to be modified, along with an associated advice operator that changes the model's action of each selected pair. 
Finally, there is a phase of policy re-derivation based on the updated dataset \citep{argall2008learning,argall2009learning, argall2011teacher}.
An advice operator can be a relative change of the current action; for instance, in a navigation task, the advice \emph{accelerate} would change the model's current velocity request, multiplying it by 1.1.
It is a relative correction because it means \emph{increase} the current action magnitude. 
An advice operator can also be the demonstration of an action, being it an absolute corrective demonstration (Section \ref{subsec:AbsoluteCorrections}). For instance, the advice \emph{stop} changes the model's velocity request to zero.
Corrective demonstrations and \gls{a-opi} were sequentially applied by~\citet{mericcli2010improving} for improving the walking stability of a Nao humanoid robot.

\paragraph{The COACH framework } Similarly to \gls{a-opi}, when actions are increased/decreased, the \acrfull{coach}\footnote{This is different to the other aforementioned COACH in Section \ref{subsub:COACHevaluative}, \acrfull{coache}, which uses evaluative feedback.} \citep{celemin2019interactive} framework employs binary feedback to indicate, for a given state, the direction in which the action taken by an agent has to change, while the magnitude of the change is set as a predefined parameter in the range of the actions. 
A parametrized policy is directly learned in the parameter space, as in policy search \gls{rl} \citep{deisenroth2013survey}. 
Differently from \gls{a-opi}, the feedback provided in \gls{coach} and the policy updates occur while the agent is interacting with the environment, i.e., during policy execution time, which allows the teacher to directly observe the effects of the corrections and correct again if required, speeding up the learning process. 
\gls{coach} was originally formulated to model the policy as a linear combination of basis functions, which allowed to solve tasks such as teaching a NAO robot to dribble a ball \citep{celemin2019interactive}. 

The method was later extended to \gls{d-coach} \citep{perez2018interactive, perez2019continuous}, which models the policy with Deep \gls{nn}s, allowing to solve tasks with high dimensional observations, like RGB images obtained from a camera such as in the problem of driving a Duckie Town car \citep{perez2019continuous}. Also solving problems like balancing a real inverted pendulum swing-up, or solving a manipulation task in a conveyor belt with partial observations by incorporating memory into the \gls{nn} architecture \citep{perez2020interactive}. 
Furthermore, \gls{coach} was combined with Policy Search \gls{rl} to learn precise motor skills, solving tasks such as the ball-in-a-cup \citep{celemin2019reinforcement}. These works present experiments in which the learning agents obtained policies with higher performances with respect to the capabilities of their human teachers, who were not always able to execute the task at hand, but still managed to teach it.

\gls{coach} is employed to learn tasks with feedback in the action space, however, corrective advice can similarly be applied to collect feedback in the state space, in tasks wherein the teacher considers that it could be more natural due to the not-so-intuitive relation or effect between the action, the current state, and the next state. 
With \gls{tips} \citep{jauhri2021interactive}, relative corrections in the state space are used for updating the policy; however, in order to find the action labels that would obtain the advised relative state correction, an additional module based on learning an inverse dynamics model is proposed. 
This inverse model works for translating the state space feedback into the space of the actions, such that the policy could be updated just as with \gls{coach}.
\gls{tips} can also be considered as the interactive version of \gls{bco} \citep{torabi2018behavioral}. 
The method was validated with a \emph{fishing}  and a \emph{laser drawing} task with a real KUKA LBR iiwa 7 robot, and a user study with simulated environments showed that using feedback in the state space can reduce the task load of the teachers.
	
\paragraph{Physical advice } Some works that are more focused on teaching behaviors with manipulators have been proposed for letting the teachers provide kinesthetic corrections over the executed trajectories.
These relative corrections are used for either updating a policy or updating the objective function that can be used in a model-based setting with a planner system. 

For instance, a policy correction by the teacher on the end-effector displacement with respect to the original trajectory is detected with tactile sensors in \gls{tpc} \citep{argall2011tactile}. 
The correction could be used for policy refinement or policy reuse. In the former, the corrections are added as new data points to the training set, whereas in the latter the corrections are used to replace some already existing data points. 
In both cases, all the data points in the set are used for re-deriving the policy after the execution. 
The approach was validated with grasping tasks using an iCub humanoid robot.

Additionally, incremental refinement of trajectories of context-de\-pen\-dent policies are performed with kinesthetic feedback in \citet{ewerton2016incremental}.
The corrections are not detected and computed with tactile sensors, but rather with the measured position difference between the desired trajectory and the one disturbed by the teacher.
A reaching task is used in the experiments for testing the method with a BioRob arm.
In \citet{canal2016personalization} kinesthetic corrections are also used to reshape a movement primitive used for a feeding assistance robot application. 

\paragraph{Relative corrections as implicit preferences } The relative corrections intended to modify a manipulator trajectory are also used as implicit preference feedback, despite it being an explicit relative correction in the state space.
The trajectory disturbed by the teacher is closer to what the teacher is expecting the robot to do (preferred option) than the trajectory intended by the robot. 
Some methods leverage this information of preference for learning a function that approximates the teacher's objective (see Eq. \ref{eq:gen_objective}), such that it could be used along with a lower-level system for computing the desired robot trajectory.
Based on this concept, \gls{tpp} was proposed and tested in robotic tasks in a household setting and pick-and-place tasks in a grocery store checkout setting \citep{jain2013learning,jain2015learning}. 
Similarly, Online Learning from physical \gls{hri} was validated in household tasks with shared workspaces \citep{bajcsy2017learning, bajcsy2018learning, losey2022physical}.

\paragraph{Conclusion } The methods based on relative corrections allow non-expert teachers to incrementally correct the agent until the unknown desired actions are found, in a guidance setting that resembles the natural way a teacher corrects a student. 
Some of these methods empower the teachers, who in some cases are not able to demonstrate the task, to teach agents to perform and reach the goals successfully.
Learners outperforming the teacher in \gls{iil} is similar to what we see in humans learning complex skills, e.g., when a sports coach guides the player to perform complex behaviors that they cannot do (anymore).
Nevertheless, learning with this feedback modality is limited to continuous action problems.

Since this feedback is directly given in either the state or action space, methods using it are also more flexible for reverting the effect of mistaken corrections. 
Moreover, there are some methods that update the policies with stochastic gradient descent and do not store the feedback in a dataset, which are even less sensitive to the occasional teacher mistakes, allowing to provide a correct label that is not conflicting with any previously stored wrong feedback.  

\section{Discussion}\label{sec:ModesDiscussion}

In this section we classified different \gls{iil} methods according to the explicit information that is given by the teacher to the learner, having two main categories: Feedback in the evaluative space, and feedback in the transition space.
They are divided into subgroups of relative and absolute feedback, therefore, the discussion sets any form of interaction within one of the four subgroups: i) Human reinforcements; ii) human preferences; iii) corrective demonstrations; iv) Relative corrections.
Each of the introduced subgroups has its pros and cons which condition the situations in which they could be applied.
In general, all these interaction modalities let the teachers train agents that perform better than policies obtained with standard \gls{il}, especially to reduce the problem of compound errors, since more complete data is incrementally collected with the teacher interventions during or after the policy roll-outs.

Some works have compared methods of different modalities of interaction and have found that users tend to prefer to interact with the learning agents by communicating information that explains or shows how to perform the task, than to provide assessments of the quality of the policy \citep{thomaz2006reinforcement,toris2012practical,amershi2014power,koppol2021interaction}.
However, this preference is not the only relevant factor that could be considered for selecting the most convenient approach to solve a specific problem.
In this Chapter, we only approach that factor and leave the others for the next chapters.
The rationale for choosing a method should include the answers to questions like \emph{what kind of information is extracted from the feedback?} (Chapter \ref{sec:ModelsLearned}), \emph{is the dynamics model known?} (Chapters \ref{sec:ModelsLearned}, and \ref{sec:AuxiliaryModels}), \emph{what kind of prior knowledge is available?} (Chapters \ref{sec:ModelsLearned}, and \ref{sec:model representations}), \emph{is there access to a reward function?} (Chapters \ref{sec:reinforcement learning with HIL}), \emph{what are the independent variables or observations of the policy?} (Chapters \ref{sec:model representations}), \emph{what kind of technology is at hand for human-robot communication?} (Chapters \ref{section:interfaces}), among others.

The growing community of learning with humans in the loop research is still mostly focused on exploring new methods and evaluating their effectiveness and efficiency. 
Research for measuring and comparing usability will help to identify what approach or method is more convenient for each kind of problem (See Chapter \ref{sec: user experience}).
Usability can be assessed by analyzing how effective is a method for obtaining a successful policy, how efficient is the learning process, how pleasant the process is for the users, how sensitive it is to human mistakes, and how easy it is for the user to learn to interact with the system.
	
Nevertheless, there are insights that can guide the selection of the interaction modality to be used for training a policy.
Depending on the used modality of interaction, the set of people who can teach a learning agent can be more or less inclusive regarding their expertise.
This is correlated with the amount of information contained within the feedback signals of each modality.

With corrective demonstrations, the feedback is the informatively richest since the teacher explicitly shows what the agent should do.
This means, that only users with high expertise in the task can teach the system.
With the relative corrections, the set of users can be widened because not only expert demonstrators can participate, but also users are enabled to teach if they just have insights about how the transitions would change with a variation of the action.
They can advise slight corrections to the agent to incrementally improve a policy.
The set of possible teachers is augmented if using human reinforcements, because then, the teachers do not require to know much about what actions should be done or how transitions should be modified, as long as they can assess locally whether each part of a behavior is appropriate (assessments that implicitly happen before any intervention with the two subgroups of the transition space feedback modalities). 
If an action is considered wrong, the teacher does not need to know which the correct one is, he/she would just punish it for the agent to try something else until it finds the appropriate behavior.
And finally, in the case of learning from preferences, the set of possible teachers is the largest one, since it includes any person who understands the objective of a task, and who can assess whether one behavior is closer to the solution than other ones, without being required to understand or specify what exactly makes the preferred behavior better.

\begin{figure}
\centering
\includegraphics[width=1.0\linewidth]{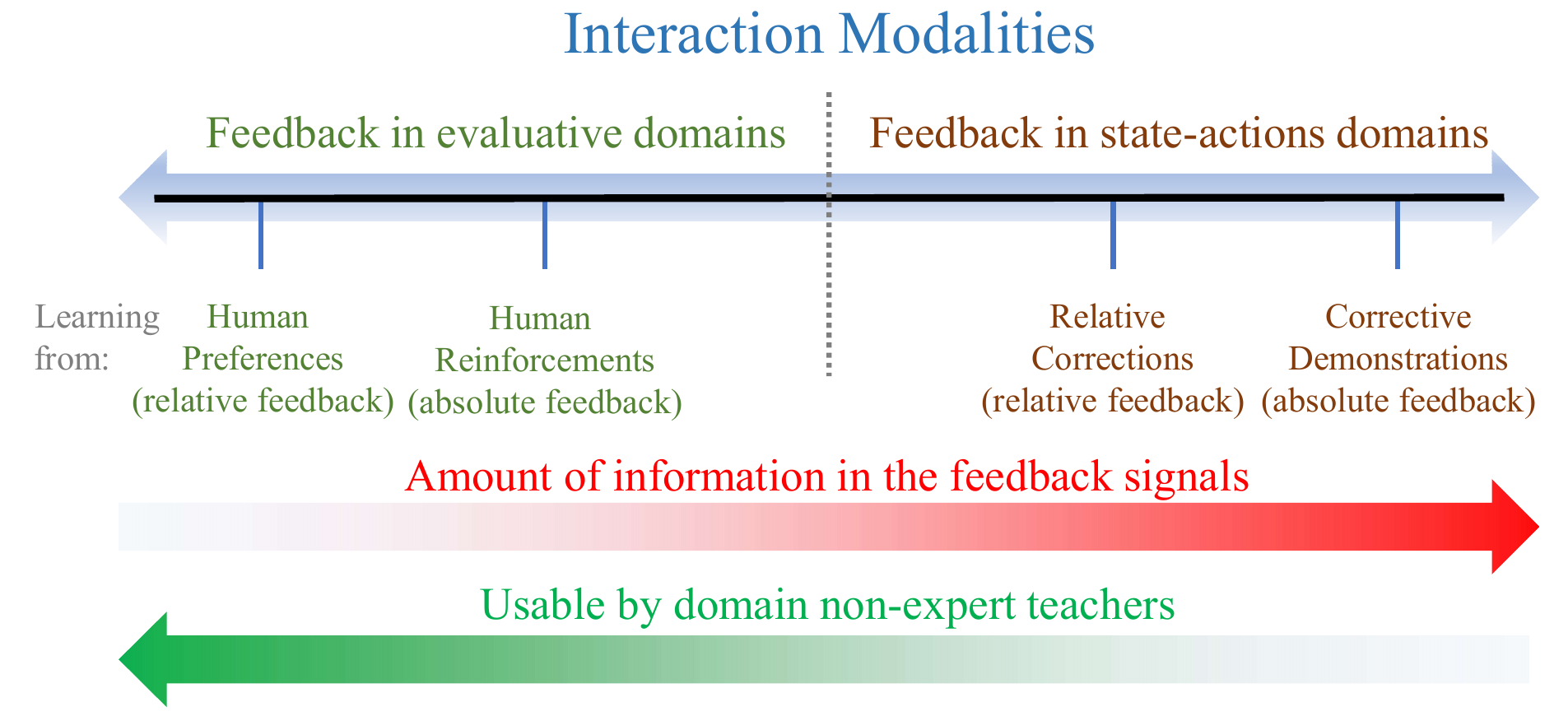}
\caption{Interaction modalities and the information contained in the feedback signals. The four modalities are organized in the plane from the right with the Corrective demonstrations modality requiring the highest expertise, to the left with the human preferences requiring the least (as shown by the green arrow).
This order has a negative correlation with the amount of information shared within the feedback signals in each modality (red arrow)}
\label{fig:modalities}
\end{figure}

As mentioned before, the corrective demonstrations are the most informative interactions, followed by the relative corrections that are defined in the same domain of actions or states, but they do not need to be strictly accurate since the accumulation of many corrections can gradually reach the desired behavior (Fig. \ref{fig:modalities}).
With human reinforcements, the feedback is a scalar evaluating the performance of each part of the policy execution, and it can be discrete or continuous.
With human preferences, the feedback contains the least amount of information because even one discrete feedback signal (or $N$ in the case of rankings) is used to compare full or partial trajectories, without assessing any individual decision.
	
Both the limitations given by human factors, or physical constraints like the ones related to learning with real physical robots that cannot be accelerated as in simulation, cannot be directly approached by adding computational power, as in the case of other \gls{ml} methods.
Therefore, some variables like the level of expertise of the teacher, the physical constraints given by the environment and the users, e.g, time, and the available interfaces compose the factors considered for selecting the right modality.
Other variables of the interactive imitation learning problem that are discussed in the next sections consider additional nuances of the approach selection. Table \ref{tab: summary chap Modalities of Interaction} provides a summary of the methods analyzed in this chapter.

\begin{table}
	\centering
	\caption{Summary of \gls{iil} methods using feedback in the evaluative and transition domain discussed in this chapter.}
	\label{tab: summary chap Modalities of Interaction}
	\begin{tabulary}{\linewidth}{|c|c|L|}
		\toprule
		\multicolumn{2}{|c|}{\textbf{Feedback}} & \textbf{List of Papers}\\
		\midrule
		\rot{19}{\textbf{Transition}} & \rot{11}{\textbf{Absolute}} & \citet{thomaz2007robot}, \citet{chernova2009interactive}, \citet{mericcli2010complementary}, \citet{mericli2011multi}, \citet{ross2011reduction}, \citet{chu2015exploring}, \citet{canal2016personalization}, \citet{chu2016learning}, \citet{fitzgerald2016situated}, \citet{laskey2016shiv}, \citet{schroecker2016directing}, \citet{zhang2016query}, \citet{krening2016learning}, \citet{laskey2017comparing}, \citet{maeda2017active}, \citet{sun2017deeply}, \citet{cheng2018fast}, \citet{fitzgerald2018human}, \citet{goecks2019efficiently}, \citet{kelly2019hg}, \citet{menda2019ensembledagger}, \citet{ablett2020fighting}, \citet{delpreto2020helping}, \citet{mandlekar2020human}, \citet{prakash2020exploring}, \citet{spencer2020learning}, \citet{hoque2021lazydagger}, \citet{bobu2022aligning}, \citet{chisari2022correct}, \citet{hoque2021thriftydagger} \\\cline{2-3}
		& \rot{8}{\textbf{Relative}} & \citet{argall2008learning}, \citet{argall2008learning}, \citet{argall2011tactile}, \citet{argall2011teacher}, \citet{mericcli2010improving}, \citet{mericcli2011task}, \citet{celemin2015coach}, \citet{jain2015learning}, \citet{ewerton2016incremental}, \citet{bajcsy2017learning}, \citet{bajcsy2018learning}, \citet{perez2018interactive}, \citet{celemin2019interactive}, \citet{celemin2019reinforcement}, \citet{perez2019continuous}, \citet{perez2020interactive}, \citet{jauhri2021interactive}, \citet{luo2021robust}, \citet{losey2022physical} \\\midrule
		\rot{17}{\textbf{Evaluative}} & \rot{11}{\textbf{Absolute}} & \citet{blumberg2002integrated}, \citet{kaplan2002robotic}, \citet{thomaz2006reinforcement}, \citet{thomaz2007asymmetric}, \citet{knox2008tamer}, \citet{mitsunaga2008adapting}, \citet{knox2009interactively}, \citet{knox2010combining}, \citet{tenorio2010dynamic}, \citet{leon2011teaching}, \citet{pilarski2011online}, \citet{suay2011effect}, \citet{knox2012learning}, \citet{knox2012reinforcement}, \citet{vien2012reinforcement}, \citet{griffith2013policy}, \citet{knox2013learning}, \citet{knox2013training}, \citet{vien2013learning}, \citet{loftin2014strategy}, \citet{macglashan2014training}, \citet{yanik2014gesture}, \citet{knox2015framing}, \citet{loftin2016learning}, \citet{krening2016learning}, \citet{macglashan2017interactive}, \citet{arakawa2018dqn}, \citet{warnell2018deep}, \citet{arumugam2019deep}, \citet{xiao2020fresh}, \citet{kahn2021land}, \citet{chisari2022correct} \\\cline{2-3}
		& \rot{6}{\textbf{Relative}} & \citet{lund1998evolutionary}, \citet{akrour2011preference}, \citet{akrour2012april}, \citet{furnkranz2012preference}, \citet{wilson2012bayesian}, \citet{jain2013learning}, \citet{akrour2014programming}, \citet{daniel2015active}, \citet{christiano2017deep}, \citet{sadigh2017active}, \citet{biyik2018batch}, \citet{ibarz2018reward}, \citet{palan2019learning}, \citet{biyik2020active}, \citet{biyik2020asking}, \citet{brown2020safe}, \citet{myers2021learning} \\		\bottomrule
	\end{tabulary}
\end{table}

\newpage
\chapter{Behavior Representations Learned from Interactions}\label{sec:ModelsLearned}


Chapter \ref{sec:modes} reviewed the different types of feedback that humans can use to transfer their knowledge about a task to robots. This knowledge is ultimately represented with a model encoding implicitly or explicitly the behavior mapping from state/observations to actions. 
To derive a behavior, different models/representations can be learned from human feedback; therefore, when solving a problem using \gls{iil}, it is not only necessary to decide which type of feedback is the most suitable for the task at hand, but also to select the representation that best fits the problem.


In this Chapter, we analyze three groups of models that have been employed in the \gls{iil} literature to solve decision-making problems by means of human feedback.  
The first type of model corresponds to the case where a policy $\pi$ is directly learned, i.e., a mapping from states to actions is obtained $\pi(s_{t})=a_{t}$.

The second group of models corresponds to state transition learning. 
In this case, the learned model $\pi^{s}$ represents a desired state transition (or its derivative) as a function of the current robot's state $\pi^{s}(s_{t})=s^{d}_{t+1}$. These methods depend on another module in charge of executing the desired state transition, e.g., a feedback controller or an inverse dynamics model $\mathcal{T}^{-1}(s_{t}, s^{d}_{t+1})=a_{t}$.

The third group consists of implicit modeling of behaviors through cost/reward/scoring or value/utility, which can be then optimized for finding the corresponding actions or transitions.

\section{Direct Policy Learning (Actions)}\label{sub:directpolicylearning}
As stated in the introduction of this chapter, \gls{dpl} stands for those methods where a policy $\pi(s_{t}) = a_{t}$ is obtained as a result of the \gls{iil} process. From the literature, it is possible to observe that methods apply \gls{dpl} by means of different feedback modalities (see Section \ref{sec:modes}), ranging from cases where desired actions are explicitly indicated to the robots, to other cases where direct policies are learned through action evaluation.

\subsection{Intuitive action spaces}
We start by considering methods that employ corrective feedback for \gls{dpl}. In such cases, it is necessary to indicate to the robot how to modify its actions to improve its behavior. Therefore, in cases where the action space of a robot is high dimensional (e.g., bimanual task) or unintuitive (e.g., underactuated hand \citep{della2018toward}), it can be very challenging for a teacher to provide corrections to the robot. Consequently, \gls{dpl} with corrective feedback is limited to those cases where the action space is intuitive for humans. An example of this is the problem of autonomous driving consisting on three actions: \emph{throttle}, \emph{steering angle} and \emph{brake}. Given that many humans know how to drive a car, it is also intuitive for them to teach a robot how to control these actions appropriately.

In these cases, absolute corrections (demonstrations) provide a well-suited framework for \gls{dpl}, since the human feedback directly indicates to the robot the desired action for a given state. Several DAgger-based methods (see Section \ref{sec:divergencemin}) have attempted to solve robotic tasks by learning policies directly. However, in many cases, they were only validated in simulated scenarios with simulated teachers \citep{prakash2020exploring,zhang2016query,menda2019ensembledagger,hoque2021lazydagger}.

As explained in Section \ref{sec:corrective_demos}, Dagger-based approaches are limited when learning from humans, as the labeling process is unintuitive and prone to errors; however, the subset of these methods based on human interventions can alleviate this issue. Consequently, methods based on human-gated \citep{kelly2019hg,goecks2019efficiently,mandlekar2020human,luo2021robust} and robot-gated \citep{hoque2021thriftydagger,laskey2016shiv,ablett2020fighting} interventions have been successfully applied to learn these policies with human teachers, solving autonomous driving and manipulation problems. It is worth noting that, although learning from interventions has been proposed in recent years as a variation of \gls{dagger} for \gls{dpl}, similar strategies were already being proposed more than a decade ago \citep{chernova2009interactive}.

\subsection{Intuitive, yet challenging action spaces}
The methods discussed above have proven to be an effective tool for \gls{dpl}; however, there are some cases where the action space of a robot might be intuitive, but challenging to demonstrate. High-frequency tasks are a good example of this (e.g., swing-up pendulum), where, although the dynamics and control inputs of the robotic system can be well-understood by the teacher, it can still be challenging for them to successfully control the robot and provide absolute corrections.

In such cases, relative corrections are a suitable alternative for \gls{dpl}, since the teacher only needs to indicate the direction where the action taken by the robot must be modified. This allows to gradually improve the behavior of the learner even when it is not possible to demonstrate the task. For instance,  \citet{celemin2019interactive} train an agent to balance a bicycle and solve the cart-pole problem through relative corrections, where it is shown that, even though the human is not able to teleoperate both tasks, it is capable of successfully teaching an agent to solve them. Moreover, one of the deep learning extensions of this method \citep{perez2020interactive} was able to successfully teach an agent to swing up a pendulum from raw pixels of an image, showing superior performance of relative corrections in this task when compared to the intervention-based method \gls{hg-dagger} \citep{kelly2019hg}.

\subsection{Non-intuitive action spaces}
Some approaches have addressed the problem of \gls{dpl} in scenarios where the action space can be difficult for the teacher to understand. For instance, giving corrections in the joint space of a robot arm corresponds to one of these cases. Providing corrections in this space can be very challenging, since, commonly, manipulators must solve tasks by controlling their end-effector. Hence, the teacher needs to have an accurate internal model of the effect that actions in joint space generate in the robot's end-effector to teach the robot how to solve the task at hand (Section \ref{sec:LfO} provides a deeper study of these cases). To address this problem, \citet{jauhri2021interactive} introduce a method for giving relative corrections in the state space of the environment (e.g., feedback about the effects of the manipulator's end-effector) and use an inverse transition model (learned from interactions with the environment) to map this feedback to the action space (joint level in the example) of the robot, allowing it to directly learn a policy in this space.

Finally, evaluative feedback can be employed to learn policies directly when the action space is not intuitive for the teacher. Evaluative feedback allows the human to teach a policy without having to understand the action space of the robot (see Section \ref{sub:EvaluativeDomain}). In most cases, evaluative feedback is employed to learn a value function. Nevertheless, \citet{macglashan2017interactive} propose to directly learn a policy through evaluative feedback by introducing an \gls{iil} method inspired by policy gradients from \gls{rl} \citep{sutton2018reinforcement}. The deep learning extension of this method \citep{arumugam2019deep} showcases experiments where an agent learns to solve simulated navigation tasks from raw pixels of an image.

As a final remark, note that methods that work in non-intuitive action spaces can also be employed to learn policies in intuitive action spaces.
\section{Learning Desired State Transition/Dynamics }\label{sec:LfO}
\gls{dstl} stands for those methods where a policy indicates the next desired state that the agent should visit $\pi^{s}(s_{t})=s^{d}_{t+1}$.
As highlighted in Section \ref{sub:directpolicylearning}, when it is not intuitive to provide feedback in the action space of the agent, solutions have been proposed for converting human feedback in state space to the desired action space.
For example, in a torque-controlled robot, the demonstrator's action sequence is not available/observable when learning from kinesthetic teaching \citep{celemin2019reinforcement}.
This is also true when humans want to imitate each other's behavior: they do not have access to the internal actions used by the demonstrator.
Instead, they can only observe the state transitions generated by those demonstrations; therefore, they need to infer the necessary actions to achieve the same transitions. 
Additionally, when dealing with robots with many \glspl{dof}, the user aims to teach behaviors in task space, and not in the actuator level of the desired task to perform.
For example, in a manipulation insertion task, the demonstrator would only focus on the end-effector position and not on the complete kinematic chain state of the robot. 
Moreover, teaching and correcting movements in task space also allow better generalization \citep{ewerton2016incremental, meszaros2022learning}.

\gls{dstl} relates to the literature of \emph{model-based} methods in that it makes use of learned/known models of the environment to achieve desired state transitions. 
The \gls{iil} community aims to leverage those models to make the teaching and the correction easier and accessible to non-expert users, who may not be familiar with the effect of actions on the operational space dynamics \citep{argall2011teacher}. 
For example, manipulation tasks can be solved by interactively learning the desired end-effector transitions. Those transitions can then be achieved by the robot using a feedback controller, such as cartesian impedance \citep{franzese2021ilosa} or velocity \citep{chisari2022correct} controllers.


\paragraph{State dependent vs State Independent transitions}
The desired state transitions can be described as a function of the current state or with a state-independent formulation; for example, as a function of time/phase using movement primitives like \gls{dmp} or \gls{pro mp} or as a function of the current state space using a \gls{gp} \citep{franzese2021ilosa} or a \gls{nn} \citep{chisari2022correct} (introduced in Section \ref{sec:model representations}). 
When learning a state-dependent transition the policy can be described as $\pi^{s}(s_{t})=s^{d}_{t+1}$ that still fits the \gls{mdp} formulation. 
On the other hand, when learning a state-independent movement primitive, the robot learns a behavior model that describes how the state evolves over time, i.e. $\pi^{t}(t)=s^{d}_{t+1}$, which can again be tracked using feedback control.
\citet{argall2011teacher} use human feedback to shape the desired trajectory of a robot starting from the assembly of primitives and using local correction refinement of the final driving policy execution. 
\citet{celemin2019reinforcement} combine interactive learning with \gls{rl} to learn a \gls{dmp}. The interactive correction on the current robot state is used as exploration for a faster convergence to optimal performance of the task. 
Similarly, \citet{schroecker2016directing} use \gls{iil} to stop the \gls{rl} algorithm, and they let the user bring the robot to the desired state at the particular time of the execution. This strategy is used to create soft via-point constraints that limit the \gls{rl} exploration and results in a faster learning convergence of a \gls{dmp}.
An active and interactive collection of state space demonstrations is proposed by \citet{maeda2017active} where multiple trajectories are collected from demonstration, and saved in a model that estimates the movement parameters as a function of the current context of the task. When the robot faces a novel situation, it actively queries the user to provide more data. 

As a final remark, the difference between desired state transition and action strongly depends on the definition of the \gls{mdp}: if the action is set to be equal to the desired robot transition, the two categorizations collapse in a single one. However, when working with real robots, the re-formulation of the \gls{mdp} needs to be supported by available models (learned or hard-coded) that handle the conversion from the newly defined action and the actual robot control. This section summarized how interactive methods took advantage of these models to obtain more efficient and user-friendly teaching of robotics tasks.

\section{Learning Reward and Objective Functions} \label{sec:learning_reward}

Contrary to learning a policy from human feedback directly, there exists a plethora of methods that fit a reward or objective function first. Such a function can then be used to either derive a policy~\citep{christiano2017deep} or directly infer the optimal action~\citep{knox2008tamer}. In the next subsections, we describe and distinguish between methods that learn a reward, cost or scoring function, and methods that learn an objective, such as value or utility functions. 

\subsection{Learning Reward, Cost, or Scoring Functions}

A common approach studied in the literature is to use human feedback to shape a \emph{reward} function, which can then be used to train a policy.
For example, in the method proposed by~\citet{christiano2017deep}, the reward function is estimated by minimizing a cross-entropy loss where the labels are the human-provided pairwise preferences. 
An alternative approach is taken by~\citet{sadigh2017active}, where the weights of a reward function are learned from human preferences via a Bayesian update.
An extension of these ideas consists of learning a reward function by integrating human preferences with expert demonstrations.
For example, in \gls{dempref}~\citep{palan2019learning, biyik2020asking}, the weights of the reward function are pretrained with the expert demonstrations via a Bayesian \gls{irl} approach, while \citet{ibarz2018reward} use the demonstrations to pretrain the policy, which is then used to generate a set of initial trajectories for the human teacher to annotate with preferences, from which the reward function is finally learned. 

While the aforementioned methods aim to learn a reward function via preference feedback, there have been works in the literature proposing to learn approximate reward functions from corrective feedback. In the works of~\citet{bajcsy2018learning, losey2022physical}, the robot is trained through physical human-robot interaction, and the weights of the reward function are updated in the direction of the \gls{map} estimate of the observed and corrected trajectories.
A different approach is taken in \gls{land}~\citep{kahn2021land}, where human intervention is used to learn a disengagement prediction model to distinguish good and bad actions. This model is then included in the \emph{cost} function used to guide a model predictive control planner.

A different solution is \gls{tpp}~\citep{jain2013learning, jain2015learning}, which consists of learning a \emph{scoring function} used to rank different trajectories. At each episode, a planner is used to sample multiple trajectories, which are ranked according to the current scoring function. If the top scoring trajectory is not considered adequate, the human teacher provides a preferred alternative, which is used to update the weights of the scoring function by gradient-free optimization. At inference time, the scoring function is used to pick the best trajectory from the ones sampled by the planner.

\subsection{Learning Value or Utility Functions}
Instead of learning a reward function, an alternative approach consists of directly training from human feedback a \emph{Q-value} function, which describes the expected future cumulative reward of each state-action pair~\citep{watkins1992q}.
As shown by~\citet{thomaz2006adding, thomaz2006reinforcement, suay2011effect}, this is achieved by substituting the environment-provided reward (which is not always readily available) with the human-provided evaluative feedback.
In these works the action space is discrete; hence, the agent can infer the action by directly maximising the Q-function by selecting the action with the highest value: \( a = \argmax_{\hat{a}} Q(s, \hat{a}) \).
A similar approach is taken in the \gls{tamer} framework~\citep{knox2009interactively}, discussed in Section~\ref{sec:tamer}, where the human reinforcement function \( H(s, a) \) is trained from human feedback and then used directly for inference: \( a = \argmax_{\hat{a}} H(s, \hat{a}) \).
Directly maximizing the value function at inference time is an effective approach, but it is only straightforward for discrete action spaces. 

A possible solution to cope with continuous action spaces is to use the learned objective function for learning an explicit model of a policy. \gls{ppl}~\citep{akrour2011preference, akrour2012april, akrour2014programming} uses the preference feedback from the teacher to learn a policy return estimate, which is then used to build new candidate policies. The policy return estimate, also referred to as \emph{utility function}, represents the quality of a given state with respect to discovering new better policies. New policies are then generated following either an evolution strategy approach or an \gls{aeus} criterion, and the process is repeated.

\section{Discussion}\label{sub:modelslearned-discussion}
In this chapter, we classified methods according to the type of model learned from human feedback. 
Different approaches exist to teach robots to solve tasks through human-robot interactions. 
Some methods directly learn a mapping from states to desired action/states and others indirectly derive policies by learning functions that evaluate the performance of the agent's actions.

We observed that depending on the task at hand, and how intuitive it is for a human to provide a specific type of feedback, different alternatives exist to address it. Therefore, although one type of feedback can be ideal for a specific use case (e.g., interventions for driving), it might not be a satisfactory candidate to solve other problems (e.g., interventions for balancing tasks).

Furthermore, we also analyzed that the learning process can be assisted by incorporating prior knowledge about robotic platforms in \gls{iil} methods. For instance, it is possible to directly learn manipulation tasks in the robot's end effector space by using the robot's geometry to map these behaviors to the actuation space. More details regarding auxiliary models are presented in Chapter \ref{sec:AuxiliaryModels}.

Finally, we observed that it is possible to employ feedback signals, that require a low cognitive load for human teachers, to implicitly learn behaviors using reward or objective functions. These behaviors can then be decoded using optimization methods such as model predictive control. 

So far, we studied the feedback modalities that humans can employ to transfer their knowledge to robots (Chapter \ref{sec:modes}) and, in this chapter, the models that can be learned from them (summarized in Table \ref{tab:table_behavior}). 
We observed that depending on the model at hand, different types of feedback can be used to learn it.
However, an important part of model learning is to select an appropriate function approximator to represent it, which was not addressed in this chapter. 
This is reviewed in depth in Chapter \ref{sec:model representations}.

\begin{table}
\centering	\caption{Summary of \gls{iil} methods discussed in this chapter.}
	\label{tab:table_behavior}
	\begin{tabulary}{\linewidth}{|Z{0.8in}|L|}
		\toprule
		\textbf{Behavior Representation} & \textbf{List of Papers}\\
		\midrule
		\textbf{Direct Policy} & \citet{chernova2009interactive}, \citet{laskey2016shiv}, \citet{zhang2016query}, \citet{macglashan2017interactive}, \citet{arumugam2019deep}, \citet{celemin2019interactive}, \citet{goecks2019efficiently}, \citet{kelly2019hg}, \citet{menda2019ensembledagger}, \citet{ablett2020fighting}, \citet{mandlekar2020human}, \citet{perez2020interactive}, \citet{prakash2020exploring}, \citet{hoque2021lazydagger}, \citet{jauhri2021interactive}, \citet{luo2021robust} \\\hline
		\textbf{Desired State Transition} & \citet{argall2011teacher}, \citet{ewerton2016incremental}, \citet{schroecker2016directing}, \citet{maeda2017active}, \citet{celemin2019reinforcement}, \citet{franzese2021ilosa}, \citet{chisari2022correct}, \citet{meszaros2022learning} \\\hline
		\textbf{Reward/{\allowbreak}Cost/{\allowbreak}Scoring} & \citet{akrour2011preference}, \citet{akrour2012april}, \citet{jain2013learning}, \citet{akrour2014programming}, \citet{jain2015learning}, \citet{christiano2017deep}, \citet{sadigh2017active}, \citet{bajcsy2018learning}, \citet{ibarz2018reward}, \citet{palan2019learning}, \citet{biyik2020asking}, \citet{kahn2021land}, \citet{losey2022physical} \\\hline 
		\textbf{Value/{\allowbreak}Utility} & \citet{thomaz2006reinforcement}, \citet{knox2008tamer}, \citet{knox2009interactively}, \citet{suay2011effect}, \citet{christiano2017deep} \\
  \bottomrule
	\end{tabulary}
\end{table}
\newpage
\chapter{Auxiliary Models}\label{sec:AuxiliaryModels}

\begin{figure}[h]
    \centering
    \includegraphics[width=0.7 \linewidth]{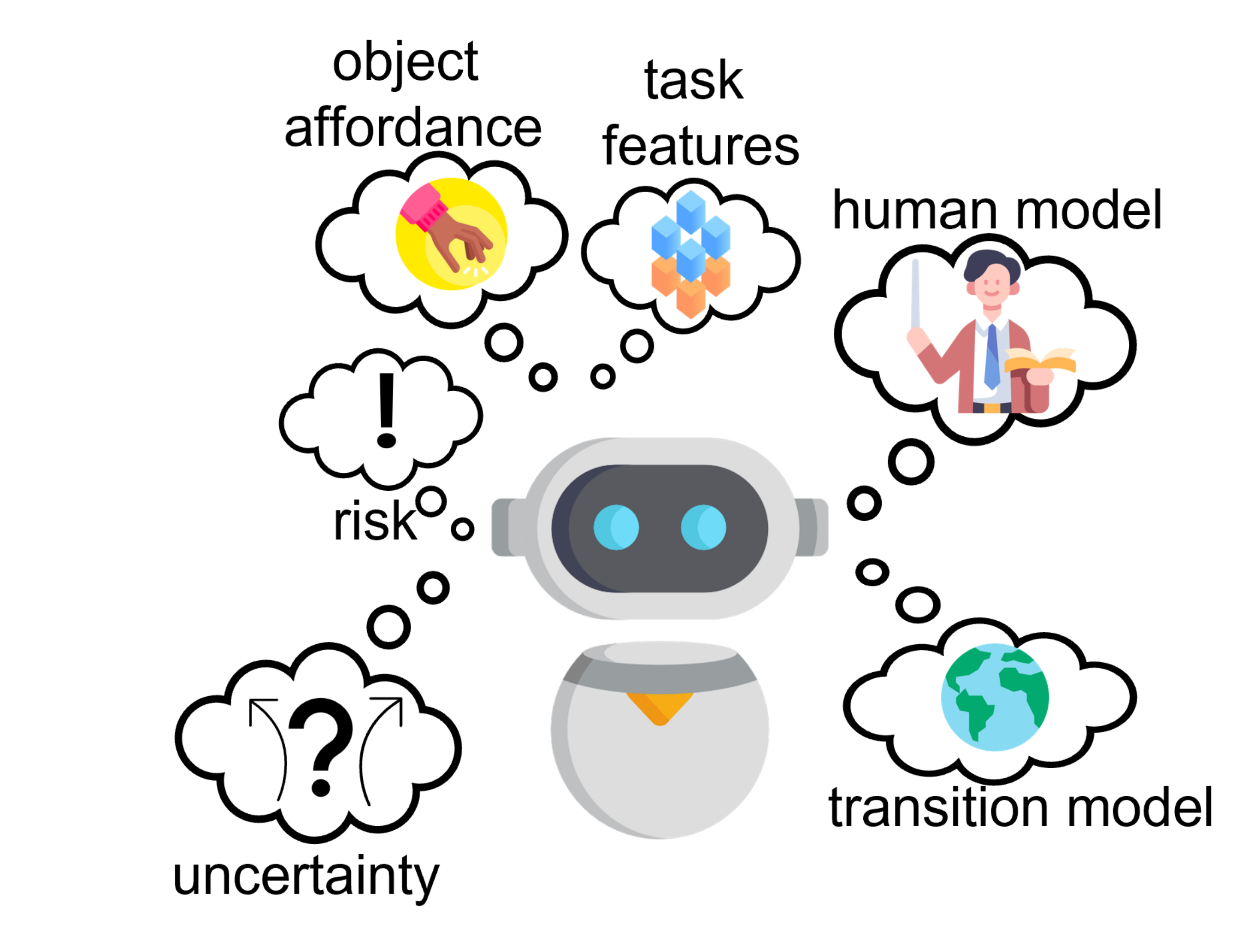}
    \caption{Auxiliary models that are commonly used in interactively imitation learning. }
    \label{fig:auxiliary}
\end{figure}

When teaching autonomous agents interactively, the quality of the human feedback directly affects the learned behavior. Chapter~\ref{sec:ModelsLearned} discussed different types of models/representations that can be used to encode the desired behavior. Besides these, it is often useful to employ additional models to enhance the training process and/or the execution of the task. In this chapter, we consider multiple types of auxiliary models (see Fig. \ref{fig:auxiliary}) that are used to improve the learning process, the teacher experience, and the policy execution with respect to different objectives such as data efficiency, safety, and credit assignment. These models can be either learned or hand-crafted and are applied either at training time or at inference time.

\section{Task Features Learning}\label{sec: task features}

Learning features for a task allows identifying a more compact and/or descriptive state space, and it can tremendously boost learning speed and generalization.
As such, learning the importance of features, or the features themselves can be a key component on \gls{iil}.
This is especially important within the \gls{iil} scope, given the necessity of having a reduced number of demonstrations and corrections and still achieving desirable performance.
Furthermore, hand-designed task features are undesirable as they might not generalize or transfer to different tasks and they require task-specific engineering effort.

Task feature learning has been applied in the \gls{iil} context in the past years, aiming for identifying features for improving the learning performance, enabling teaching with high-dimensional signals, or aligning the agent's and the human's knowledge \citep{bobu2022aligning}.

\paragraph{Dimensionality reduction}

Dimensionality reduction approaches target finding a compact set of structured and informative features from high-dimensional signals.
Such approaches aim to enable \gls{iil} methods to work on high-dimensional observations without requiring large amounts of data as is the case with most deep learning methods.
Auto-encoders have been successfully applied in order to reduce the high-dimensional state spaces by learning a smaller set of informative features.
They have been used in \gls{d-tamer} \citep{warnell2018deep}, in \gls{d-coach} \citep{perez2018interactive}, and by \citet{luo2021robust}, however, in those works, the auto-encoders are pre-trained, instead of being interactively learned from human input.

In this scope, \citet{perez2019continuous} extend \gls{d-coach} by making both the auto-encoder and policy to be learned online.
Results show data efficiency improvement w.r.t.\ the previous version of \gls{d-coach}, which is due to the fact that both the auto-encoder and the learner agent are trained using the same distribution, i.e., the states encountered during training, while auto-encoders trained offline are likely to encode only a set of all the possible observations the learner can obtain.

\citet{perez2020interactive} extend these architectures to enable \gls{iil} from camera images in environments that are non-fully observable.
A \gls{lstm} model is used to introduce recursion in the learned \gls{nn} for dealing with partially-observable processes.
The approach is applied to a fruit selection task, in which fruits approach the robot through a conveyor belt and a camera captures a region of the belt different from the robot's action region, making it necessary to learn to predict where the fruits will be and when they will arrive at the robot's reach.


\paragraph{Interactive Feature Learning for Inverse Reinforcement Learning}

Relevant task features can also be learned interactively within a \gls{irl} setting. \citet{bobu2021feature,bobu2022inducing} present \gls{ferl}, a framework that interactively learns arbitrarily complex non-linear features, such as distance to and between objects. Then, these features are used to obtain a policy via \gls{irl}.
At the beginning of the first step, the robot does not have prior knowledge about the features, and the user provides information about them through the so-called \emph{feature traces}. This feedback consists of trajectories obtained from the user, who guides the robot from states where the features are strongly activated to states where they are not, according to the teacher's judgment.
In the second step, the learned features are combined using standard reward learning frameworks, which are used to infer the policy.
Results show that the learned features improve generalization capabilities w.r.t. non-interactive \gls{irl}-based approaches \citep{finn2016guided,wulfmeier2016watch}, which simultaneously learn task features and reward.

\section{Object Affordances}
\label{sec:objectaffordance}
Many real-world robotic applications require finding a sequence of actions applied to relevant objects to bring the environment to an intended state. To be effective, sometimes it is useful to design appropriate abstractions of the environment, to focus only on the actions that generate the intended (meaningful) effect on the objects. This can be achieved via affordance modeling.
An affordance model accounts for a high-level behavior modeling approach that learns the relationships between the robot, its actions, and their effect on objects \citep{gibson1977theory}.
For a given object, an affordance $\AFFORD_\INTENT$ represents a subset of the state-action space ($\AFFORD_\INTENT \subset \STATES \times \ACTIONS$), that leads to the intended effect $\INTENT$ (mapping from states to distribution over states, denoting high-level effect, e.g. \emph{grasped}) \citep{khetarpal2020can}. There are different approaches to discovering meaningful affordances as described later in this section. Once discovered, affordances might offer a kind of generalization across different objects of the same class. High-level decision-making can utilize then affordances to achieve desired goals/transitions in the environment efficiently and effectively.

\paragraph{Learning Object Affordance from Vision}
Vision has been used since the beginning of affordance learning in the literature.
\citet{thomaz2007robot, thomaz2009learning} address human-guided exploration for learning affordances.
In this approach, the robot learns object affordances through a combination of self-exploration and human guidance. The teacher guides the robot by i) providing evaluative feedback, ii) suggesting to perform certain actions (e.g., \emph{try action X on the object Y}), iii) drawing attention to an action-effect observation (e.g., \emph{look it's Z}) and iv) controlling the environment so that the appropriate cues are most noticeable for making the learning process efficient (placing an object in areas or poses which increased the likelihood of finding affordances).

For instance, \citet{thomaz2009learning} experiment with a robot configured as an upper torso humanoid. It learns about a set of five objects with different geometrical shapes and colors. The robot learns different affordances, such as \emph{lift-able}, \emph{open-able}, \emph{roll-able}, \emph{move-able} and \emph{tip-able}. The results from these experiments show that a non-expert user is able to teach the robot to learn the affordances more successfully than when the robot learns by itself.

Although human guidance results in efficient exploration strategies, it is cumbersome to have humans provide an exhaustive set of interactions for each affordance. Therefore, to reduce the burden on humans, \citet{chu2015exploring} introduce the \emph{human seeded exploration} strategy, which is a combination of self-exploration and human-guided exploration. In this strategy, first, a human teaches the affordances using kinesthetic demonstrations. The robot uses the distribution of the demonstrations for searching in the action space and finding affordances in new situations. The experiments were conducted with a robot having 7 \gls{dof} arms. This approach results in improved success rates with fewer object interactions for learning affordances.

\paragraph{Learning Object Affordance from Haptic Feedback}
The previously mentioned approaches consider learning affordance models using visual information. In contrast, \citet{chu2016learning} propose a complementary haptics affordance model, which is fused with visual affordance to develop a multi-modal model. It characterizes how a particular action-object pair feels and, therefore, can aid in better task completion. During each interaction, the human teaches the robot via \emph{environmental scaffolding}, i.e., moving objects slightly to perturb the action context.

\paragraph{Learning Object Mapping Using Affordance}
\citet{fitzgerald2016situated,fitzgerald2018human} develop an \gls{iil} method to enable agents to map representations of objects between environments based on their affordances. These models are used to transfer a learned task into a new environment. For example, a glass (\emph{fill-able}, \emph{drink-able}, \emph{pour-able}) in an environment $E_a$ can be mapped to a pitcher (\emph{fill-able}, \emph{pour-able}) in another environment $E_b$, for instance, on the basis that both objects are \emph{pour-able} if the task requires pouring. However, these objects could not be mapped together if the task requires drinking since the pitcher is not drinkable.
The human teacher assists the agent by indicating the correct mapping for some of the objects. Through this assistance, the algorithm infers the mapping function between source and target objects, such that the remaining objects can be mapped autonomously for task completion. The results show that the agent can use human guidance to quickly infer a correct object mapping, requiring assistance with only a few steps.

\paragraph{Learning Affordance from Natural Language Advice}
In order to learn a task from users with non-\gls{ml} expertise, an \gls{iil} method should be able to understand simple human explanations. Most approaches require the users to provide state-specific advice, which might not always be intuitive for them. Alternatively, it is possible to provide object-specific advice through natural language \citep{krening2016learning}. For example, consider that a human wants to teach an agent to play a Super Mario game. It is more intuitive for the human to advise which actions to take with respect to an object (e.g., \emph{jump on an enemy}), rather than state-specific advice such as \emph{hold the jump key for 10 frames when Mario is within 2.5 horizontal blocks of an enemy with a velocity of 3.2 units/frame}. Hence, object-focused feedback helps generalize over the state space.

\section{Forward and Inverse Transition Models} \label{sec:transition_models}

Transition models can be useful for training autonomous agents as they encode information about how the state evolves given a certain action, and can potentially result in better sample efficiency and improved generalization capabilities.

A useful application of a transition model consists of the search for an optimal trajectory given a cost function. In the setting of interactive learning, \citet{losey2022physical} use the optimization-based motion planner TrajOpt~\citep{schulman2014motion} to find the best new trajectory according to a reward function shaped by human corrections. Similarly, various methods that learn a reward function by preference feedback make use of the dynamics model of the system to optimize the output trajectory~\citep{sadigh2017active, palan2019learning}.

Another use of transition models is trajectory generation for sampling-based approaches.
In \gls{tpp}, \citet{jain2015learning} employ a model of the robot kinematics and of the obstacles in the environment within a sampling-based planner to generate collision-free trajectories. The learned \emph{scoring function} obtained from the human feedback is used to pick the optimal path.
A further sampling-based approach is employed in \gls{land}~\citep{kahn2021land}, where at each step multiple roll-outs are generated, and the first action of the best trajectory is applied to the agent in a model predictive control fashion. The cost function for the selection includes a model, trained from human feedback, which outputs a sequence of disengagement probabilities, i.e. the probability that the robot needs to be stopped by the human teacher because it moves towards an unsafe region of the state space.

In the field of \gls{iil}, another application of transition models is to facilitate the teaching process, making it more intuitive for non-expert demonstrators that are not familiar with the effect of controlling robot actions.
The \gls{tips} framework of~\citet{jauhri2021interactive} proposes to learn a forward dynamics model which maps a state and action pair \( (s_t, a_t) \) to the next state \( s_{t+1} \). This forward model is continuously updated from the recorded robot transitions during the task's episodes and used to allow the user to give feedback on the desired state dynamics rather than the desired action. This approach is advantageous for all applications where providing feedback in the action space is difficult or unintuitive.
\section{Confidence, Novelty and Risk Models}\label{sec:confidence}
When learning from non-expert users, it is desirable to minimize the amount of feedback the teacher needs to provide. At the same time, the robot should not attempt dangerous actions in regions where the learned policy is not well trained. To achieve these objectives, different researchers use confidence, novelty and risk detection models to support the user during the teaching. 

\subsection{Novelty and Confidence model}
The use of Bayesian methods (and their approximations), through the estimation of epistemic uncertainty, provides a successful tool for performing safe learning, i.e., calling attention when in a region of high uncertainty or asking explicitly for more data (active learning). 
\citet{chernova2009interactive} propose a confidence-based interactive method where the policy outputs the action as well as its confidence. When the learner executes the action, it also checks its confidence score against a threshold. If below, the user is asked to intervene.
Similarly, \citet{franzese2021ilosa} encode the desired robot motion from demonstrations and interventions, and use a measure of confidence to constantly direct the robot to regions of minimum uncertainty (high confidence).

To enhance exploration when learning from interaction, \citet{kulak2021active} adopt a measure of confidence  
to explicitly bring the robot to regions of high uncertainty and actively ask for feedback to have a heterogeneous set of demonstrations to learn from.
\citet{menda2019ensembledagger} and \citet{kelly2019hg} propose to derive the action and confidence estimate as the mean and variance of the predictions of an ensemble of \glspl{nn}. The threshold over which the user's help is queried is tuned with previously observed interventions of the user. 
\citet{subramanian2016exploration} estimate a value function and use two statistical measures (leverage and discrepancy) to compute the influence of visited states on the value estimation. When it is high, the teacher is actively requested to perform a demonstration to move the robot to that state to increase exploration. 
In the context of Bayesian \gls{irl}, \citet{cui2018active} propose an active and interactive framework. The agent queries the user for segmenting and labeling (segment of) trajectories as good or bad. Given the probabilistic formulation of the reward function, the robot can generate trajectories that maximize information gained after the actual labeling from the user. Finally, the user can label different steps as good or bad, and this label is then used to update the reward function using a soft-max update rule. 

\subsection{Risk Detection and Safety Enhancement}
While very important, the novelty of the observation is not the only factor that leads to a high probability of failure. A state can be risky even though not novel.
In the context of \gls{irl}, to generate risk-aware performance-based trajectories, \citet{brown2018risk} examine the robot’s policy and evaluate per-state policy loss. The states that have high loss/low cumulative reward are classified as risky states.  
Starting from these critical states, the algorithm samples trajectories and asks the human supervisor to critique them. With human feedback, the policy is updated to reduce the risk of failing when approaching those states. 

\citet{zhang2016query} propose \gls{safe dagger}, a safety rule to enhance \gls{dagger}, where during the data collection, not only the action policy is learned, but also a safety policy that returns a binary classification value, i.e. safe/unsafe, given the state-action pair from the primary policy.
\gls{lazy dagger}~\citep{hoque2021lazydagger} extends this simple rule with a hysteresis model to reduce the amount of switching between the supervisor and the policy.
Similarly, \citet{ablett2020fighting} propose to fit a discriminator that is able to classify state-action pairs as dangerous and assign them a probability of failure. 
In order to avoid querying the user too often, an additional parameter is learned from interaction, which evaluates if the current discrimination is too dependent (high user burden) or too independent (leads to failure), and updates it accordingly. 
Finally, \citet{hoque2021thriftydagger} propose to learn a Q-function to estimate the probability of failure from a certain state, based on past experience. This measure is used to request the user to intervene.

\section{Human Models for Feedback Interpretation} \label{sec:feedback_interpretation}
When receiving feedback from humans, it is important to take into account the delayed reaction and the decaying significance of past state-action pairs. 
The authors of \gls{tamer} \citep{knox2009interactively} propose a `\emph{Credit Assigner} module, intended for environments of \emph{high frequency} regarding the human response capabilities. 
The module aims to solve a temporal credit assignment problem. A human trainer is not able to assess the effect of each action at each time step, so this produces a delay between the action execution and the human response.
The Credit Assigner proposed in \gls{tamer} approaches this problem by associating the feedback not only to the last state-action pair but to a past window of pairs. Each pair is weighted with the corresponding probability computed with a model of the human delay probability density function \citep{knox2009interactively}. 
While being determined experimentally, this model of the human delay appears to correspond to the one found empirically in psychological studies \citep{sridharan2011augmented}.

A similar idea is presented by~\citet{loftin2016learning} with the \gls{isabl} framework. They propose a probabilistic model of the human teacher feedback which describes how a trainer decides to provide an explicit reward or explicit punishment. The model takes into account the overall teaching strategy, and it is also able to learn from the actions for which the human does not provide any feedback.
Finally, \citet{celemin2015coach} propose to employ a model of the human teacher to predict what feedback the agent will receive on a given state. This model can then be used to adapt the step size of the relative corrections from the teacher, being able to incorporate the intentions of the teacher encoded on the past feedback, i.e., for a specific state, either applying a large change to the policy or fine-tuning it.

\section{Discussion}
In this Chapter, different auxiliary models are discussed that contribute to improving the overall interactive learning process. Task features learning addresses the problem of obtaining a good representation of the environment that a policy can use to learn data efficiently and generalize well. Object affordances are useful for finding appropriate abstractions of the objects to manipulate and to focus the learning process on the effects that certain actions generate. Transition models are helpful, as they describe how the state evolves given a certain action, and are often employed for trajectory generation, either via sampling or optimization. Confidence, novelty and risk model are employed to assess the safety of the agent behavior in a given state, and to react accordingly if needed, e.g. by querying the human teacher. Finally, models of the behavior of the human teacher can also be used. They aim to improve different aspects of the learning process such as credit assignment or adaptive step size in the feedback interpretation. 
Hence, each type of auxiliary model has its advantages, and the choice of adopting one over the other mainly depends on the task at hand: task features can be useful for problems with otherwise high-dimensional representation, object affordances are mainly tailored for manipulation tasks, transition models are needed for scenarios where planning is required, while confidence and risk models are helpful for safety-critical application. Moreover, such models are not mutually exclusive, and a combination of multiple of them is possible.
In the next section, the existing types of representation or function approximation that are commonly used in the \gls{iil} setting are discussed.


\newpage
\chapter{Model Representations (Function Approximation)} \label{sec:model representations}
\begin{figure}
    \centering
    \includegraphics[width=0.8\linewidth]{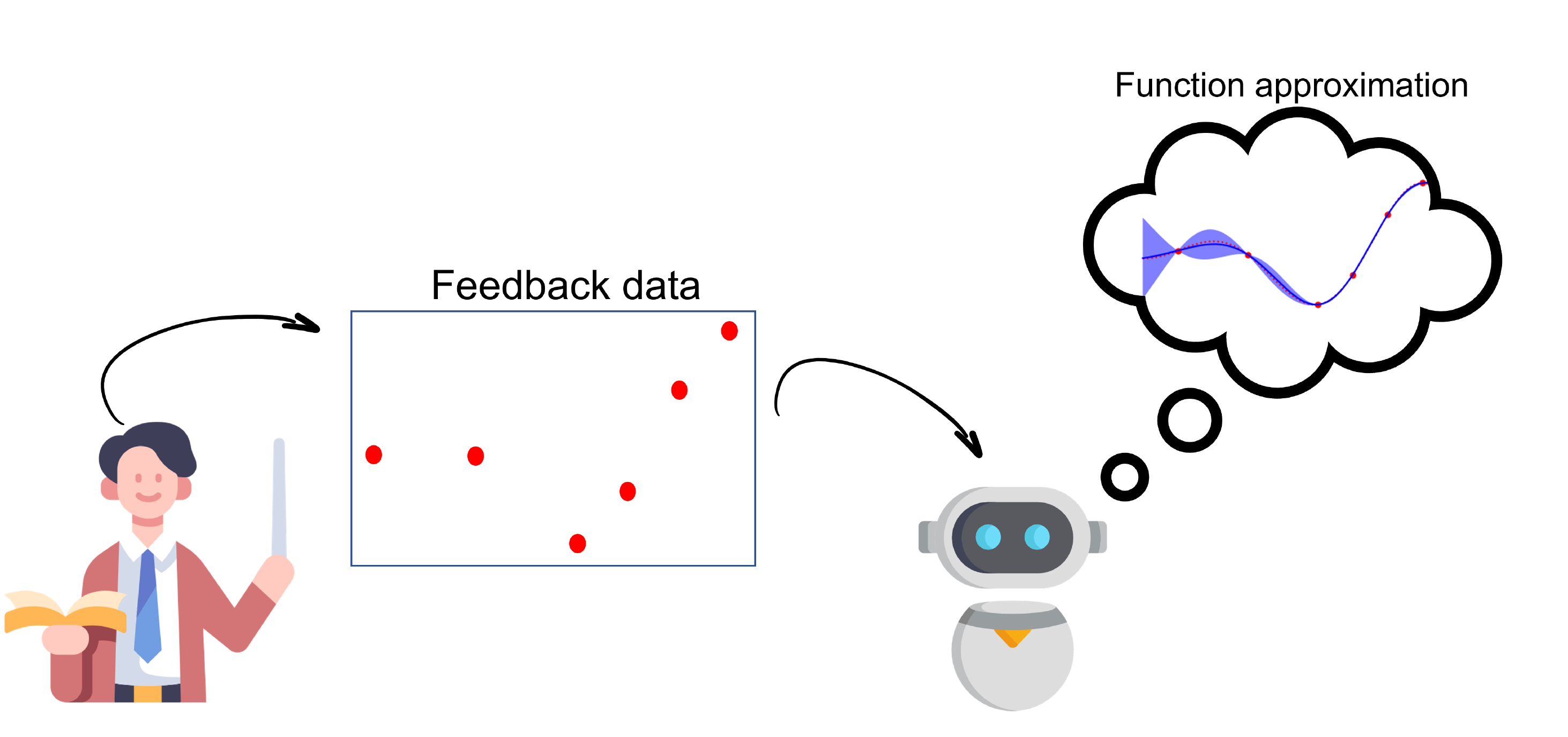}
    \caption{Function Approximations: the feedback of the teacher is used by the robot to reconstruct a function. }
    \label{fig:function_approx}
\end{figure}
In Chapter \ref{sec:ModelsLearned}, different models learned from interactions are analyzed, namely policies, rewards/objectives, and desired state transitions. These mathematical objects are estimated using finite data but are required to act on a continuous space, e.g. to generalize to previously unseen states, see Fig. \ref{fig:function_approx}.
To achieve the desired approximation, depending on the characteristics of the task at hand, design choices must be made on the function model, e.g. linear, non-linear, parametric, non-parametric, etc. 

The appropriate model representation enables, for instance, to cope with small/big databases, noisy/conflicting data, and a continuous stream of new information coming from the environment or from the teacher.
This section summarizes regressors and classifiers that have been adopted in the field of \gls{iil}.
Many of these methods could be considered as a particular case of a unified model \citep{stulp2015many}, showing the usability of different function approximations to different interactive learning algorithms. 

\section{Linear Models}
The simplest example of function approximation is linear regression, where the objective is to map inputs $X$ to a continuous output $Y$ with a linear dependency by means of a parameter vector $\Theta$, i.e,  
\begin{equation}
    Y = X \Theta.
\label{sec:eqlinear}
\end{equation}

\subsection{Combination of Features}
More generally, the function can be obtained with a linear combination of hand-crafted features that have a particular meaning in the context of the studied problem. The prediction is obtained as 
\begin{equation}
    Y=\Phi(X) \Theta,
\end{equation}
where $\Phi$ corresponds to the chosen features.

For example, \citet{spencer2020learning} obtains the value function as a linear combination of different features, i.e. boundary violation, absolute curvature to the next step, distance to the nearest obstacle, and distance to the goal path.
\citet{subramanian2016exploration} use this type of model to represent the value function for solving a task with reinforcement learning. These models are also employed to compute statistical measures that were used to actively query the user when necessary.
In the field of reward learning from demonstration and correction \citep{bajcsy2017learning} or preferences \citep{sadigh2017active}, a linear model of hand-crafted features is investigated for solving the \gls{pomdp} through online inference. This function approximator has the advantage of easily integrating prior knowledge in the model while keeping a linear complexity. 

\subsection{Radial Basis Functions}
\gls{rbf} correspond to a specific type of feature model that has been widely used in the \gls{iil} literature.
Its value depends only on the distance between the input and some fixed points. 
\citet{knox2009interactively} use \glspl{rbf} policies to encode non-linear policies that are updated online from human reward. 
Similarly, \citet{celemin2015coach} also use \glspl{rbf} controllers where the policy is updated with feedback in the action domain.
\citet{macglashan2017interactive} use an actor-critic strategy to update the policy parameters. In fact, using the same update rule based on gradient descent, the user can provide policy-dependent feedback, i.e., the expected value function of that state, or advantage function, that describes how much better or worse an action selection is compared to the agent's performance following its policy.
The RBF models allow fitting smooth functions, ideal for direct robot control with a small set of parameters. The location and the width of the basis functions can be chosen based on prior knowledge. The main advantage of these models is the stable update of their parameters, each new data point has a local effect on the function after the update, something very convenient for incremental learning approaches.

\subsection{Locally Weighted Regression}
One alternative for solving Eq. \ref{sec:eqlinear} is finding the optimal parameters $\Theta$ with the pseudo-inverse of X, $\Theta= (X^TX)^{-1}X^TY$.

This formulation can be extended to fit nonlinear relations by augmenting the parameter space with a weight matrix $W$, known as \gls{lwr}:
$\Theta= (X^TWX)^{-1}X^TWY$, where $W$ is a symmetric matrix and each element is defined by a function that tells you how much the individual values of $X$ and $Y$ should be considered when fitting a line through the neighborhood of $X$.
Hence, given a database, the prediction can be obtained as a weighted average of all the terms. 

\gls{lwr} has been applied for fitting non-linear functions without the requirement of choosing the degree of the function approximator nor the number of necessary parameters such as in the RBF controller. 
For example, \citet{argall2008learning,argall2011teacher} employ \gls{lwr} to fit a policy from state-action pairs that are collected from demonstrations or interactively updated using a teacher's advice. When feedback is provided, a new data point is aggregated only if the query input is not close to the original data, avoiding the collection of conflicting or redundant labels in the database.

\section{Gaussian Process}
\acrfullpl{gp} provide the means for making predictions while incorporating prior knowledge about the distribution of the data \citep{Rasmussen2005}.
We assume that all training and test labels are drawn from an $(n+m)$-dimensional Gaussian distribution $\mathcal{N}(0, \Sigma)$, where $n$ is the number of training points, $m$ is the number of testing points and $\Sigma$ corresponds to the covariance matrix. When querying the model on $m$ test inputs, its prediction is an m-dimensional gaussian distribution, with mean and variance, after conditioning the $m+n$ distribution on the $n$ training points. The variance quantification is a measure of the epistemic (model), uncertainty. The statistical formulation and the quantification of uncertainty make the \gls{gp} desirable when limited data are available; however, on the other hand, they do not scale well with big databases or high-dimensional inputs. 

%
\citet{wout2019learning} use \glspl{gp} for computing the uncertainty of the policy used for active queries, for approximating the human model used for adapting the learning rate, and for deciding  whether the new data instances interactively obtained are used to be appended in the dataset or to modify a data point of the dataset. 
\glspl{gp} have been successfully applied for learning reward functions using human preferences \citep{biyik2020active}.
The uncertainty quantification of the \gls{gp} is employed to generate active queries to the user, increasing data efficiency.

\citet{meszaros2022learning} and \citet{franzese2021ilosa} propose to use \glspl{gp} for motion learning. The motion is supported by minimizing the epistemic uncertainty of the \gls{gp}, which is also used to interactively update the database.

\section{Gaussian Mixture Model}
While in \gls{gp} the hypothesis is that the labels are sampled from an infinite dimensional Gaussian distribution, in a \gls{gmm} the hypothesis is that the input and output pairs are samples of a joint distribution defined by a superposition of $m$ Gaussians of dimension $(j+k)$, being $j$ and $k$ the dimension of the input and output features, respectively.
The prediction is obtained as a conditioning of the joint distribution on the provided input.
\citet{chernova2009interactive} use \glspl{gmm} for classifying discrete actions and for generating an active and interactive agent, which queries the user in uncertain situations. Moreover, it also allows the aggregation of new labels, which are used for retraining the classifier.
In that work, the \glspl{gmm} are used for quantifying aleatoric uncertainty i.e., the uncertainty in the data.
This type of uncertainty is commonly employed to capture the variability of the demonstrations or to spot conflicting labels.

\section{Support Vector Machine}
\gls{svm} is a supervised classification algorithm that maps each data point to an $n$-dimensional space and identifies the hyperplane that best separates the points belonging to different classes. 

In the context of interactive learning, \citet{laskey2016shiv} combine the use of \gls{svm} with \gls{dagger} for the classification of the states as risky according to the proximity to the decision boundary of the \gls{svm}. The use of this strategy shows a reduction of the asked query to the user when performing interactive learning. 

\section{Neural Networks}\label{sub:neuralnetworks}
In general, \glspl{nn} correspond to a family of function approximators where the approximation is achieved through a composition of multiple operations, known as \emph{layers} \citep{goodfellow2016deep}. Currently, \gls{nn} are known as a tool that can scale  well in terms of size, input/output dimensionality, and amount of training data \citep{goodfellow2016deep}, which opens opportunities for new technologies and theory. The advances that \glspl{nn} have experienced in the past decade created a turning point in different research fields, and \gls{iil} has not been an exception. 
%

Despite the opportunities that \glspl{nn} present, different challenges must be overcome to use them successfully, where requiring large amounts of data, vanishing gradients, and catastrophic forgetting are some of them. Hence, several \gls{iil} works focus on unfolding the advantages of \glspl{nn} while overcoming these different challenges \citep{warnell2018deep, mandlekar2020human, perez2020interactive, arumugam2019deep}.

Inspired from the \gls{rl} literature, several \gls{iil} methods employ \emph{replay buffers} throughout the agent's learning process to avoid locally overfitting to the last set of feedback signals and to use the learning data more efficiently \citep{arumugam2019deep,warnell2018deep}. However, to use replay buffers successfully, the methods must be able to learn off-policy, which is discussed in depth in Chapter \ref{sec:on off policy learning}.

Similarly, another strategy that has been borrowed from the \gls{rl} literature is to add a penalty in the update rule of the policy to avoid large changes in the policy's parameters each time the policy is updated. \citet{mandlekar2020human} includes a penalty loss in a learning from intervention framework where two buffers are used, one that stores the interventions and another one that stores the transitions executed by the learner. Every time the policy is updated, data is sampled from both buffers; hence, the learner improves its behavior from the interventions while avoiding drastic changes in its behavior, as it is also trained from data generated by itself. \citet{chisari2022correct} extended this idea by including evaluative feedback in the learning framework. Therefore, instead of storing every learner's transition into a buffer, only the transitions with positive feedback are stored, avoiding the policy to learn from its own erroneous behavior.

In contrast, \citet{prakash2020exploring} focus on the challenge that unbalanced datasets can present to \glspl{nn}. \gls{dagger} is employed to collect autonomous driving demonstrations online; however, in such complex scenarios, there are situations that have a low probability of occurrence. Consequently, the dataset that the \gls{nn} uses for learning is largely occupied by frequent and similar situations and will have few instances of uncommon states. Hence, the network is likely to forget these unlikely situations and not behave properly when they occur. \citet{prakash2020exploring} propose to detect such situations with a measure of epistemic uncertainty using an ensemble of \glspl{nn} and increase their likelihood of being sampled from the learning dataset. 

Finally, \gls{srl} is another technique used to avoid overfitting and speed up the learning process. \gls{srl} methods, along with the \gls{iil} loss, optimize for auxiliary loss functions that are employed to learn state representations from high-dimensional data \citep{bohmer2015autonomous}. For more information regarding these methods, the reader is referred to Section \ref{sec: task features}. 

\section{Movement-Conditioned Models}\label{sec:movementmodels}
When dealing with robotics tasks, commonly, the user aims to teach a desired movement to the robot. 
For these models, the input can be the progress (or phase) of the movement or the current robot state. The output can be the desired position, velocity or acceleration. 

\subsection{Dynamic Movement Primitive}
\glspl{dmp} \citep{saveriano2021dynamic} generates a movement as the superposition of an attractor model and a non-linear function (known as a \textit{forcing term} $f(s)$). This forcing term enables the generation of complex trajectories while the attractor leads the robot towards the desired goal. 

\citet{schroecker2016directing} investigated the learning of \glspl{dmp} with \gls{ps} and initial demonstration or interactive corrections. In particular, the teacher gives a set of via points at the beginning of the training or the execution of the training can be stopped and the robot moved to the desired position at a particular time. The parameters of the \glspl{dmp} are updated every time a correction is provided in order to maximize the probability of actually going towards that new via-point. 
Alternatively, \citet{celemin2019reinforcement} used a different strategy for the correction that does not require stopping the robot. In fact, it allows the user to give corrective feedback on the desired state by modeling the motions with \glspl{dmp}.

\subsection{Probabilistic Movement Primitive}

A movement may be described through a combination of different primitives. For example, motions reaching toward different objects on a table may have similar starting behavior but, depending on where the object is located, their shape may vary. 
\glspl{pro mp} are a variant of \glspl{mp} that enables the capture of the probability distribution of the different demonstrations \citep{paraschos2013probabilistic}, as a combination of multiple primitives.
%
%
For example, multiple demonstrations with different goals would generate trajectories of variable shapes, with increasing variance towards the end. Then, when conditioning the motion on a point near the end of the trajectories, one primitive would be sampled according to this conditioned distribution.

In the context of Active Learning, it is important to adapt the motion not only from the human feedback, but also to minimize the risk of collision during the interaction with the human. Therefore, a modification of the learned \gls{pro mp} to perform collision avoidance is introduced by \citet{koert2019learning}. 

Alternatively to \glspl{pro mp}, the variability of the demonstrations can be conditioned on the context, e.g., the goal position, the mass of the manipulated object, etc.
This approach is proposed by \citet{maeda2017active} using \gls{gp} and by \citet{kulak2021active} using \glspl{gmm} for interactive learning of the forcing term of a \gls{dmp}.

\subsection{Kernelized Movement Primitives}
A formulation of \glspl{mp}, known as \glspl{kmp}, is introduced by \citep{huang2019kernelized}, where a multi-output formulation is proposed to embed the variability of multiple demonstrations.
Although this method is general, it is also tested in an \gls{iil} setting. To adapt the robot's trajectory when the environment changes, the use of a force sensor installed at the end-effector of the robot is used to measure corrective forces exerted by the human.

\section{Discussion}
In this section, an overview of different regression and classification methods is introduced with their application in the context of \gls{iil}. An important feature of function approximation methods in \gls{iil} is the possibility to perform an online update of the function when interactive feedback is provided, and to be able to deal with conflicting data (e.g., \gls{pro mp}). 

Additionally, the use of bayesian methods, like \glspl{gp}, provides a quantification of the epistemic uncertainty that has been used for actively querying the user to enhance a safer exploration of the robot and interaction with the human. Finally, recent developments of \gls{nn} allow dealing with high-dimensional sensory inputs while interactively aggregating data from humans. These advances are opening the possibility for \gls{iil} algorithms to be applied to new settings where it was not possible before, such as household environments. 
\newpage
\chapter{On/Off Policy Learning}\label{sec:on off policy learning}

Machine Learning methods intended to solve sequential decision-making problems (like \gls{rl} or \gls{il}) feature different algorithmic properties related to what kind of data is used for learning, when and how it is generated, and how it is used for updating the policy.
Depending on how these questions are approached by the method designer, learning processes could be classified as on/off-line learning and on/off-policy learning.

The chronological evolution of the focus on the way the learning data is generated for \gls{il} has been relatively opposite with respect to \gls{rl}.
Initially, the main idea of \gls{rl} was the autonomous learning of a policy by trial and error, while the agent is interacting with the environment, i.e., collecting the data samples while testing the learning policy.
However, in recent years, researchers have dedicated efforts to an additional branch for applying the \gls{mdp}s properties, and \gls{rl} concepts, for learning from prerecorded data without further agent-environment interaction, as is the case with offline RL \citep{levine2020offline}.
On the other hand, \gls{il} was studied for many years only to find methods that could replicate behaviors contained in static datasets of expert demonstrations, and only later it has been explored the idea of incrementally collecting data from the teacher who observes the learning agent performance.

Due to the different development of these two learning paradigms, general common concepts have been independently introduced.
In this section, a discussion intending to unify the definitions of these concepts given in both the \gls{rl} and \gls{il} literature is presented, while trying to keep the \gls{rl} definitions as the reference. 

\section{Online and Offline Learning}\label{sub:on/off-lineLearning}
Depending on when the collection of data used for learning is carried out, the learning methods could be classified into offline or online learning.
In \gls{rl}, the offline learning setting is defined as the situation when \emph{``the agent no longer has the ability to interact with the environment and collect additional transitions using the behavior policy. Instead, the learning algorithm is provided with a static dataset of transitions and must learn the best policy it can using this dataset''} \citep{levine2020offline}.
In contrast, in the online learning setting, the experience the agent gathers for learning increases with new interactions with the environment, allowing it to improve the current policy.

The projection of these definitions into the world of \gls{il} matches completely with the classification of interactive and non-interactive methods.
Offline learning covers the standard \gls{il} methods that sequentially record demonstrations in a static dataset, and later obtain a policy with the recorded data. 
Online \gls{il} methods cover the group of \gls{iil} approaches because they feature the ability to collect more data with a dynamic dataset during learning.
Since in \gls{il} the data collection depends on a teacher, the continuous feedback sampling of online learning involves the teacher in the loop as it has been defined for \gls{iil}.

As mentioned in the introduction of this chapter and considering the introduced definitions, we could say that \gls{rl} was initially developed for online learning, and only recently its potential for learning offline has been studied, while \gls{il} was first formulated offline, and recently extended to the online setting.

In both \gls{rl} and \gls{il}, the agent learns from the obtained feedback, provided by the environment or teacher intervention, respectively.
Both learning paradigms aim at a similar objective in the offline case, since both try to obtain a policy that reproduces the behavior recorded in the data.
In other words, offline \gls{rl} also tries to imitate the demonstrations collected in a dataset; however, it makes use of a reward function that supports the process of defining which decisions in the demonstrated data are more relevant and which ones are less convenient for attaining the task goal.

\section{On-policy and Off-policy Learning}\label{sub:on/off-policy-learning-sub}

\begin{figure}
    \centering
    \subfloat[On-policy Learning]{\includegraphics[trim={15 90 270 90},clip,width=0.5\textwidth]{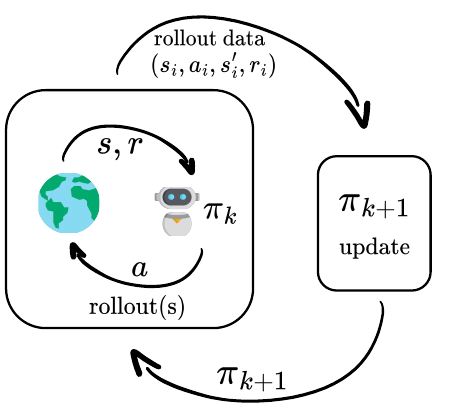}} 
    \subfloat[Off-policy Learning]{\includegraphics[trim={15 85 270 90},clip,width=0.5\textwidth]{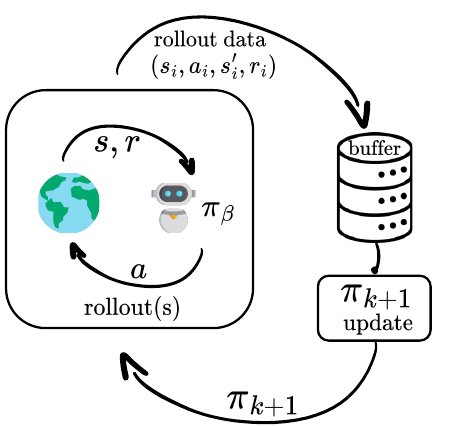}} \\
    \subfloat[Offline Learning]{\includegraphics[trim={15 90 190 93},clip,width=0.75\textwidth]{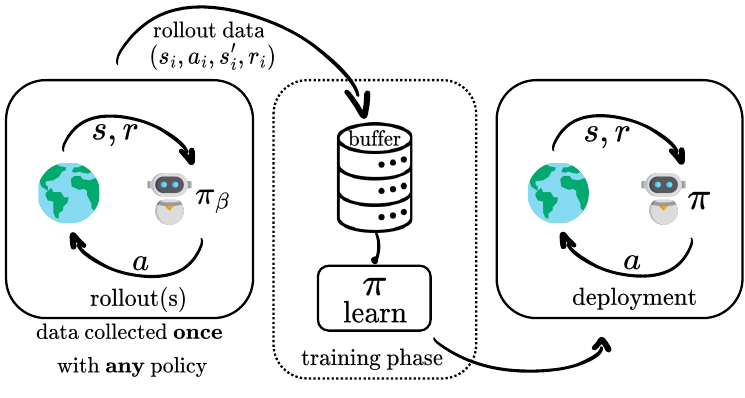}} 
    \caption{Different learning schemes used in \gls{rl} that are applicable to \gls{il}. In on-policy learning, the target policy $\pi_k$ is the same as the behavior policy. This policy collects the data used in the update that leads to $\pi_{k+1}$ (a). In off-policy learning the behavior policy $\pi_{\beta}$ is different from the target policy, allowing the use of a replay buffer. However, in practice, $\pi_{\beta}$  results from combining $\pi_k$ with exploration noise or input from the teacher (b). In offline learning the behavior policy $\pi_{\beta}$ used for obtaining the data is completely different from the policy $\pi$ obtained in the learning process, which is not considered for the data collection (c). This figure is inspired by \citet{levine2020offline}, with some modifications.}
    \label{fig:online_offline_RL}
\end{figure}

Since, historically, offline learning has been predominately applied in \gls{il} methods, two ideas that become evident in online learning scenarios have been mostly ignored in its literature: \emph{on-policy} and \emph{off-policy} learning (see Figure \ref{fig:online_offline_RL}). 
These ideas have been well defined and deeply studied by the \gls{rl} community, and they play a fundamental role in the understanding and design of the learning methods. 
In this section, we argue that the relevance that on-policy and off-policy learning has in \gls{rl} also transfers to \gls{iil}. 
However, although some works have used these concepts in the context of \gls{iil} \citep{laskey2017dart, arumugam2019deep, balakrishna2020policy}, they are still not clearly defined in this field. Therefore, to analyze the relevance that on/off-policy learning has in \gls{iil}, it is necessary to first clearly define it for this case.

Below, we introduce these concepts from the original definitions in the literature of \gls{rl}, and thereafter they are extended to the \gls{iil} case. 

\subsection{On/Off-Policy Learning in Reinforcement Learning}\label{sec:on-off-learning}

\citet{sutton2018reinforcement} define these concepts stating: \emph{``on-policy methods attempt to evaluate or improve the policy that is used to make decisions, whereas off-policy methods evaluate or improve a policy different from that used to generate the data''}.
The policy that is being learned is often referred as \emph{target policy} $\pi^{\text{t}}$, and the policy used to generate the learning data as \emph{behavior policy} $\pi^{\text{b}}$. Then, on-policy learning occurs when the learning data comes from trajectories generated by the target policy, i.e., $\pi^{\text{t}}=\pi^{\text{b}}$.
In contrast, if $\pi^{\text{t}} \neq \pi^{\text{b}}$, the learning is off-policy. Note that, consequently, offline \gls{rl} requires off-policy learning.

These concepts can be formally defined from the \gls{rl} objective and from how it is commonly optimized. From Section \ref{sec:background}, Eq. \eqref{eq:rl_objective}, we have that this objective commonly corresponds to the maximization of the discounted expected return
\begin{equation}
    \POLICY^* = \argmax_{\POLICY \in \Pi} \EXPECT_{\tau \sim p_{\pi}(\tau)} \left[G(\tau)\right],
    \label{eq:rl_objective_2}
\end{equation}
where $G(\tau) = \sum_{t=0}^{T}\gamma^{t} \mathcal{R}(s_{t}, a_{t})$ corresponds to the return.

In practice, since we do not have analytical access to this expectation, to find the policy $\POLICY^*$ that maximizes the presented objective, it is necessary to empirically collect information from, ideally, every possible trajectory $\tau$ (or transition $(s_{t}, a_{t}, s_{t+1}, r_{t+1})$ given the Markov assumption) and shift the behavior of $\pi^{\text{t}}$ towards the trajectory distribution that maximizes Eq. \eqref{eq:rl_objective_2}. However, in most realistic scenarios, it is not possible to sample the complete state-action space; hence, a policy is commonly chosen to sample this space as diversely and exhaustively as possible while keeping the problem tractable. This policy is $\pi^{\text{b}}$. Then, at every update iteration, the data collected by $\pi^{\text{b}}$ is employed to estimate the current expected return of the trajectory distribution induced by $\pi^{\text{t}}$, and $\pi^{\text{t}}$ is modified such that this expectation increases. However, if data is generated by sampling trajectories induced by $\pi^{\text{b}}$, \emph{how is the expected return computed with respect to $\pi^{\text{t}}$?} There are two options, 1) on-policy learning, i.e, directly improve $\pi^{\text{b}}$ at every iteration ($\pi^{\text{b}}=\pi^{\text{t}}$), 2) off-policy learning, i.e., $\pi^{\text{b}} \neq \pi^{\text{t}}$ and employ a strategy to compute the expected return of $\pi^{\text{t}}$ from trajectories collected by $\pi^{\text{b}}$. Consequently, at every learning iteration, the estimated objective of an on-policy learning method corresponds to 
\begin{equation}
    \textbf{on-policy: } \hat{\EXPECT}_{\tau \sim p_{\pi^{\text{b}}}(\tau)} \left[G(\tau)\right],
    \label{eq:on-policy}
\end{equation}
where $\hat{\EXPECT}_{\tau \sim p_{\pi}(\tau)}$ corresponds to the expectation estimated by sampling data from the environment following a policy $\pi$.

In contrast, the estimated objective of off-policy methods, even though the data comes from $\pi^{\text{b}}$, corresponds to 
\begin{equation}
    \textbf{off-policy: } \hat{\EXPECT}_{\tau \sim p_{\pi^{\text{t}}}(\tau)} \left[G(\tau)\right].
    \label{eq:off-policy}
\end{equation}

Note that off-policy methods are defined as those that are able to learn off-policy, which indicates that it is also possible to learn on-policy with these methods, as they are capable of learning from data generated by any policy, which includes the target policy \citep{sutton2018reinforcement}.

The \gls{rl} literature provides a vast family of on-policy and off-policy learning methods, to study how the concepts of on/off-policy are applied in practice we can analyze some examples.

\subsubsection{SARSA and Q-Learning}
To illustrate the difference between on-policy and off-policy methods, let us study SARSA \citep{rummery1994line} and Q-Learning \citep{watkins1989learning}, two seminal RL methods. SARSA is on-policy and Q-Learning is off-policy. These methods employ \gls{td} learning to compute estimates of the expected return and solve eq. \eqref{eq:rl_objective_2}. \gls{td} combines ideas from \gls{mc} methods and \gls{dp}, i.e, trajectories are empirically sampled from the environment, but the final outcome is estimated based on current models of the environment (which is known as \emph{bootstrapping}), instead of only using the sampled data. SARSA and Q-learning employ \gls{td} learning to estimate the action-value function $Q(s_{t}, a_{t})$, which estimates the expected return of a policy given its current state and selected action and derive a policy from it. Hence, the environment can be sampled following $\pi^{\text{b}}$ and bootstrapped at every time step to get the following sample/estimation of the return:

\begin{equation}
    G(\tau)_{t}=\underbrace{r_{t+1}}_{\text{sample}} + \underbrace{\gamma Q(s_{t+1}, a_{t+1})}_{\text{estimation}}.
    \label{eq:q-bootstrap}
\end{equation}
Then, $Q$ can be updated by computing the error of this \gls{td} estimate with respect to the current estimation of $Q$ for a given time step, which is known as the update rule of SARSA:  
\begin{equation}
    \textbf{SARSA: } Q^{\text{new}}(s_{t}, a_{t}) \leftarrow Q(s_{t}, a_{t}) + \alpha [\underbrace{\underbrace{r_{t+1} + \gamma Q(s_{t+1}, a_{t+1})}_{\text{TD target}} - Q(s_{t}, a_{t}}_{\text{TD error}})],
    \label{eq:SARSA}
\end{equation}
where $\alpha$ is the learning rate of the update. Note that the only variable that indicates that Eq. \eqref{eq:SARSA} is following $\pi^{\text{b}}$ is $a_{t+1}$, as the other variables are a consequence of the action taken by $\pi^{\text{b}}$ one time step before, which can be ignored at $t+1$ given the Markov assumption. Hence, it is possible to remove the dependence of the TD-target from $\pi^{\text{b}}$ if instead of using the action $a_{t+1}$ sampled from $\pi^{\text{b}}$ in this estimate, a different one is chosen. This idea can be followed to create an off-policy variation of SARSA, known as Q-learning.

Q-learning defines its target policy as the optimal policy according to the current estimation of $Q$, i.e., $\pi^{t}(s_{t}) = \argmax_{a} Q(s_{t}, a)$. Then, Eq. \eqref{eq:SARSA} can be modified by replacing the term $Q(s_{t+1}, a_{t+1})$ with the $Q$ value of $\pi^{\text{t}}$, making the return estimation to be according to $\pi^{\text{t}}$ instead of $\pi^{\text{b}}$, i.e, 

\begin{multline}
    \textbf{Q-learning:} \\
    Q^{\text{new}}(s_{t}, a_{t}) \leftarrow Q(s_{t}, a_{t}) + \alpha [r_{t+1} + \gamma \red{\max_{a}Q(s_{t+1}, a)} - Q(s_{t}, a_{t})],
    \label{eq:q-learning}
\end{multline}

where the modified value with respect to Eq. \eqref{eq:SARSA} is highlighted in red. Note that in the special case where the behavior policy is greedy with respect to the current estimate of $Q$ (e.g., when using an $\epsilon$-greedy exploration strategy and $\epsilon$ tends to zero ), the SARSA and Q-Learning update rules are equivalent \citep{rummery1994line} because the target and behavior policies are the same, i.e., Q-Learning becomes on-policy.

\subsubsection{Importance Sampling}
Another well-known approach for designing off-policy learning methods is \emph{importance sampling}.
Importance sampling allows methods that in nature are on-policy, such as SARSA, to become off-policy by weighting the \gls{td} errors with the importance sampling ratio \citep{sutton2018reinforcement, mahmood2017incremental}. 
The importance sampling ratio is employed to estimate the update of the $Q$ function of the target policy from data generated by a different policy, e.g., the behavior policy.
Closely related to the methods studied above, the method \gls{td-dis}\footnote{The acronym \gls{td-dis} is introduced in this work for simplicity, given that no acronym is proposed in \citep{precup2000temporal}.} \citep{precup2000temporal} can be analyzed in this case. 
\gls{td-dis} method can be seen as an off-policy extension of SARSA by means of importance sampling \citep{rubinstein2016simulation, hammersley1964monte}. The update rule of \gls{td-dis} is
\begin{multline}
    \textbf{TD-DIS: } \\
     Q^{\text{new}}(s_{t}, a_{t}) \leftarrow Q(s_{t}, a_{t}) + \alpha \rho_{t} [r_{t+1} + \gamma Q(s_{t+1}, a_{t+1}) - Q(s_{t}, a_{t})],
\end{multline}
where 
\begin{equation}
    \rho_{t} = \frac{\pi^{t}(a_{t}|s_{t})}{\pi^{b}(a_{t}|s_{t})}
\end{equation}
is the per-step importance sampling ratio between the target policy and the behavior policy. The similarities between SARSA and \gls{td-dis} are evident; $\rho_{t}$ is the only variable that differentiates both methods and allows the update rule of \gls{td-dis} to be employed with data collected by the target policy. In the special case where this method is used on-policy, the behavior and target policies become the same (i.e., $\rho_{t}=1$) and \gls{td-dis} becomes equivalent to SARSA.

\section{On-Policy/Off-Policy Learning in Imitation Learning}\label{sub:onoffpolicy-imitationlearning}

According to \citet{osa2018algorithmic}, the terms on-policy and off-policy, in the \gls{il} literature, are mentioned for the first time in \citet{laskey2017dart}. The authors use the terms on-policy and off-policy according to which policy is used to sample data from the environment. 
If the current agent's policy is used to sample data, then the method is on-policy; if the teacher's policy is used to sample data, then the method is off-policy. 
Although this definition may seem equivalent to the one in \gls{rl}, there is a difference: in \gls{rl}, these definitions are about the data that is used in the evaluation or improvement of the \emph{current} agent's policy. 

This difference is important because online \gls{rl} and \gls{iil} are iterative learning processes (i.e., the agent's policy is evolving over time while it interacts with the environment), which means that the distribution of the data generated by an older version of the agent's policy is not the same as the one generated by the current agent's policy. 
The on/off-policy definitions provided in \citet{laskey2017dart} allow on-policy methods to use data generated by older versions of the agent's policy (i.e., other policies) when improving its behavior, which is not consistent with the \gls{rl} definition. 

As an example, \gls{dagger} \citep{ross2011reduction} has commonly been defined as being on-policy and Behavioral Cloning as off-policy \citep{osa2018algorithmic, laskey2017dart, balakrishna2020policy}. 
Nevertheless, in \gls{dagger}, data is constantly being aggregated in a dataset that is used to update the agent's policy iteratively. 
Consequently, data generated with a different policy than the target policy is used in the update rule, and, therefore, from an RL perspective, it would be an off-policy method. 
From this point of view, \gls{dagger} and Behavioral cloning are in the same category. 

Instead, we argue that the ideas of on/off-policy learning can be transferred differently to \gls{iil}. In this section, we focus the analysis on the per-step feedback case as described in Section \ref{sec:iil-objective}, as it is the case that most resembles \gls{rl}.

\subsection{From Reinforcement Learning to Interactive Imitation Learning}
From the definition of on/off-policy learning in \gls{rl} provided in Section \ref{sec:on-off-learning}, we can recall that off-policy learning occurs when data collected with one policy (i.e., behavior policy) is employed to update a different one (i.e., target policy). 
Consequently, off-policy learning allows updating a policy from data that has no dependence on it.

This same idea can be employed to study on/off-policy methods in \gls{iil}, i.e., if the data used in the update rule of the learner's policy follows a different policy, the learning method is off-policy; otherwise, it is on-policy. The only difference is that, in this case, the learner collects the feedback signal when interacting with the environment, instead of the reward signal like in \gls{rl}. As mentioned in Section \ref{sec:mdp_iil}, the feedback signal can be understood as a generalization of the reward.

To observe this more clearly, we can analyze methods from the two paradigms that lead to \gls{iil} methods (see Section \ref{sec:perstep}): 1) Value Maximization and 2) Divergence Minimization.

\subsection{Value Maximization methods}
Since these methods derive from the \gls{rl} literature, they optimize the \gls{rl} objective, and, therefore, the definitions provided in Section \ref{sec:on-off-learning} can be directly used to define them as being on-policy or off-policy. Let us study two of these methods: \gls{coache} \citep{macglashan2017interactive} and \gls{tamer} \citep{knox2008tamer}.

\paragraph{COACHe}
\gls{coache} is derived employing the \emph{policy gradient theorem} of \gls{rl} \citep{sutton2018reinforcement}. This theorem allows to directly improve the parameters $\theta$ of a policy $\pi$ by computing the gradient of its value function and applying stochastic gradient ascent. Consequently, \gls{coache} applies the following update rule to its policy:

\begin{equation}
    \theta^{\text{new}} \leftarrow \theta + \alpha \nabla_{\theta}\pi(s_{t},a_{t}) \frac{h_{t+1}}{\pi(s_{t},a_{t})},
\end{equation}

where $\alpha$ is the learning rate and $h_{t+1}$ the human feedback. Here, $h_{t+1}$ can be interpreted as replacing the \emph{advantage function} used in this type of policy gradient methods, which describes how much better or worse an action would perform compared to the agent's action when following the agent's policy. Consequently, to improve the agent's policy with this method, it is necessary to learn \textbf{on-policy}; otherwise, $h_{t+1}$ would indicate the advantage of an action with respect to a policy different from the agent's policy, making its update incorrect.

\paragraph{TAMER}
\gls{tamer} can be interpreted as a method that maximizes the Q function for deriving a policy but assumes that the policy behaves \emph{myopically} (i.e., $\gamma=0$). Therefore, we can observe that if we assume a myopic behavior, Eqs. \eqref{eq:SARSA} and \eqref{eq:q-learning} reduce to the same solution, which corresponds to the update rule employed by \gls{tamer}

\begin{equation}
    Q^{\text{new}}(s_{t}, a_{t}) \leftarrow Q(s_{t}, a_{t}) + \alpha [h_{t+1} - Q(s_{t}, a_{t})],
\end{equation}

where the reward $r_{t+1}$ is replaced by the human feedback $h_{t+1}$. Consequently, \gls{tamer} interprets the feedback signal as a Q-value, which does not depend on the agent's policy \citep{zhang2019leveraging}. Moreover, it does not depend on \emph{any} policy, but only on immediate actions. Consequently, with \gls{tamer}, it is possible to update the target policy with data collected by any policy, making it an \textbf{off-policy} learning method. 
Note that although it is likely that the teacher will provide feedback as a function of the learner's policy \citep{knox2009interactively,macglashan2017interactive}.
For instance, \gls{tamer} is proposed considering assumptions such as \emph{``The trainer can evaluate an action or short sequence of actions, considering the long-term effects of each''} (effects with respect to the policy) and \emph{``a human trainer’s reinforcement function, is a moving target. Intuitively, it seems likely that a human trainer will raise his or her standards as the agent’s policy improves''}. However, from an algorithmic perspective, given the proposed implementation, the feedback is policy-independent. 

\subsection{Divergence Minimization methods}
Let us  recall Eq. \eqref{eq:dagger_form}, which summaries \gls{iil} methods based on Divergence Minimization and rewrite it in its \gls{mle} form \citep{ke2020imitation}, instead of the \gls{kl} divergence between two policies; then, in each training iteration we solve 
\begin{equation}
    \max_{\pi \in \Pi} \EXPECT_{s \sim p_{\pi}(s), a \sim \pi^{h}}\left[\ln(\pi(a|s))\right].
\label{eq:divergence2}
\end{equation}

This form of the equation is useful for analyzing the on/off-policy nature of these methods because it explicitly shows the state and action distributions. Here, we analyze two \gls{iil} methods derived from this equation: \gls{dagger} \citep{ross2011reduction} and \gls{coach} \citep{celemin2019interactive}.

\paragraph{DAgger}
In \gls{dagger}, Eq. \eqref{eq:divergence2} is minimized by setting the feedback signal to $h_{t}=a_{t}$. Hence, $h_{t}$ directly corresponds to a label of the optimal action (according to the teacher) for a given state. This method assumes that for every sampled state from which this equation is minimized, the label $h_{t}$ does not depend on the current agent's policy, as it only follows $\pi^{h}$. Therefore, it is possible to update the agent's policy from data generated by any policy, which makes \gls{dagger} an \textbf{off-policy} learning method. 

Nevertheless, there is one important remark to make. Given that the state distribution of the samples used to update Eq. \eqref{eq:divergence2} depends on the behavior policy, a different behavior policy will, inevitably, yield different solutions. However, this is also the case for \gls{rl} methods, so this definition is still consistent with them. 

\paragraph{COACHc}
In \gls{coach}, the assumption is that the feedback signal corresponds to an error signal that indicates the direction in which the current agent's policy should be modified to improve its performance (feedback is only meaningful for improving the current behavior policy, and not future versions of it). Therefore, in this case, for every training iteration, an action label is generated as a function of the learner's policy with the form $a=\pi^{l}(s) + e \cdot h_{t}$, where $e$ is a hyperparameter defined as the error magnitude. Consequently, Eq. \eqref{eq:divergence2} gets modified, since the actions do not distribute as $\pi^{h}$ anymore, but rather as a different distribution that depends on $\pi^{l}$. Therefore, it is only possible to update \gls{coach} with samples collected by the agent's policy, making this method an \textbf{on-policy} learning method.

\section{Discussion}\label{sub:DiscussionOnOffpolicy}
The concepts discussed in this section are as important in \gls{iil} as in \gls{rl} because they are agnostic of the feedback source used for policy improvement (teacher or environment).
Instead, they are related to the way the learning experience is obtained and used in the policy updates.

The replay of recorded experience and the way it is implemented is one of the main features that come into the discussion of On/Off-policy learning.
But unlike \gls{rl}, wherein the reward function (that could be deterministic or stochastic) is (time or policy) invariant, \gls{iil} methods could have in some cases feedback of the teacher that depends on the performance of the policy.
Therefore, depending on the assumption about the teacher's feedback within a learning method, it is relevant to evaluate what kind of learning is the most convenient for the method implementation, such that it leverages that assumption.

Since in online learning the experience is incrementally collected, there are additional challenges when fitting function approximators with this kind of data.
The sequential nature of these problems makes the data have spatio-temporal correlations, therefore not following the IID assumption of most \gls{ml} approaches.
Additionally, when training \gls{nn}s from a static dataset, the iterative process of updating the model normally reduces the error for most training data as long as there are sufficient iterations. 
That is because some data points require more update steps following the cost function gradient than others.
However, when data samples are obtained incrementally while also learning, it is difficult to control the model to avoid either overfitting or underfitting the data.
In both cases, there is the additional issue of the model being changed for other input-output mappings different from the ones used in the update, which counts as losing the already acquired knowledge, and is known as ``catastrophic forgetting''.

Experience replay is a technique introduced for breaking those correlations in the collected data during the policy update. 
It also helps to have a good balance for not overfitting to the most recent training data, while keeping the old experience in the \emph{memory} of the model, i.e., it helps to deal with the three problems previously mentioned.

For instance, in methods wherein the teacher provides evaluative feedback at any time step, there could be two different cases: 
\begin{enumerate}
    \item When the human feedback is assumed to replace the \gls{mdp} reward and used within an \gls{rl} implementation, the feedback is assumed to be consistent in all the state-action space, such that the \gls{rl} learning properties hold.
    In this case, the old feedback samples are never conflicting with the new ones, therefore, old feedback signals are always usable, and the choice of on/off-policy learning is left to the \gls{rl} implementation, being both valid.
    \item When it is assumed humans consider past and future in their evaluative feedback signals, and it is used as something equivalent to value function (e.g., \gls{tamer}), i.e., the feedback depends on the policy.
    Consequently, feedback given over state-action pairs of old policies could be contradictory with respect to the one obtained with the current policy.
    This assumption requires giving priority to the feedback given to the execution of the current policy, hence on-policy learning would be more appropriate.
\end{enumerate}
Since the discussions of On/Off-policy learning are relatively new and not consolidated in \gls{il}, this dimension of the algorithmic features space has been neglected in some implementations of \gls{iil} methods. 
Some algorithms have considered a learning strategy that does not align with the assumptions of the required human feedback.
It is not simple to implement methods whose algorithmic features match the feedback assumptions because the limitations created by the aforementioned problems (non-IID, over/under-fitting, catastrophic forgetting) condition the learning strategies.

The most common case of having inconsistent implementations is when the feedback is policy-dependent, but experience replay is required for stable learning, i.e., on-policy learning deals better with the assumed policy-dependent feedback, but the need for experience replay makes it necessary to learn from off-policy data.
As mentioned before, importance sampling helps for decreasing the priority of data obtained with different policies \cite{degris2012off}, which is convenient for learning from off-policy data with methods whose assumptions align with the on-policy learning conditions.
The \acrfull{coache} algorithm \cite{arumugam2019deep} is a good example of \gls{iil} with a policy-dependent feedback assumption (naturally on-policy), which benefits of off-policy learning for stability, but using importance sampling to prioritize the data in the updates according to the distribution of the current policy.

\newpage
\chapter{Reinforcement Learning with Human-in-the-Loop}\label{sec:reinforcement learning with HIL}
 
 %

As explained in the \nameref{sec:background} section, \gls{rl} represents a learning framework where an \emph{autonomous agent} learns the desired behavior by interacting (in a form of \emph{trial-and-error}) with the \emph{environment} that provides the \emph{reward} (as a feedback signal). Defined as such, \gls{rl} offers a general framework and, in general, all other entities except the autonomous agent (e.g., other agents/humans) can be considered as a part of the environment. However, considering \emph{humans} as a distinct entity from the environment enables the design of algorithms that take advantage of this setting. 
There are some \gls{rl} problems that also consider interactions with humans that act in the environment (e.g., human collaboration). In contrast to these, in this section, we focus only on those problems where a human is influencing the \emph{learning loop} by providing feedback to the agent (e.g. has the role of an observer that provides feedback), and not just acting in the same environment as the agent. 
In these cases, we refer to the human as a teacher, as explained in Chapter \ref{sec:background}.

Looking from a \gls{rl} perspective, if the teacher is influencing the learning loop and it is the only one providing feedback to the agent (i.e., demonstrations, corrections, etc.), the problem reduces to the \gls{iil} problems. %
If an autonomous agent receives a feedback signal from both the environment and the teacher, it is learning with a \emph{RL with Human-in-the-Loop} method. 
We define as \gls{rlwhil} all \gls{rl} algorithms where the agent learns a desired behavior and a teacher is influencing one or more components of the reinforcement learning loop (e.g. reward, policy, exploration, etc.).

\section{Other related approaches}

As it was already discussed in Chapter \ref{sec:introduction}, several \gls{rl} approaches rely on human demonstrations. For instance, the approaches that pre-train a policy to warm-start the learning process, storing pre-recorded demonstrations for either off-policy \gls{rl} or offline \gls{rl}, or inferring rewards from demonstrations like in \gls{irl}. However, we do not consider these as \gls{rlwhil} as the human input is not part of the learning loop.
Several approaches rely on human demonstrations to initialize the policy and subsequently use \gls{rl} to further improve it (e.g.,
\gls{power} \citep{kober2008policy}, 
\gls{loki}  \citep{cheng2018fast}).
Pre-training the policy in a supervised manner from demonstration data before interacting with the real task is especially important for 
alphaGo \citep{silver2016mastering}, to have a reasonably good policy for starting self-play.
\gls{dqfd} \citep{hester2018deep} 
also pre-train the network with expert demonstrations and keeps them in a replay buffer, along with the data obtained from its own experience.

\section{Historical perspective}
Historically, \gls{rlwhil} evolved with \gls{rl} research from the beginning.
%
%
If we look for the origins of the idea of using teacher advice, we can find that they go even to the roots of founding the \gls{ai} field, with John McCarthy's \emph{Advice take} \citep{mccarthy1958programs} that was proposed to improve behavior by taking pieces of advice.
Some works already in the 90s try to combine Reinforcement Learning approaches with human advice  \citep{utgoff1991two,whitehead1991complexity,clouse1992teaching,lin1992selfimproving,maclin1994incorporating}.
\citet{utgoff1991two} is mainly inspired by advancements in the Checkers game-playing agents. The initial game-playing agent was based on \emph{learning from experience} \citep{samuel1959studies} and extended later with utilizing expert choice, typically called a \emph{book move} \citep{samuel1967studies}. 
The main idea by \citet{utgoff1991two} is that these are two fundamentally different kinds of training information useful for learning. Additionally, they argued that expert choice does not have to be recorded in a \emph{book} in advance. By watching an expert in action, or asking an expert what to do in a particular situation, an agent would be able to learn from an expert whenever it is available. The agent is able to query an external agent about the correct action to take when the confidence in its own decisions is low. Their agent is used to control a search algorithm by considering as an action which node to explore next, like in the checkers engine\citep{samuel1959studies}. 
\citet{clouse1992teaching} extend the idea beyond a search algorithm to the cart-pole task and provide an online mechanism that allows an external agent to guide it while it performs the task.

On the other hand, \citet{whitehead1991complexity}, inspired by the idea that \gls{rl} can be viewed as an online search where an agent explores unknown environments instead of a simulated model, extend blind exploration (like in blind search e.g. Dijkstra's algorithm) with two cooperative mechanisms:  \gls{lec} and \gls{lbw}.
In the work by \citet{maclin1994incorporating}, 
an \emph{advice-giver} watches the learner and occasionally makes suggestions to help it, expressed as instructions in a simple programming language (e.g., rules to avoid an enemy). These pieces of advice are integrated and refined by \gls{rl}.

As one of the most recognized examples of \gls{rlwhil} approaches, \citet{isbell2001cobot} 
present Cobot, an application of \gls{rl} for LambdaMOO, a complex, open-ended, multi-user chat environment that can take proactive actions and adapt the behavior from multiple sources of human reward. After 5 months of crowd-sourced training, and 3171 reward and punishment events from 254 different users, Cobot learns nontrivial preferences for a number of users. 

\section{Reinforcement Learning with Human-in-the-Loop approaches}
\begin{figure}[h]
   \begin{minipage}{0.5\textwidth}
     \centering
     \includegraphics[width=1\linewidth]{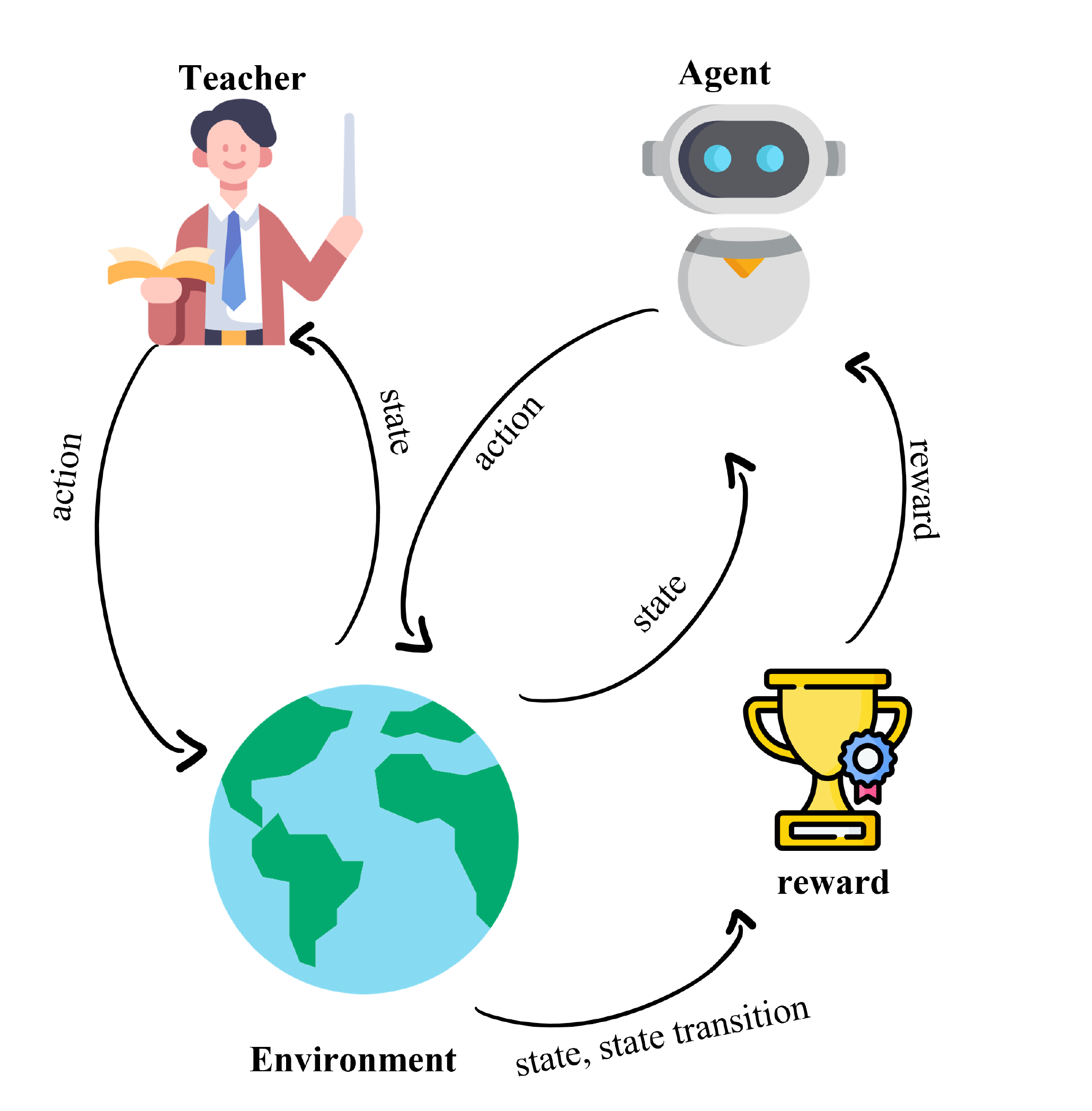}
     \caption{\\ Human-guided exploration}\label{fig:human_exploration}
   \end{minipage}\hfill
   \begin{minipage}{0.5\textwidth}
     \centering
     \includegraphics[width=1\linewidth]{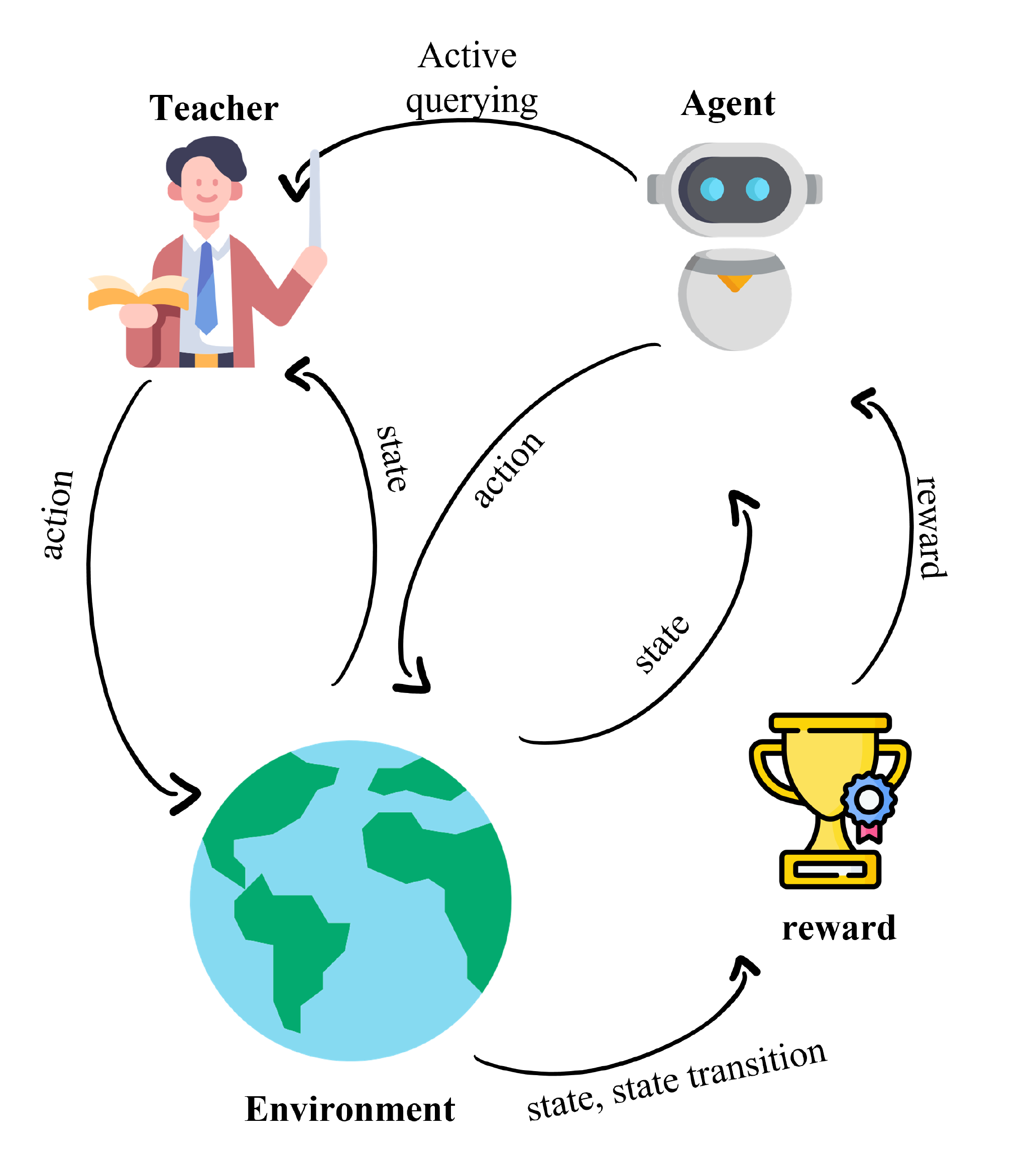}
     \caption{\\ Human intervention for Safe-RL}\label{fig:human_intervantions}
     \end{minipage}
\end{figure}

\begin{figure}[h]
   \begin{minipage}{0.5\textwidth}
     \centering
     \includegraphics[width=1\linewidth]{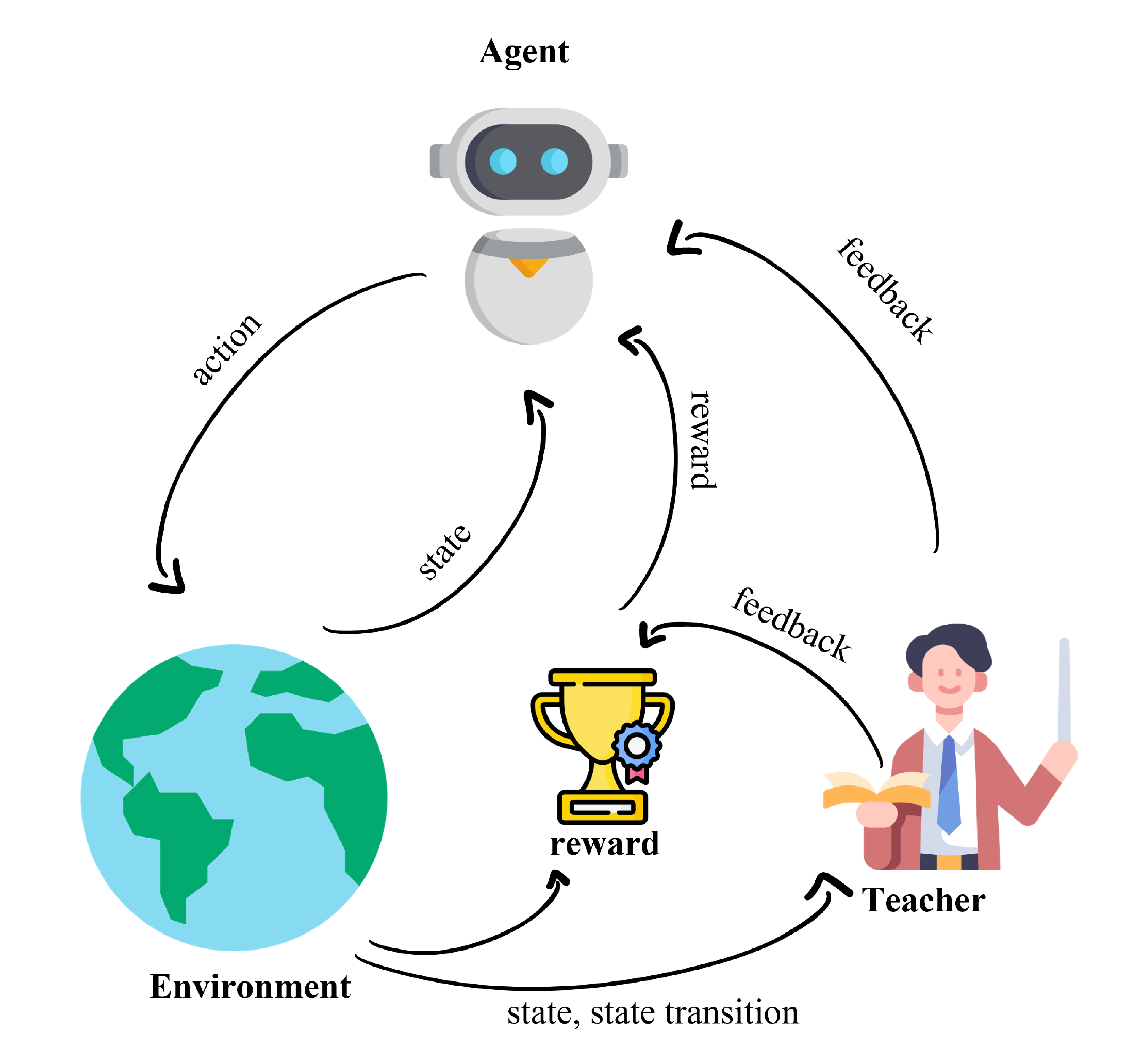}
     \caption{\\ Reward Shaping}\label{fig:reward_shaping}
   \end{minipage}\hfill
   \begin{minipage}{0.5\textwidth}
     \centering
     \includegraphics[width=1\linewidth]{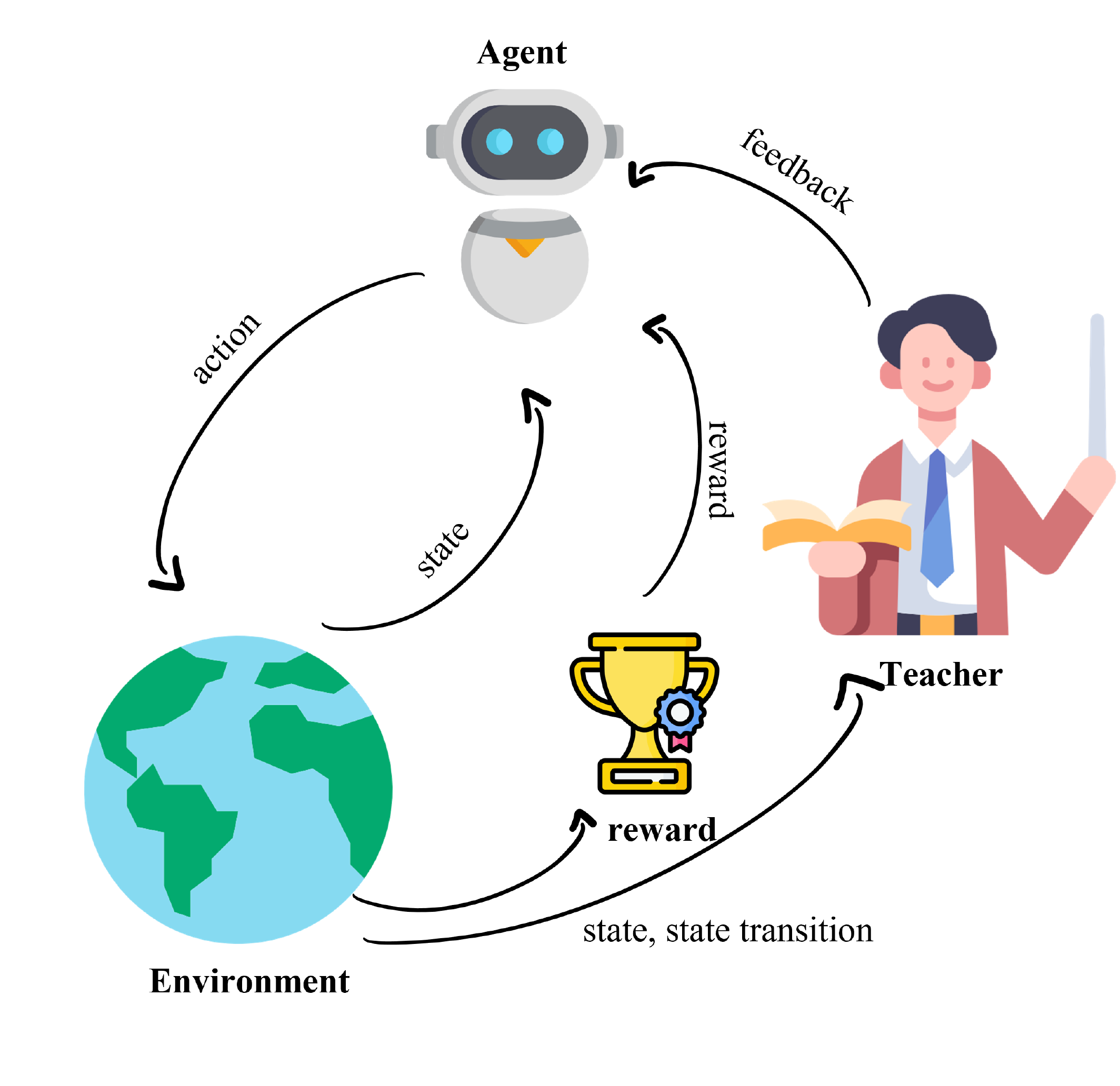}
     \caption{\\ Policy Shaping}\label{fig:policy_shaping}
     \end{minipage}
\end{figure}
A few surveys on the topic of \gls{rlwhil} have been recently published, providing a deeper insight into the field \citep{zhang2019leveraging,arzatecruz2020survey,lin2020review,najar2021reinforcement}.
Although there is no consensus, one way of clustering these approaches is based on the way human feedback is integrated into the learning loop (e.g. as a reward, as a policy correction, as a guiding policy for exploration, etc.). Based on that, \gls{rlwhil} approaches can be clustered into four main categories:

\begin{itemize}    
    \item Human-guided exploration, see Fig. \ref{fig:human_exploration}
    \item Human intervention for Safe-RL, see Fig. \ref{fig:human_intervantions}
    \item Reward shaping with Human feedback, see Fig \ref{fig:reward_shaping}
    \item Policy shaping, see Fig. \ref{fig:policy_shaping}
\end{itemize}
which we briefly summarize here.

\subsection{Human-guided exploration}

As known from many \gls{rl} applications, training \gls{rl} agents requires extensive trial-and-error experience. A necessary component for improving the policy is \emph{exploration}. To accumulate high returns, the agent needs to follow actions that were leading to high returns in the past. On the other hand, the agent sometimes needs to try-out actions that it did not try before. This, ideally, allows it to find strategies that obtain higher returns and, therefore, improve the policy. 
There are several environments that are particularly hard for learning from the perspective of exploration, mainly \emph{sparse-reward} environments and environments containing \emph{fatal failures}.

Sparse reward functions normally return a constant signal for most of the states, while they provide a different value in a few states, normally for indicating success and/or failure. One example of such an environment is a puzzle, where the agent receives a reward only once the puzzle is successfully solved. Obtaining useful information from random exploration in sparse reward settings makes learning challenging. Therefore, utilizing human feedback for guiding the agent during the exploration phase has been proposed to avoid this problem.   

In \citet{suay2011effect}, a human provides rewards but also anticipatory guidance on the selection of future actions. 
The users can guide during a short period before actions are selected by the robot, such that they limit the set of actions that can be explored in a specific state.
This can be interpreted as directing the robot's attention towards the target object or area.
\citet{subramanian2016exploration} present a model-free policy-based approach called \gls{efd}. The user feedback is used to shape an exploratory policy and bias a \gls{rl} agent’s exploration to cover the search space effectively.
The algorithm can query for demonstrations from the user by highlighting the states in a \gls{gui}. The user can help the learner by providing demonstrations (reaching the states) using the \gls{gui}. The user can alternatively choose to ignore the query or stop interacting with the algorithm. These demonstrations are used to learn an exploration policy to reach desired states. 
In some applications (e.g., letter writing), it was shown to be useful that users can select some desired regions to direct exploration of the learning algorithm.
\citet{schroecker2016directing} enable to interactively add \emph{via-points} that restrict the search space to trajectories that will be close to the defined via-points at the specific time. This in turn significantly reduces the number of samples necessary to learn a good policy.
%
Besides approaches where the teacher is directly guiding an exploration policy, there are also approaches where it is done in an indirect way, such as using gaze or natural language.
\citet{saran2021efficiently} propose an approach that utilizes indirect information gained from a human observer by detecting the human gaze, while the agent is interacting with the environment. The algorithm guides the agent to explore the regions that are targeted by the human gaze.
Human corrections can also be considered as the exploration disturbances of a stochastic policy in \acrlong{ps} approaches \citep{celemin2019reinforcement,celemin2019fast}, or as guidance in which the \gls{rl} agent memorizes the advice on where to explore \citep{scholten2019deep}.

\subsection{Human Intervention for Safe-RL}

There are environments in which the safety of the system is not guaranteed for all possible transitions in the state space. Learning policies in such scenarios with a high risk of failure poses a difficult challenge, especially for executing exploratory actions. Such problems are the focus of the research of \gls{saferl}. 
As seen in a \gls{saferl} survey \citep{garcia2015comprehensive}, one approach to \gls{saferl} is to use external knowledge in a form of teacher advice, e.g., the learner agent asks for advice when the confidence level is low \citep{utgoff1991two}.
%
Safety can be guaranteed using shields \citep{alshiekh2018safe}, which are models that prohibit actions leading to catastrophic failures. However, shields are usually conservative and might limit exploration. Designing a shield that is not conservative but still safe is rather challenging. 
Some approaches utilize \emph{teacher-oversight} to interactively learn a model of a safety shield \citep{marta2021humanfeedback}. 
A similar approach (although not based on standard \gls{rl}) is presented by \citet{kahn2021land}, where a model of human intervention probability is learned from human interventions. It is later used in a model predictive control fashion to find actions that have a lower probability of human intervention. 

\subsection{Reward Shaping with human feedback}

Besides the exploration process, human teachers can provide insights into the long-term benefit of an action. This can be done in the form of a reward signal in addition to the one coming from the environment. The idea of \emph{reward shaping} consists of designing a reward function that not only defines the main objective but also provides useful guidance for effective exploration~\citep{ng1999policy}.
However, designing a reward function in advance that balances the goal definition and the guidance task is often very difficult, and the agent might get stuck in a local maximum, e.g., exploiting the local reward without reaching the goal, also known as \emph{reward hacking} \citep{amodei2016concrete}. 
The human-provided reward can be also considered as a type of reward shaping. It can be used to guide the agent to learn the task faster.
As it is provided interactively, human feedback can be even used to correct the negative effects of reward hacking. If the user notices the agent is exploiting the reward used for guidance without progressing on task execution, it can locally correct it by adding a negative reward.

One of the first approaches of this family is \gls{interactive rl}, as presented by \citet{thomaz2005real}. A human teacher can, in real-time, provide a reward signal in addition to the reward provided by the environment.
%
\gls{tamer} \citep{knox2008tamer}, is also used in an extension with \gls{rl}, where \gls{tamer} is used first, and then the policy is fine-tuned with \gls{rl} \citep{knox2010combining,knox2012reinforcement}.
Similarly, \citet{xiao2020fresh} present \gls{fresh}, an approach where a human is presented with trajectories from a replay buffer and then provides feedback on actions (good or bad) or even on states. A \gls{nn} is then trained to generalize this feedback to unseen states and actions, and the feedback of the \gls{nn} is converted to shaping a reward that augments the reward provided by the environment.
Recently, \citet{arakawa2018dqn} introduce an algorithm called DQN-TAMER as a combination of \gls{tamer} and DQN. The method learns two Q functions in parallel, one from the environment reward and one from the user feedback. Then, the action is computed as the action that maximizes the combination of both functions. This approach is not directly Reward Shaping but rather Value Shaping.
\subsection{Policy shaping} 
There are also works combining \gls{rl} methods and teachers in the learning loop for directly obtaining an explicit policy representation. 
This has been studied especially from the perspective of using absolute corrective feedback as in \gls{dagger} algorithms.
Methods like \gls{aggrevate} \citep{ross2014reinforcement} extend the incremental collection of demonstrations with the use of cost (reward) functions provided by the environment, which are considered for improving the policy on top of the information provided by the teacher.
\citet{chang2015learning}, propose a similar learning scheme, although their paper is focused on the problem of learning from a \emph{poorly performing reference policy} (teacher), showing that the feedback of the environment helps to improve the knowledge obtained from the teacher.
Later on, \citet{sun2017deeply} extended  \gls{aggrevate} for using it with differentiable policies, such that it is possible to use powerful expressive models (e.g., deep \glspl{nn}).

This line of research intends to leverage the benefits of both \gls{il} and \gls{rl} worlds, similarly to the domains of other Sections of this Chapter.
In order to be able to obtain policies outperforming the teacher, it can be more effective to optimize the environment objective function based on the demonstrated trajectories, instead of optimizing the discrepancy between the learner and the teacher. 
Therefore, such a strategy can relax the requirement of having expert or near-optimal demonstrators in the learning loop.



\section{Discussion}
 
This chapter provides a concise, non-exhaustive overview of the \gls{rlwhil} field from the \gls{rl} perspective. Human-in-the-loop approaches have a long history in the field of \gls{rl}, and in many applications, they are critical for efficient and safe learning. From \gls{iil} perspective, they offer an approach to overcome suboptimal human feedback. The approaches can be analyzed based on the way the human input is integrated into the learning loop, making the distinction between approaches improving exploration (efficiency and safety), approaches modifying the reward and approaches directly improving the policy.

\newpage
\chapter{Interfaces}\label{section:interfaces}


\begin{figure}
    \centering
    \includegraphics[width= 0.8\linewidth]{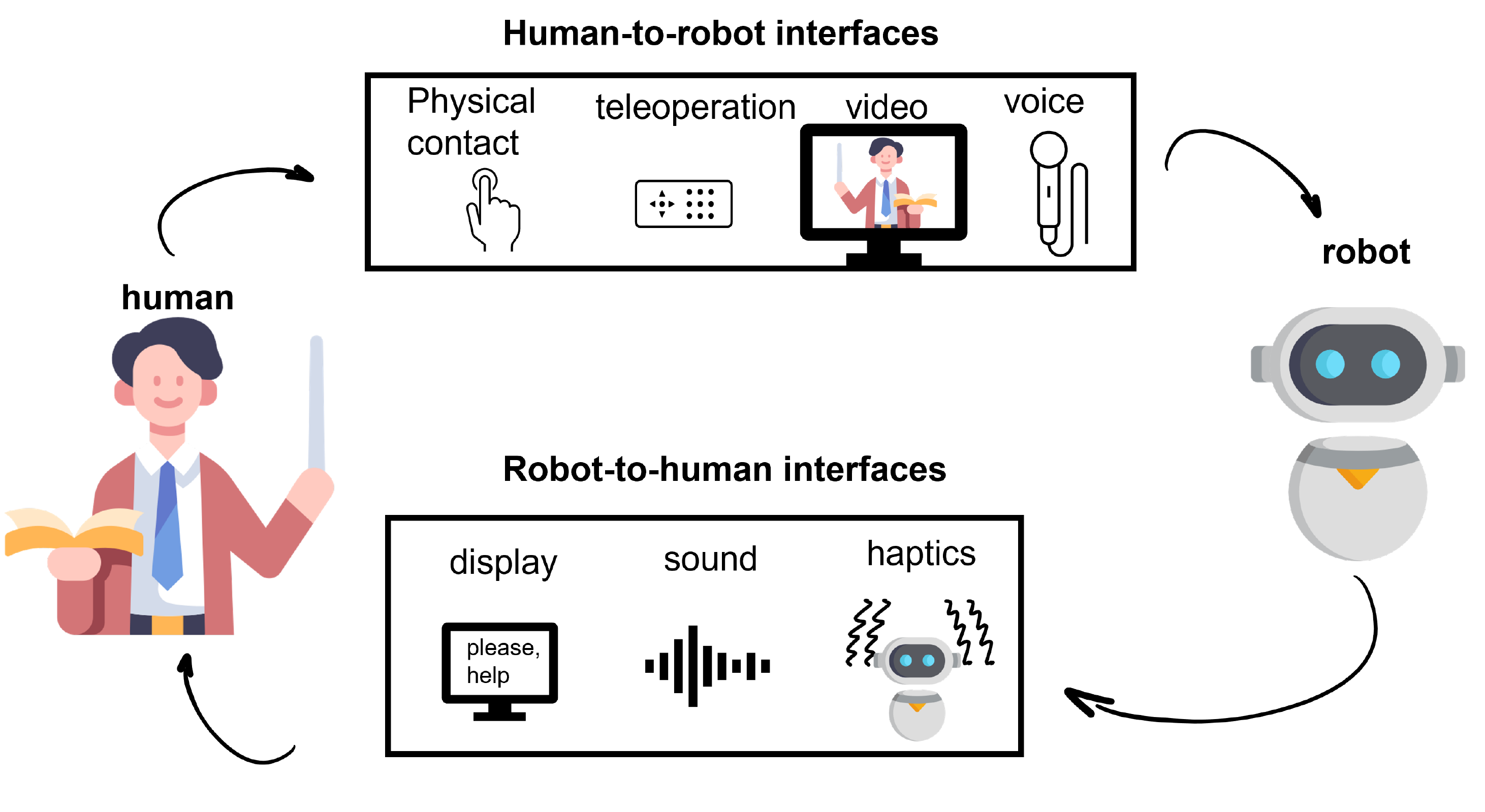}
    \caption{Human-to-robot interfaces and robot-to-human interfaces commonly employed in \gls{iil}.}
    \label{fig:interfaces}
\end{figure}
In an interactive learning setup, the human teacher and the robot are frequently exchanging information.
It is natural for humans to communicate in many different forms, making discussions about \emph{how} and \emph{what} to communicate with the robot to be often overlooked.
However, from the robot's perspective, the type, the format, and the content of the received information can greatly impact its learning process.
Furthermore, choosing an adequate way for the robot to communicate with the human can improve the latter's understanding of the system, which will in turn improve the quality of information the teacher provides to the robot.

While Chapter~\ref{sec:modes} discusses the types of feedback that the teacher can provide, this chapter focuses on the channel used for exchanging this information. 
According to \citet{dudley2018review}, an interface is the bridge in charge of the bidirectional feedback between the user and the system, as illustrated by Figure~\ref{fig:interfaces}.
Here, we define interfaces as the physical channel used to capture/provide data together with required software that processes the raw data into the desirable information type.

First, we summarise the different interfaces used by humans to provide feedback, highlighting which type of information is more appropriate to provide using each interface, and the respective learning methods that can efficiently explore that information.
Second, we review interfaces used by the robot to provide information to the human, highlighting which information is provided back to the human and how it is used in the learning loop.

\section{Human-to-Robot Interfaces}\label{sec:human2robot interface}
In \gls{iil}, most methods are based on passive learners, which means the interactions are only sent from the teachers towards the agent, without having active responses or queries from the robot. The users transfer their knowledge to the learner through the feedback signals using different kinds of interfaces, as discussed in this Section.

\subsection{Physical Contact with the Robot Embodiment}\label{subsec: kinesthetic teaching}

Being able to teach the robot by physically moving it is a promising method to teach generic tasks since it does not require any extra interface with the user \citep{amor2009kinesthetic}. This approach is generally called kinesthetic teaching.
Kinesthetic teaching is especially motivated by its simplicity, which is a key feature to enable the employment of robots by non-expert users in everyday tasks.

Teaching trajectories is arguably the most common use of kinesthetic input, which allows the user to provide demonstrations in a natural manner \citep{wrede2013user}.
For example, kinesthetic information can be used to learn user preferences \citep{jain2013learning,canal2016personalization,bajcsy2017learning,bajcsy2018learning,losey2022physical} such as distance to objects for safer manipulation.
\gls{lira} \citep{franzese2021learning} shows how it is possible to solve ambiguities from kinesthetic demonstrations; when the learning agent recognizes an ambiguous situation, it requests the user to move the end-effector towards the correct action, allowing it to learn the correct one. 
\gls{ferl}~\citep{bobu2022inducing} presents a method to learn non-linear features from kinesthetic demonstrations of partial-trajectories. Such features allow for improving learning efficiency and generalization capabilities.

Kinesthetic teaching has been successfully applied to robotic arms; however, such an interface might not be feasible or safe in other scenarios.
For example, the robot's size has to be within the human manipulability spectrum, making it not applicable for nano- or industrial-sized robots.
Also, high-speed robots render kinesthetic teaching infeasible and/or unsafe, especially on tasks requiring high-speed controlling (e.g., balancing tasks).
Due to such limitations, in some learning settings, other interfaces can be more adequate for teaching robots interactively.

\subsection{Physical Contact with External Device}\label{subsec: keys}

Keys are present in most computational units (e.g., keyboards, joysticks, remote controllers, touch pads) and are arguably the most standard way for users to input information into systems.
Key-based interfaces are convenient since they are readily available in most systems and can be mapped to provide different types of feedback to robotic systems.
Keys can be used to provide full demonstrations as shown by \cite{palan2019learning}.

Even though many \gls{iil} approaches use keys for their convenience to create proof-of-concept tests using simulations \citep{thomaz2007asymmetric,argall2011teacher,akrour2014programming,spencer2020learning,zhang2020deep,myers2021learning}, games \citep{knox2009interactively,subramanian2016exploration,christiano2017deep,warnell2018deep,xiao2020fresh,jauhri2021interactive}, or real robots \citep{argall2008learning,suay2011effect,knox2013training,kelly2019hg,hoque2021lazydagger}, there are three points one must consider about key-based inputs.

First, it is undesirable to have an interface with many keys available in \gls{iil}, complicating the usage of the system.
As such, keys are better used on systems with limited user input requirements, such as learning from human reinforcements (see Section \ref{subsub:HumanReinforcements}) or from human preferences (see Section \ref{subsec:learning from human preference}).

Second, the meaning of each key can be programmed at a software level, namely \emph{mapping the keys}.
Such mapping can be task-specific, which is common in cases the keys are used to provide feedback in the state/action domain (see Section~\ref{sub:state-actionDomain});
for example, \citet{ross2011reduction} associates keys to actions in a video game (left, right, jump, speed). 
However, such mappings can also be independent of the task, which is the case for learning from human reinforcements (see Section \ref{subsub:HumanReinforcements});
for example, \gls{tamer} \citep{knox2008tamer} expects binary feedback that indicated the optimality of the policy, making it task agnostic.

Furthermore, aiming to improve the inherently limited amount of information that keys can carry, \citet{loftin2016learning} propose to use a statistical user-feedback model, which allows extrapolating cases in which the user did not provide feedback.


An option to encode more information in a single input is to use analog sticks (joysticks), which allow mapping multiple values depending on how far the stick is displaced by the user. They are commonly used in driving scenarios due to their precision for controlling continuous values \citep{corrigan2016engagement}.
For instance, an analog stick is used by \citet{ross2011reduction} to provide demonstrations of the correct steering of a kart in a racing game to train with the \gls{dagger} framework.
A digital sliding bar is used by \citet{wilde2022learning} in order to provide an analog value for user preferences in a \gls{lfp} setup.

Despite the above considerations, 6 \glspl{dof} interfaces (e.g., space mouses, cellphones) allow the user to directly control the robot's end-effector position and orientation directly \citep{luo2021robust}, and to give continuous signals (as analog sticks).
Such interfaces allow interactive teaching in scenarios where the human cannot provide kinesthetic demonstrations. For example, \cite{celemin2019reinforcement} propose to use keys to train a robotic arm to perform the \emph{ball-in-a-cup} task using a combination of \gls{rl} and \gls{ps}.
Furthermore, keys and joysticks can also be used to teach the speed and shape of dynamic grasping movements as proposed by \cite{meszaros2022learning}.

As such, we can conclude that interfaces such as keys, joysticks, and 6 \gls{dof} mouses can be used not only for their commodity value but also for allowing \gls{iil} methods to be applied to systems that cannot accept other types of interactions (e.g., kinesthetic).
Furthermore, it is a common practice to implement modules that convert other types of sensors into discrete or analog signals, which can be used by learning methods in the same fashion as the ones presented in this section.

\subsection{Contact Free Interfaces}
The previous sections presented interfaces whose raw output is compatible with the type of information expected by the learning method.
However, interfaces such as video or voice require more complex processing for mapping from the communication channel to the learning method's input; also, such interfaces are interesting in the \gls{iil} context to enable natural communication between the human and the robot.

Normally, processing these signals involves different kinds of pattern recognition methods that extract the relevant information from the raw signals, which are then mapped into discrete or continuous signals.
Next, we summarize works that interface raw video, motion capture, depth sensors, and voice signals with \gls{iil} methods.

\paragraph{Video}
Using videos of humans performing tasks allows the user to naturally perform a task with minimal hardware requirements, and without directly dealing with the robotic system.
Nevertheless, learning from videos is not trivial given their large state spaces, making them inefficient to be trained based on human demonstrations.
Furthermore, videos from humans are also susceptible to the \emph{correspondence problem} \citep{nehaniv2002correspondence}, which is usually mitigated by using videos featuring only the user's hand, which is mapped to the end-effector's position.

Videos have been used in \gls{iil} through pre-trained models that extract the information of interest from the images.
This idea is applied to \gls{iil} by \citet{celemin2019interactive}, where a hand gesture recognition module is employed to provide feedback for the \gls{coach} framework, and by \citet{arakawa2018dqn}, who use a facial expression detection mapped to pre-defined positive/negative reward values to be used with the \gls{tamer} framework.

\paragraph{Depth Sensors and Motion Capture}

Depth sensors and motion capture provide information on depth, pose estimation, and motion tracking, which is often necessary to perform dynamic tasks.
Motion capture usually uses cameras or an array of inertial sensors \citep{hindle2021inertial} to provide tracking information of objects and the human body, to be used as a human-robot interface without directly manipulating the robot.

These systems have been successfully used to track the hand of the user, which is mapped to the robot's end-effector.
As such, the user can perform a task naturally, without handling the robot, which has been shown as preferable by the users in a \gls{lfd} setup~\citep{hedlund2021effects}.
Motion capture has also been used by \cite{leon2011teaching} to record demonstrations of the position of objects manipulated directly by the human teacher, which allows obtaining a first policy that is later on refined interactively.
For instance, a Microsoft Kinect V2 has been used by \citet{najar2016training} to create a discrete signal from head movements (nodding and shaking), which are mapped as positive or negative feedback to the learning agent, similarly to the pressing of a key.

The \gls{tics} framework \citep{najar2020interactively} explores learning the mapping between movements captured with a Kinect and a discrete set of pre-defined actions.
This model helps the overall training process, which in turn refines the mapping in an iterative process. This makes the system robust to different users since each user associates a different meaning to the actions.
A similar idea was previously presented by \citet{grizou2013robot}, where a robot learns to perform a task and the association of pre-defined commands identified through a voice recognition module.


\paragraph{Voice}
A direct way to create a voice interface is to use pre-established commands, simplifying its recognition, where specific voice commands are used as a feedback signal (e.g., good, or bad), or to trigger specific operation modes \cite{kaplan2002robotic,nicolescu2003natural,leon2011teaching}.
Also, \citet{tenorio2010dynamic} propose to parse voice commands with a language recognition module to identify a pre-established vocabulary where all words have an associated reward value which is used to compose a final evaluative feedback reward; creating an analog input (discretized according to the vocabulary) to the learner.

However, such a simplistic approach cannot capture the information richness of human speech, leading to the adaption of learning-based natural language modules.
Focusing on language grounding, \cite{blumberg2002integrated} propose an interactive click trainer which learns to associate voice commands with the actions of a simulated dog. Similarly, \cite{macglashan2014training} propose a method that grounds text commands.
Voice commands have also been used in a simulated environment by \citet{cruz2015interactive}, where language commands are processed by an \gls{asr} module that interactively teaches affordances. Similarly, \citet{krening2016learning} parse natural language sentences into \emph{advice} or \emph{warnings} analysing sentiment. Finally, \citet{cruz2018multi} integrate motion capture commands and voice commands using a multi-modal estimation module to teach affordances.

\subsection{Interactions with Multiple Interfaces}
Works leveraging multiple types of human input are common in the \gls{iil} literature, which is motivated by two advantages.
Firstly, each type of human-robot interaction is more adequate to capture different stages of the learning process.
Secondly, different types of information are used during the learning process, which can achieve a faster or more consistent learning rate~\citep{koert2020multi}.


As an example of the first advantage, within \gls{iil} it is common to obtain an initial policy from demonstrations, which is then refined using interactive corrections.
For example, \cite{prakash2020exploring} learns an initial policy from demonstrations in a driving simulation, which is time-consuming but only requires recorded demonstrations, then the policy is interactively refined by the user.
\citet{franzese2021ilosa} initiate a policy using kinesthetic demonstrations, which is later refined through teleoperated corrections in order to learn the movement stiffness, allowing for performing force-interaction tasks without the necessity of force sensors.
This idea is also applied by \citet{celemin2019reinforcement}, where a policy is initialized using kinesthetic demonstrations and is interactively improved using a combination of corrections and \gls{rl}.


As for the second advantage, its motivation lies in the fact that each learning modality has its own shortcoming; thus, using complementary methods might allow overcoming the limitations of single interfaces.
This idea has been explored by \citet{jain2015learning}, where humans provide feedback first using keys to rank policies displayed in a simulation environment, and then provide kinesthetic corrections to the trajectories; the first feedback form is easier to provide and allows teaching simpler trajectories, while the second allows teaching more difficult parts of tasks.
Also, in the \gls{ceiling} framework \citep{chisari2022correct}, an initial policy (obtained from demonstrations) is interactively improved using evaluative feedback from key presses, and corrective feedback from teleoperation (joystick); the latter allows corrections, while the former allows the user to provide feedback in difficult situations in which the user could not correct the robot.

\section{Robot-to-Human Interfaces}\label{sec:robot2human interface}

Even though most works in \gls{iil} focus on how to capture information and learn from humans, some work has been paying attention to the opposite communication direction.
This channel allows the robot to communicate the necessity of corrections or input, as in the case of robot-gated or active learning methods, in which the user is asked for feedback only in specific cases which improve the learning process (e.g.,  \cite{sadigh2017active,cui2018active,brown2018risk,biyik2018batch,biyik2020asking,hoque2021thriftydagger,franzese2021learning}).
Additionally,~\citet{li2016using} and~\citet{koert2020multi} show that providing the human with uncertainty and performance information about the robot's actions can improve the teacher's feedback quality, improving the learning process, i.e., the teacher concurrently learning how to teach the robot.
Thus, understanding how and what information to communicate to the human can be a key to enabling non-experts to interactively teach robots.

\paragraph{Displays}

Using the computer screen is the most common way for the robot to communicate with the human.
Screens are suitable for displaying simulations or behavior estimates that provide insights to the user, allowing them to preview the robot's behavior and act accordingly.

The displaying of simulations has been used by \gls{lfp} methods, in which a few simulation trajectories are displayed to the user, who selects the most adequate ones according to personal preferences \citep{jain2015learning}. Furthermore, these methods can enable learning from multiple teachers \citep{wilde2020improving}.

The video interface is also part of \gls{vr} kits, which usually consist of a wearable headset that streams a camera image to the user and a joystick/controller, providing standard keys and/or analog sticks for controlling the robot.
For example, \citet{delpreto2020helping} use a \gls{vr} kit to teach grasping tasks using a master-apprentice model.


\paragraph{Voice}

Using verbalization of pre-defined sentences is a straightforward way to create a communication interface from a robot to a human teacher, and it has been used in \gls{iil} by \citet{maeda2017active}, whose active learning method asks for demonstrations of reaching objects specified a priori based on uncertainty metrics.

Nevertheless, learning what needs to be communicated with the user in \gls{iil} setups is subject to recent works.
For example, \citet{shridhar2020ingress} propose to interactively learn a model for composing questions that help to disambiguate the selection of objects for manipulation tasks.

\section{Interface Design}\label{sec:InterfaceDesign}
We should consider three main aspects to successfully complete an \gls{iil} interfacing system that connects both an agent and a human: 1) the hardware interface (discussed in this Chapter), 2) the modalities of interaction (Chapter~\ref{sec:modes}), and 3) also the user experience (Chapter~\ref{sec: user experience}).
Interface design in \gls{iil} has been a less explored topic until today. However, works in \gls{iml} already identified the main factors influencing these aspects. 

First, both the needs of users and the model should be taken into account in interface design, user studies can reveal false assumptions or user patterns and difficulties, which can be used to leverage the system usability and efficiency \citep{amershi2014power}.

Second, the preferences of the users on models, features, and interaction modes should be considered, e.g., to allow non-expert users to build an accurate mental model of the system, hence the best feedback possible \citep{chatzimparmpas2020survey,dudley2018review,mohseni2019multidisciplinary}.

Third, excessive querying should be avoided in order to reduce the undesired cognitive load of the user \citep{amershi2014power}. Instead, querying should be done to promote understanding and encourage trust in the model, helping the user to understand the source of failures and solve them.

Finally, user studies can help to develop systems in which experts and non-experts are capable of understanding what's intended from the interaction with the agent  \citep{dudley2018review}.
For example, \citet{chatzimparmpas2020survey} show that users tend to better interpret \gls{ml} models through data visualizations.

\section{Discussion}

Since the human is the source of information in \gls{iil} systems, special attention should be paid to the interface design for these systems.
Furthermore, robot-gated methods (Chapter \ref{subsec:AbsoluteCorrections}) can interface with the human, in order to teach them how to provide better feedback, improving the overall learning process.

Nevertheless, interfacing sensors with learning methods cannot always be done directly, especially on high-dimensional problems, such as learning from video and voice, since it is required to map this sensor information into feedback signals that have to be compatible with the already established learning methods.
However, using pre-defined, or pre-trained mappings implies that the information provided by the teacher is not fully explored (as these interfaces are not adaptable), leaving open the challenge of interactively learning directly from such signals.
Note that existing research outside the \gls{iil} scope focuses on learning from natural language sentences \citep{williams2018learning}, or on using videos of demonstrations \citep{yang2019learning}, and it can be used as a base for extending interactive methods.
Other methods in the \gls{iil} literature focus on merging different types of signals using multi-modal learning methods, e.g., voice and video are used by \citet{jang2022bc} as input. Such models could pave the way to create interfaces in future works.

Finally, despite the interface and modality of interaction (Section~\ref{sec:modes}) being related to the information used for teaching, the human aspect has also to be taken into consideration in order to achieve an efficient interactive learning setup.




%

\newpage
\chapter{User Studies in IIL}\label{sec: user experience}

A major factor contributing to effective and efficient collaboration between robots and humans is the improvement of the human experience in teaching and collaboration. The topic of this chapter pertains to user studies in \gls{iil}. We report central guidelines on how to design user studies in \gls{iil}, based on findings of several papers on user studies in \gls{iml} and \gls{iil} specifically. This Chapter discusses how to set up a user study and what evaluation metrics to use.

\section{Study Setup}
A user study targets comprehending user needs, behaviors, and motivations through a variety of evaluation methods such as surveys, interviews, and questionnaires, in order to improve the user experience in a certain interaction scenario. User studies require previous approval from an ethics committee. Researchers should communicate to the ethics committee any potential harm that may be caused to the participants, as well as actual or potential conflicts of interest \citep{kuniavsky2003observing}.

To effectively set up a user study in \gls{iil}, a researcher should take into consideration system aspects as well as human aspects. 
Although there is flexibility in setting up a user study, researchers should take into consideration the following aspects:
\begin{itemize}
    \item What is the study goal?
    \item What are the most significant metrics to support the study goal, and how to collect data related to those metrics?
    \item Select participants according to the study goals. How many participants? Experts or non-experts? Do they need to get familiar with the task (s)? How to design the familiarisation phase?
\end{itemize}
These items are discussed in the following Sections.

\subsection{Study Goal}
The study goal is defined according to the motivation of the study.
Some user studies aim to verify if non-experts can teach complex tasks to an agent. These studies make use of different metrics. For example, \citep{meszaros2022learning,franzese2021ilosa,perez2020interactive,perez2019continuous} apply metrics such as success rate and time of completion to evaluate teaching efficiency in manipulation and navigation tasks. 
\citet{sadigh2017active, bajcsy2017learning, bajcsy2018learning, losey2022physical, biyik2018batch, jain2015learning,chu2015exploring,cakmak2012designing} use model training error and subjective evaluation metrics (user's perception of tasks and policies) to measure efficacy and efficiency in teaching successful robot behaviors. 

Another goal is to compare different learning methods using non-expert users. These studies measure the learning efficiency of an \gls{iil} method and its relation to human performance. \citet{jauhri2021interactive, delpreto2020helping, biyik2020asking, cui2019uncertainty, palan2019learning,  chisari2022correct,  he2020learning, hoque2021lazydagger} employ robot learning metrics (task accuracy, success rate, training time, reward maximization) as well as human performance metrics (workload assessment and model perception) to compare different learning methods.

Some studies also aim to analyze how the teaching strategy used by the users influences the results. For example, \citet{loftin2016learning} conduct a user study to determine the rate of success of teaching behaviors such as punishment-focused, reward-focused or balanced. \citet{vollmer2018user} evaluate different teaching strategies such as comparative, error based, or spontaneous. These strategies are rated using ground truth data and its effect on task success or failure. 

\subsection{Participants}
Human-centered evaluations require objective profiling of the participants. Users' expertise in \gls{ml}, \gls{iil} and users' age and gender are aspects that may influence interaction with the learning method/task. Hence, the outcomes of user studies may vary due to the fact that different users may show differentiated interaction performance profiles, for a similar interaction method/task.

\subsubsection{Participants' Expertise} 
There are two kinds of participant expertise to be considered, expertise in the task domain, and expertise in robotics and \gls{ml}. Users' expertise can be categorized into high, medium, and low. For example, participants with high expertise in \gls{ml} tend to make predictions faster about how the system learns compared to participants with low expertise. Therefore, a proper selection of study participants should be taken into consideration. Most of the user studies in \gls{iil} either report very low \gls{ml} expertise or none \citep{biyik2020asking,palan2019learning,chisari2022correct,he2020learning,thomaz2006reinforcement,franzese2021ilosa,meszaros2022learning,perez2019continuous,jauhri2021interactive,delpreto2020helping,bajcsy2018learning,bajcsy2017learning}.
On the other hand, \citet{jain2015learning,bajcsy2017learning,palan2019learning} require that participants have medium task domain expertise in robot manipulation tasks. Finally, high domain expertise is requested in a driving task (the participants needed to own a driving license and in some cases 8 years of experience driving cars) \citep{cui2019uncertainty,sadigh2017active}. 

\subsubsection{Age and Gender} 
Participants' age, gender, demographic categories and distribution should be taken into account to structure user studies. The most targeted age group in user studies lies between 18-37 years, with some studies comprising older participants. Gender specifications of the participants are only reported in a few studies \citep{bajcsy2017learning,palan2019learning,loftin2016learning}; however, no correlations are found between gender and the interactive process outcomes.

\subsection{Number of Participants} 
The number of participants to be recruited for a study depends on the study goal. For example, pilot studies whose goal is to evaluate the feasibility of an approach to be used in a larger scale study include small samples, while large-scale studies include larger samples. The number of participants in a user study is also limited by the experimental setups. The setups with real robots, controllers, and interfaces require preparation, calibration, and resetting in each trail which constrains the size of a user study. We observe user studies with a real-world experimental setup to include, on average, 7-15 participants. 
However, simulated or grid-based high-level tasks offer the opportunity for larger-scale user studies, with, for instance, 40 \citep{he2020learning} or 150 \citep{loftin2016learning} participants.

\subsection{Preparation Phase}
The preparation phase identifies the type and amount of training provided to participants before they interact with the system being evaluated. The aim of a preparation phase is to avoid participant interaction with the experimental setup to be influenced by confounding effects, such as lack of clarity of task requirements and task execution, or insufficient familiarity with the study interface. The training phase needs to balance the amount of information provided to participants against the potential introduction of bias towards a system, technique, or model. 
Participants that do not need to know the robotics hardware details or controller interfaces are just provided with a brief session about the tasks and related high-level instructions
\citep{delpreto2020helping,hoque2021lazydagger,loftin2016learning,biyik2020asking,he2020learning,perez2019continuous,biyik2018batch,sadigh2017active,chu2015exploring,vollmer2018user}. On the other hand, when participants need prior knowledge in robotics (hardware details, controller interface), a more dedicated session for familiarization is scheduled before the main study \citep{meszaros2022learning, franzese2021ilosa, chisari2022correct, bajcsy2017learning,bajcsy2018learning,palan2019learning,jain2015learning}.

\section{Evaluation Methods}
In this Section, we describe evaluation methods and metrics that can be adopted in \gls{iil}. These include, \emph{robot learning performance metrics} and \emph{human performance metrics}. 

\subsection{Robot Learning Performance Metrics}
\gls{iil} makes use of quantitative metrics to evaluate robot learning performance. Deciding on how to make use of these metrics depends on the goal and objective of the learning model. Common metrics to evaluate robot learning performance in \gls{iil} are listed below.
  
\subsubsection{Cumulative Reward} 
A widely used metric to evaluate robot task performance in \gls{iil} is the cumulative reward during task execution. In the \gls{iil} case, the reward function would need to be additionally designed by the creator of the user study for a given task, for evaluation purposes. This reward is also communicated to the study participants so that they are consistent with the true reward for the task during their interactions.
Examples of tasks that make use of the cumulative reward metric include manipulation reaching task \citep{cheng2018fast,perez2019continuous,perez2020interactive}, writing symbols \citep{schroecker2016directing,jauhri2021interactive}, autonomous driving \citep{menda2017dropoutdagger}, UAV racing task \citep{li2018oil}, balancing task on Cartpole \citep{wilson2012bayesian,knox2012reinforcement,vien2012reinforcement}, games like atari, pacman, frogger \citep{christiano2017deep, cederborg2015policy, subramanian2016exploration}, and OpenAI Mujoco based locomotions \citep{reddy2019sqil, hoque2021lazydagger, cronrath2018bagger}.

\subsubsection{Success Rate} 
Success rate evaluates the robustness of a learning method in performing a task. This metric is defined by calculating the ratio between the number of successful executions of the task and the total number of attempts. This metric is more useful for tasks in which success is binary (i.e, completed or not) and it requires low design effort (in contrast to the reward function). It is calculated among several trails/episodes and it is used for different tasks such as grasping  \citep{delpreto2020helping}, contact-rich manipulation \citep{mandlekar2020human, chisari2022correct, ablett2020fighting}, reaching manipulation \citep{perez2020interactive}, autonomous driving \citep{cui2019uncertainty,prakash2020exploring}, mobile robot navigating to a target \citep{argall2008learning}, drone navigation to reach a goal with tolerance \citep{blukis2018following}, drone perching \citep{goecks2019efficiently}, game of maze \citep{le2018hierarchical}, and industrial manipulation \citep{luo2021robust,hoque2021thriftydagger}.
    
\subsubsection{Model Training Error} 
This metric measures the accuracy of a model in fitting the observed data, and it is useful to evaluate or debug the model, and also observe how good it resembles the data.
This metric is used in tasks such as autonomous driving \citep{biyik2018batch, sadigh2017active}, walking \citep{akrour2014programming}, and reaching \citep{bajcsy2018learning,losey2022physical,biyik2020asking, palan2019learning}.
Furthermore, this metric is also used for performance evaluation in reconstructing trajectories for a task of drawing symbols \citep{celemin2019reinforcement}, and driving a mobile robot \citep{spencer2020learning,kelly2019hg}. 
    
\subsubsection{Task Completion Time} 
This metric represents the total time taken by the algorithm to learn to solve the task. Examples on studies that have used this metric include item sorting, where human guidance is used to help the agent learn the task quickly \citep{suay2011effect}, mobile robot navigation task \citep{tenorio2010dynamic, macglashan2017interactive}, game of soccer \citep{mericcli2010complementary}, insertion time in industrial assembly tasks \citep{luo2021robust}, high-level grid-based tasks \citep{peng2016need}, and constrained manipulation \citep{hoque2021thriftydagger}. Some authors use switching time to evaluate the learning algorithm performance \citep{hoque2021lazydagger}. 
    
\subsubsection{Safety Performance} 
The safety of an agent depends on the combined performance of the robot and the human, e.g. when learning, an autonomous driving car may drive off the road. This metric can be measured by the uncertainty of the agent's actions. The higher the uncertainty, the lesser the safety. \citet{menda2017dropoutdagger, menda2019ensembledagger} use safety performance evaluations by calculating the average reward resulting from the combined robot and human actions.
     
\subsubsection{Task Specific Metrics} 
Specific metrics evaluate task performance in a certain domain. In a manipulation task, \emph{number of knocked items} is an effective way to measure the accuracy of the task by observing how many objects other than the target object are displaced or knocked over by the agent/robot, e.g., in a constrained manipulation task with cluttered objects \citep{laskey2016shiv}.
In an autonomous driving application, \emph{infraction rate} is a leading performance metric. Infractions refer to breaking the driving rules of traffic (e.g., collisions, intersections with the opposite lane, driving onto the curb etc.), leading to potential collisions and human injuries. \citet{cui2019uncertainty} and \citet{prakash2020exploring} used infraction rate metrics to derive the quality of autonomous driving performance. The distance traveled by the autonomous vehicle between infractions has been used as another performance indicator also \citep{cui2019uncertainty, kahn2021land}.
For path following tasks, \emph{mean path deviation} measures the average error of the robot over multiple trials, e.g. lane deviations in autonomous driving. \citet{bajcsy2018learning} use this metric in a task of object placement with human preference. The evaluation focuses on observing whether the robot follows the human preferred path and the mean deviation resultant from it. \citet{kelly2019hg} and \citet{ross2011reduction} measure the mean deviation from the lane per meter/lap in autonomous driving. \citet{bootsma2021interactive} measure the root mean square error of deviation from the desired path for mobile robot navigation.

\subsection{Human Performance Metrics}
Human Performance Metrics evaluate different human performance capabilities \citep{abdel2005human}. According to \citet{sperrle2021survey}, there is no established methodology to evaluate \gls{hcml} systems due to the field's novelty. The authors emphasize that human factors such as effort and trust (cognitive and emotional elements), as well as placing humans as actors in the design of \gls{ml} models need to be taken into consideration in order to improve evaluations of the \gls{hcml} process.

Unfortunately, there are no methods in the literature for evaluating every feedback signal provided by the teacher, since different sequences of signals can obtain the right effect on the learned model.
Therefore, the only way to evaluate whether the given feedback is correct is by the testing of the policies.

Below we describe some of the Human Performance Metrics applied in \gls{iil}, like human workload and users' perception of robot model/behaviors. According to the literature, the previous metrics evaluate different aspects of human performance in \gls{hcml} \citep{cui2021understanding,kulesza2014structured,mohseni2019multidisciplinary}; hence, they are presented in two different sections. The aforementioned metrics should be combined with the Robot Learning Performance metrics to better evaluate and improve the outcomes of the \gls{hcml} process. 

\subsubsection{Number of Interactions/Interventions from a Teacher} 
One of the main goals related to \gls{iil} is to minimize human effort during the teaching experience. The amount of feedback provided by the teacher for successful task execution is a quantitative measure of human effort --- the amount of feedback tends to correlate with the effort levels. This metric is used in \gls{iil} to evaluate human effort during the teaching experience \citep{bajcsy2018learning, bootsma2021interactive, celemin2019interactive, cronrath2018bagger, delpreto2020helping, franzese2021ilosa, goecks2019efficiently, hoque2021lazydagger, jain2015learning, laskey2016shiv, le2018hierarchical, myers2021learning, najar2016training, peng2016need, tenorio2010dynamic} 
    
\subsubsection{Interaction Time} 
Interaction time is a metric that allows measuring human effort in \gls{iil} tasks. Commonly, it includes the total study duration for a teacher to successfully train a model \citep{chisari2022correct, franzese2021ilosa, perez2019continuous, losey2022physical, bajcsy2018learning, jain2015learning}. \citet{hoque2021lazydagger} also consider the time for switching between interactive mode and autonomous robot mode.
    
\subsubsection{Teacher's Cognitive Load and Engagement Levels} 
This metric represents the cognitive load levels and engagement levels of the teacher during the learning cycle, which can be measured via the NASA Task Load Index (NASA-TLX), Van der Laan questionnaires \citep{van1997simple} for perceived workload \citep{meszaros2022learning, delpreto2020helping, jauhri2021interactive,hoque2021thriftydagger}, and/or via physiological measures \citep{ferraz2019multisensory, cui2021understanding}. The NASA-TLX is a widely used subjective assessment tool that rates human perceived workload and performance for a certain task. The total workload is divided into six subjective subscales comprising i) Mental Demand, ii) Physical Demand, iii) Temporal Demand, iv) Performance, v) Effort and vi) Frustration. Subjective measures of cognitive load such as the NASA-TLX tend to have a high variance across humans \citep{cui2021understanding}. Additionally, it can be helpful to make use of objective measures, e.g., physiological metrics such as electroencephalographic analysis and physical exertion monitoring, both recently used in human-robot interaction to objectively quantify human cognitive load and physical load. Increases in human workload correlate with a decrease in task performance in general human-robot interaction tasks \citep{ferraz2019multisensory}. 

\subsubsection{User Perception of Robot's Behavior} 
The \emph{user mental model} is measured by asking users their view on the logic behind the model decision-making process, i.e. users' perception of how correctly the robot learns the task. In a robotic task, the user observes the execution of the robot and evaluates its behavior. Evaluation methods include interviews, think-alouds, self-explanations, Likert-scale questionnaires, users' model output prediction, and users' model failure prediction \citep{biyik2020asking, palan2019learning, bajcsy2018learning, losey2022physical}.
\citet{sadigh2017active} evaluate users' perception of whether the robot understands and executes the driving task as desired by the users. \citet{jain2015learning} measure users' perception on validity/correctness of robot motions.

\subsubsection{User Trust and Reliance} 
\emph{User trust and reliance} represents ratings for whether the users trust the robot or not, e.g., to operate safely in a factory, office, or household environment. Evaluation methods include subjective measures, e.g. interviews, self-explanations and/or Likert-scale questionnaires; and objective measures, e.g. user perception of model competence to execute a task \citep{bajcsy2018learning,delpreto2020helping}. \citet{delpreto2020helping} ask users to rate (on a scale from 1-7) how likely they would trust a robot in a variety of settings and tasks in the context of grasping items - whether they would trust the robot to pick up different objects, work with power tools, or operate in a classroom. 
    
\subsubsection{Easiness of Interactions and Usability} 
\emph{Easiness on interactions and usability} represents users' ease and confidence when interacting with a robot in order to complete a task. This metric can be measured by subjectively asking questions to users \citep{ bajcsy2017learning, bajcsy2018learning, biyik2020asking, cui2019uncertainty, franzese2021ilosa, jain2015learning, palan2019learning}. \citet{cui2021understanding} report that several researchers have been using the System Usability Scale to measure the perceived usability of a \gls{ml} system --- this scale evaluates users' ease and confidence when using a system. The System Acceptance Scale can also be used in  \gls{ml} to evaluate the acceptance of new technologies due to its simplicity and reliability \citep{van1997simple}. 

\section{Discussion}
Even though there have been a few improvements in user studies in \gls{iil}, many aspects have not yet been covered in the literature, e.g. objective measures of human performance as well as evaluation of the users' mental model. Nevertheless, they are necessary to perform more accurate evaluations on human response to interactions with artificial agents and to improve overall efficacy and efficiency in the human-agent interaction process. 

A systematization on how to design user studies in \gls{iil} is also necessary to better inform researchers on how to carry them in this field. User studies in \gls{iil} should target clear communication between a human and an artificial agent, in addition to a user-friendly, intuitive, and enjoyable experience for the human supervisor. 

\newpage
\chapter{Benchmarks and Applications}\label{sec:BenchmarksAndApplications}

Evaluating robotic systems is a well-known and difficult challenge due to the wide variety of robots, tasks, and environments tailored for each robotic system \citep{behnke2006robot}.
Such difficulties are exacerbated in the context of \gls{hil} learning, where the performance of learning methods is highly influenced by the data used \citep{bouthillier2021accounting}, which often comes from humans.
Furthermore, the different modalities of interactions (see Section \ref{sec:modes}) influence the quality and amount of information that humans provide as feedback during learning, creating the necessity to compare not only the learner's performance but also the human aspect of an \gls{iil} system (see Section~\ref{sec: user experience}).

The lack of reproducibility of experiments has been debated for years in the field of cognitive robotics.
In order to leverage the opinions of experts on how to evaluate such systems, \citet{aly2017metrics} present the discussion entitled \emph{``Towards Intelligent Social Robots - Current Advances in Cognitive Robotics''} \footnote{Held at the \emph{``15th IEEE-RAS Humanoids Conference, Seoul, South Korea, 2015''}, \url{https://intelligent-robots-ws.ensta-paristech.fr/}}.
They recognize the difficulties in testing and comparing complex robotic systems in a fair manner, as well as a lack of proper methods and tools to do so.

A common way to compare \gls{iil} methods is to evaluate their performance on specific applications, which can be tailored to demonstrate specific aspects and contributions of learning methods or to achieve task completion.
A different approach is to leverage simulation and datasets to achieve some degree of uniformity.
Sections~\ref{sec: applications},~\ref{sec: datasets},~and~\ref{sec: benchmarks} present the main applications, datasets, and applications used for evaluating and comparing \gls{iil} approaches, respectively.

\section{Applications}\label{sec: applications}
During the research and development of \gls{iil} methods, a vast number of applications have been proposed which can potentially benefit from the intrinsic characteristics of this field.
On one hand, such applications can be designed to validate or demonstrate research ideas, showing gaps in previous methods and how they are addressed by the proposed one, or to evaluate the learner's capabilities, such as learning rate or generalization. These applications are called \emph{testbed applications}.
On the other hand, applications can target the consumer or products, whose requirements and preferences in a specific task are the target of evaluation. These applications are called \emph{use-case applications}.

\subsection{Testbed Applications}
We categorize the testbed applications into their task domains (e.g., robot manipulation or robot navigation), identifying representative tasks and successful demonstrations of \gls{iil} methods. 
\subsubsection{Manipulation}
\paragraph{Pick and Place} 
This task refers to picking an item in one location, moving it, and placing the item in another location.  It is a fundamental automation task that is commonly used as a sub-task in other long-horizon manipulation tasks. Because of this, it is commonly found in a variety of robotics applications. 

Human-like picking is a challenging robotic task that requires learning the constraints and adaptations for non-zero-velocity end-effector movements, orientations, and gripper width precisely for a successful pickup; otherwise, the task will probably fail. An \gls{iil} approach is suitable to learn this task efficiently by improving an initial demonstration through interactive corrections. 
\citet{meszaros2022learning} applied an \gls{iil} approach for the task of non-zero velocity picking of objects using a 7 \gls{dof} Panda robot. The users could overcome the limitations they had during the demonstrations and teach the desired behaviors using corrections.

\begin{figure}
    \centering
    \includegraphics[scale=0.5]{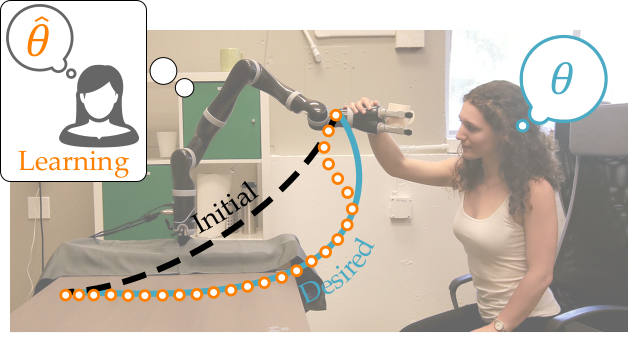}
    \caption{A robot learns from physical human interaction to understand the task objective (i.e., go closer to the table). Source: \citep{losey2022physical}.}
    \label{fig:cup2table}
\end{figure}
Additionally, \citet{bajcsy2017learning,bajcsy2018learning,losey2022physical} develop an online physical human-robot-interaction method to learn successful object placement tasks with user preferences.
Figure~\ref{fig:cup2table} shows the experimental validation by \citet{losey2022physical} where the robot learns object placement with human preferences.

\paragraph{Contact-rich Manipulations} 
Contact-rich manipulation tasks are space-critical tasks, meaning that certain regions of the state space require precise sequences of actions to make meaningful progress. These regions are a bottleneck for successful task execution because a small deviation from the correct policy may lead to failure, e.g., the \emph{peg-in-hole} task. Task failures due to inaccuracies in the agent's actions while traversing critical state-space regions can be avoided using \gls{iil}.

\begin{figure*}[h!]
 \centering
   \subfloat[] { \includegraphics[scale=0.375]{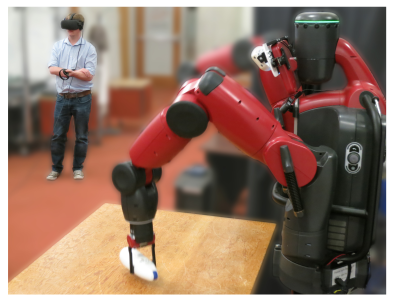} }
   \hfill
   \subfloat[] {  \includegraphics[scale=0.325]{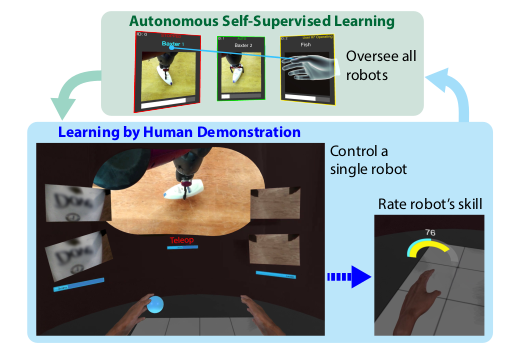}}  
    \caption{Figure (a) shows the setup where a person supervises a robot for a grasping task.
 (b) shows the virtual reality interface where the user remotely gives grasping demonstrations.  Source: \citep{delpreto2020helping}. }
 \label{fig:grasping}
 \end{figure*}

One of the methodologies to traverse such bottlenecks is to learn a policy by requesting feedback from an expert when the agent is not able to find solutions.
\citet{delpreto2020helping} apply an \gls{iil} approach where the policy predicts a vector of confidence scores for four different gripper orientations, and the one with the highest confidence is selected.
The robot autonomously attempts a predicted grasp and detects whether it can lift the object and hold it for a fixed amount of time. 
If the robot repeatedly fails to execute a successful grasp orientation, it requests user assistance, as shown in Fig.~\ref{fig:grasping}.

Instead of agents requesting help, another way of traversing the bottleneck in these tasks is learning based on human-gated interventions. The expert supervises the agent's task execution and provides intervention when considered necessary. \citet{mandlekar2020human} use this approach in two contact-rich manipulation tasks for a 6-DoF robot manipulator: a threading task and a coffee machine task. 
The threading task consists of the robot carefully grasping a rod and inserting it into a hole. The coffee machine task consists of the robot grasping a coffee pod, inserting it into the coffee machine, and closing the lid. These tasks contain critical regions while performing sub-tasks like grasping, inserting, and closing.

The presented methods assume that the expert is always able to correctly intervene when necessary. However, this might not be possible with non-expert users. For such a scenario, \citet{chisari2022correct} combines human interventions with evaluative feedback. The user intervenes to correct undesirable behavior. The evaluative feedback is used to select or discard the part of the trajectory which the user cannot correct. The approach is applied to learn contact manipulation tasks: pushing a box, picking up a cube, and pulling a plug from real-world high-dimensional image observations using a real KUKA robot manipulator. This approach shows that the combined feedback-based strategy is more advantageous than \gls{bc} \citep{osa2018algorithmic} and different interactive learning strategies \citep{kelly2019hg,mandlekar2020human}. 

Finally, \citet {franzese2021ilosa} employ demonstrations and corrections on desired end-effector transitions, to learn policies able to perform well during critical regions traversal. The approach is successfully validated in tasks of plug removal/insertion, pushing box and wiping whiteboard using a 7 \gls{dof} Panda robot manipulator. 
The policies also learn less stiff control behavior in free regions which is desirable for safe manipulations in human-robot environments.

\paragraph{Ball in a Cup}
This is a challenging robotic manipulation task where there is a cup attached to the robot's end-effector and a ball attached to the cup by a thread. The goal is to move the end-effector to swing the ball making it land in the cup.
This task is a challenging manipulation benchmark in robotics, as it corresponds to an underactuated system that depends on hard-to-model physical factors.
This task has been approached using an \gls{il} based \gls{ps} optimization on a real 7 \gls{dof} Barrett WAM robot manipulator \citep{kober2008policy}. 

\begin{figure}
    \centering
    \includegraphics[scale=0.5]{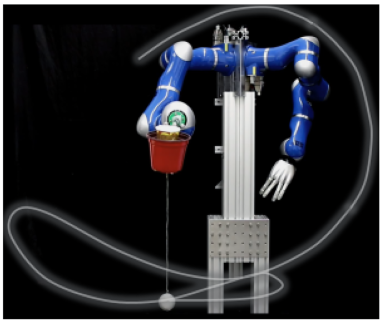}
    \caption{Ball-in-a-cup set up used in \cite{celemin2019reinforcement} composed by a 7\gls{dof} KUKA lightweight arm and an OptiTrack system, which tracks the positions of the ball and the cup.}
    \label{fig:ballCup}
\end{figure}

Later, \citet{celemin2019reinforcement} apply an \gls{iil} method for learning the ball-in-a-cup task by combining human teacher's knowledge during the \gls{ps} exploration process. The teacher provides relative corrections to the robot's end-effector.
Experiments are carried out using a \gls{ps} method on the experimental setup shown in Figure~\ref{fig:ballCup}.
Results show that the human interactions lead to an improvement in convergence speed of the \gls{ps} method by an order of 4 to 8.

\paragraph{Writing Symbols} 
The task of writing symbols consists of generating an accurate end-effector trajectory with a particular shape. Unfortunately, a combination of improper user interfaces and insufficient user expertise may limit obtaining high-performance writing demonstrations. 

\begin{figure*}[htb]
 \centering
   \subfloat[]{ \includegraphics[scale=0.65]{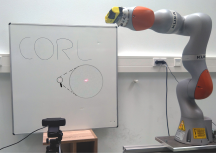} }
   \hfill
   \subfloat[] {  \includegraphics[scale=0.65]{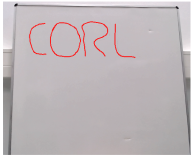}}   \caption{Figure~\ref{fig:drawing} a) shows the robot being taught to move the laser point to draw the characters and (b) the characters drawn by the robot (laser point trajectory tracked by the camera) after about 7 minutes of training per character. Source: \citep{jauhri2021interactive}.}
 \label{fig:drawing}
 \end{figure*}
In absence of good initial demonstrations, \citet{schroecker2016directing} develop an \gls{iil} method that uses soft via-points based demonstrations to initialize a writing policy and interactively refine it through a \gls{ps} process.   
This approach is validated for the task of writing symbols in simulations. The results show accurate and smooth symbol reproduction without having good quality human demonstrations of the task.

\citet{celemin2019reinforcement} propose a mechanism for writing tasks using interactive corrections during robot execution.
This approach is validated for writing symbols using a real 6 \gls{dof} robot arm. The results show that a reduction of $84.4\%$ in symbol reproduction error is achieved using the interactive \gls{ps} method in comparison to a non-interactive one.

\paragraph{Item Sorting} 

\begin{figure}
    \centering
    \includegraphics[scale=0.5]{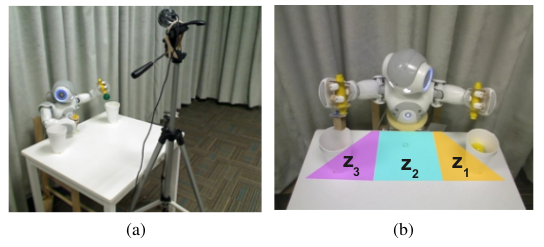}
    \caption{Experimental setup for object sorting using human guidance. Source: \citep{suay2011effect}.}
    \label{fig:sorting}
\end{figure}

The task of item sorting consists of ordering items based on their categories. 
\citet{suay2011effect} apply an \gls{iil} approach for using human evaluative feedback and human guidance in a \gls{rl} setting for an object sorting task. The item sorting robotic task setup, as seen in Figure~\ref{fig:sorting}, consists of an Aldebaran NAO humanoid robot in front of a table with different objects placed in three zones, along with storage bins for each category. 
This study shows that the robot learns to use its camera to identify the characteristics of the objects and pick them up and place them in the appropriate cup.

\begin{figure}
    \centering
    \includegraphics[scale=0.5]{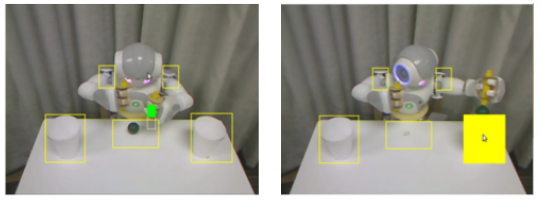}
    \caption{Interface for human guidance. Figure shows (left) how to give a positive reward by left click and mouse drag upwards; (right) shows how a user guides the robot by selecting a region of interest out of five rectangular regions, each associated with an action set. Source: \citep{suay2011effect}.}
    \label{fig:interface-chernova}
\end{figure}

The task of item sorting in a dynamic environment, e.g., on a conveyor belt, requires time-critical movements of the robot to grasp the moving items and sort them as per their category. A task of fruit sorting on a moving conveyor is shown using \gls{iil} by \citet{perez2020interactive}. The setup consists of a conveyor belt transporting oranges and pears, a 3 \gls{dof} robot arm, and a head-mounted camera. The robot uses raw camera images as input and selects oranges with its end-effector and moves away from the pears.

\subsubsection{Navigation}

\paragraph{Autonomous Driving}
Autonomous driving is a challenging robotics problem due to varying environments, terrain, topologies and dynamic vehicle/pedestrian interaction or disturbances. This area of research aims to develop intelligent autonomous vehicles to achieve safe and high-performance mobility \citep{perezdattari2022}. The problem consists of steering the vehicle based on sensory observations (camera, radar, lidar etc.) to weave around the obstacles and navigate on the road. 
\begin{figure}
    \centering
    \includegraphics[scale=0.5]{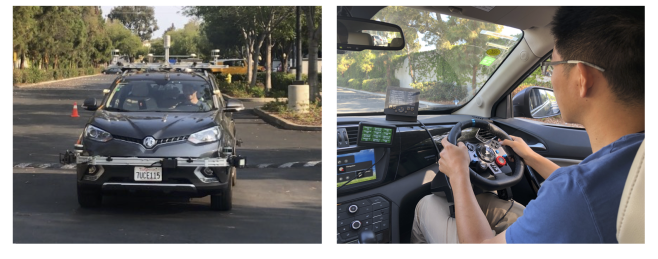}
    \caption{Test vehicle (left) and expert driver interface (right). Source: \citep{kelly2019hg}.}
    \label{fig:driving}
\end{figure}

\citet{kelly2019hg} allow human experts to take control when they deem it necessary, and to maintain exclusive control authority until they manually hand control back to the agent. The approach is experimentally validated in driving a vehicle without collisions along a road with other stationary vehicles as obstacles. The vehicle and testing setup are shown in Figure~\ref{fig:driving}. 

In addition to safe driving performance, it is also important to minimize the expert's burden during the interaction. Hence, a technique for querying the human expert is presented in \citep{cui2019uncertainty}. In this approach, the agent predicts uncertainty to anticipate risky states for the vehicle and switches control to the human expert to prevent dangerous situations. The experimental results from simulated driving tasks in CARLA driving simulation environment \citep{dosovitskiy2017carla} demonstrate that the uncertainty estimation method can be leveraged to reliably predict risky states and minimize human efforts.

\paragraph{Mobile Robots}
To improve the performance and efficient exploration of mobile robot applications, \gls{iil} approaches are explored in the literature. \citet{knox2013training} demonstrate the usefulness of the \gls{tamer} framework for interactive navigational behaviors on a real mobile-dexterous-social robot platform called Nexi. The task consists of navigation to a marker that can be moved by the human trainer. The experiments show that a robot can learn to perform sequential navigation tasks using only real-valued feedback on its behavior from a human trainer. 

\citet{perez2018interactive} show an application for learning to drive a Duckiebot autonomously through the desired track from the project Duckietown \citep{paull2017duckietown}. The robot uses raw visual information from an onboard camera to follow a path without leaving the road. The human teacher observes the task and advice linear and angular velocity corrections to the robot's actions using a keyboard in case of deviations. The robot is able to learn the task of navigation from scratch using only human corrections via the \gls{d-coach} framework in approximately 6 minutes. 

\citet{bootsma2021interactive} show learning navigation behavior for an autonomous mobile robot by leveraging the strengths of different sensors, under human supervision. They demonstrate that the learning method can prevent dangerous navigation behaviors. The experiment was carried out on a real mobile robot ROSBot, equipped with a 2D lidar and a camera. 
\begin{figure*}[h!]
 \centering
   \subfloat[] {\includegraphics[scale=0.5]{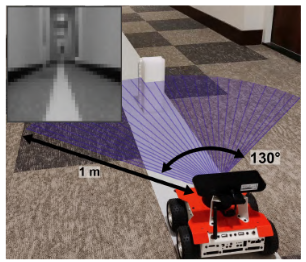}}
   \hfill
   \subfloat[]{  \includegraphics[scale=0.45]{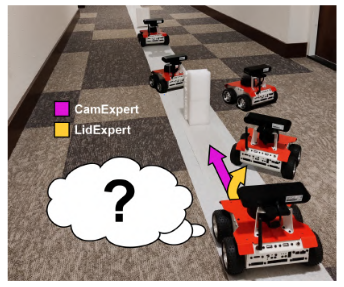}}   \caption{Learning navigation behavior for an autonomous mobile robot. The figure shows conflicting decisions based on two different sensory feedback. In conflicting situations, an expert is queried to take control and the correct action of a sensor-fused policy. Source: \citep{bootsma2021interactive}.}
 \label{fig:mobile-lira}
 \end{figure*}
The experimental setup is shown in Figure~\ref{fig:mobile-lira}.

\paragraph{Drones} Drone navigation consists of successfully maneuvering to reach a target while avoiding obstacles. Drone navigation is a challenging task since it can include non-linear dynamics, blurry images from the moving drone, and the presence of environmental gust disturbances. 

\begin{figure*}[h!]
 \centering
  \subfloat[]
   {
       \includegraphics[width=32ex, height=21ex]{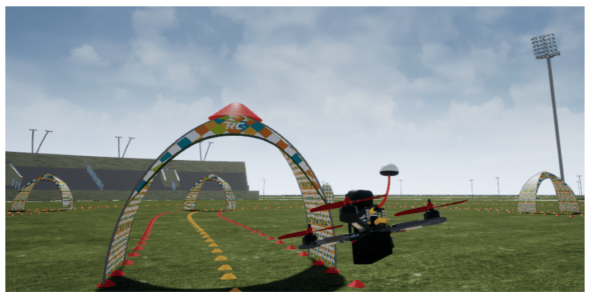}
    } 
     \hfill
   \subfloat[]
   {    
   \includegraphics[scale=0.3]{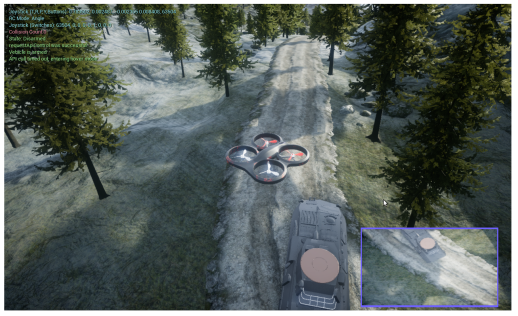}
   }
 \caption{Figure shows (a) Drone Navigation and (b) Drone Perching. Source: \citep{li2018oil},\citep{goecks2019efficiently}.}
 \label{fig:drone}
 \end{figure*}

The approach introduced by \citet{li2018oil} enables learning from multiple non-expert teachers by discarding bad drone maneuvers. 
The capabilities of this approach are demonstrated in drone navigation through racing tracks in the Sim4CV racing environment \citep{muller2018sim4cv} shown in Fig. \ref{fig:drone}(a). 

Moreover, \citet{goecks2019efficiently} demonstrate an aerial robotic perching task using a drone in AirSim simulator \citep{shah2018airsim}. 
The simulated setup in Figure~\ref{fig:drone}(b) shows the drone with a downward-facing camera hovering over the moving landing platform. 

\subsection{Use-case Applications}
Use-case applications consist of applications for a targeted audience. \gls{iil} research can be employed to provide targeted solutions to specific problems by considering domain knowledge.
\subsubsection{Assistive Robots}


\begin{figure*}[h!]
 \centering
   \subfloat[User feeding]
   {    
   \includegraphics[height=15ex, width=22ex]{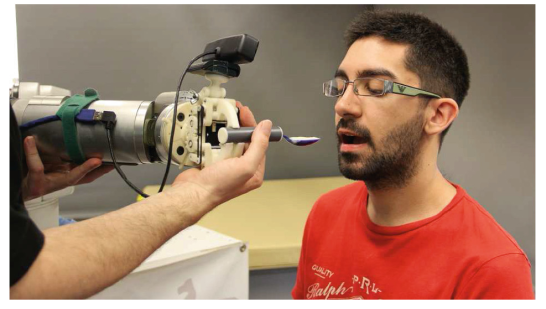}
   }
   \hfill
   \subfloat[Shoe fitting ]
   {
       \includegraphics[height=15ex, width=22ex]{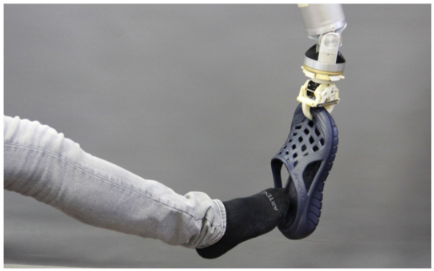}
    }  
     \hfill
   \subfloat[Dressing garment]
   {
       \includegraphics[height=15ex, width=22ex]{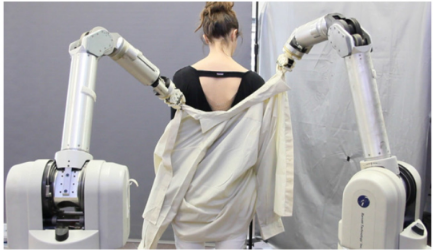}
    }  
 \caption{Assistive personalized application. Source: \citep{Canal_THRI2021,canal2016personalization}.}
 \label{fig:assistive-2016-canal}
\end{figure*}

The Healthcare industry offers opportunities for many important use-cases of an assistive robot e.g., feeding, dressing and shoe fitting for disabled people. 
A possible way to empower such attributes is through \gls{iil}. 
For instance, \citet{Canal_THRI2021,Canal_icra2018,canal2016personalization} develop a robot personalization framework for three different assistive applications i.e., feeding, shoe fitting and dressing, where the robot performs each task in a different manner based on corrective feedback from the user. The experimental setups for different applications are shown in Figure \ref{fig:assistive-2016-canal}.

\subsubsection{Household Robots} 
Robots are expected to serve alongside humans in household environments. In these environments, the successful execution of the task depends not only on accurate robot motions, but also on safety measures (not damaging fragile items or not hurting nearby humans) and on a clear understanding of human preferences.  
Although complex behaviors like robot motions can be taught using human demonstrations, safe behavior and human preferences are difficult to anticipate and model.
\begin{figure*}[h!]
 \centering
   \subfloat[Ranking-based preference feedback: (Left) Robot ranking of trajectories and (Middle) displays top three trajectories on a touch screen device (iPad here). (Right) User selects the trajectory as per his preference.]{\includegraphics[scale=0.25]{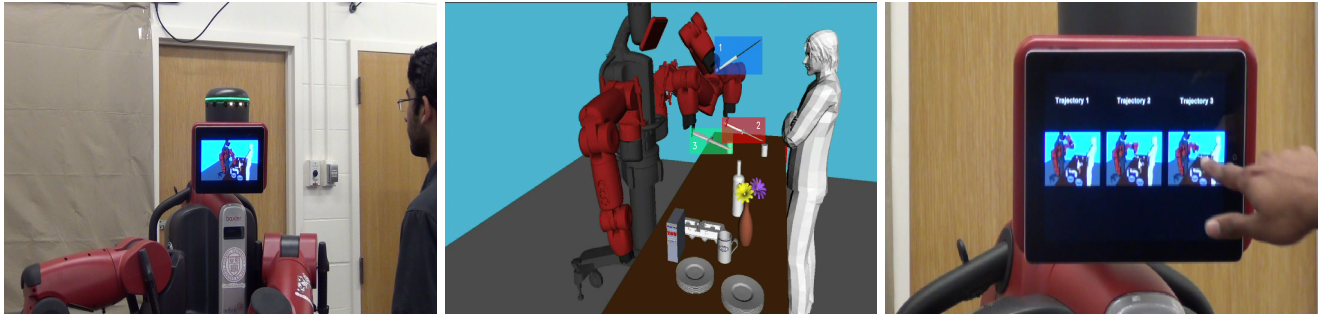} }
   \hfill
   \subfloat[Demonstration-based preference feedback. (Left) robot displays the planned trajectory (waypoints 1-2-4) and human corrects waypoint 2 because it is very near to flower (Right) demonstration on waypoint 2] {\includegraphics[scale=0.25]{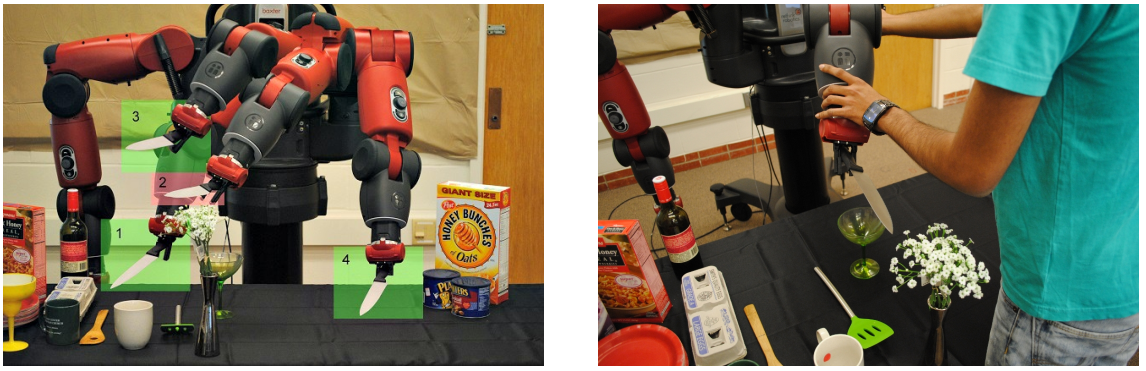}}  
 \caption{Learning user preferences for household manipulation tasks. Source: \citep{jain2015learning}.}
 \label{fig:house}
 \end{figure*}
 
\citet{jain2015learning} demonstrate a case study using PR2 and Baxter robots working in human-centered environments for tasks like household chores and grocery checkouts, as shown in Figure~\ref{fig:house}. They propose an \gls{iil} framework to incorporate user feedback and improve its result as per user preferences.
The approach is experimentally validated on 35 robotic tasks in household settings. 

\begin{figure}[h!]
    \centering
    \includegraphics[scale=0.3]{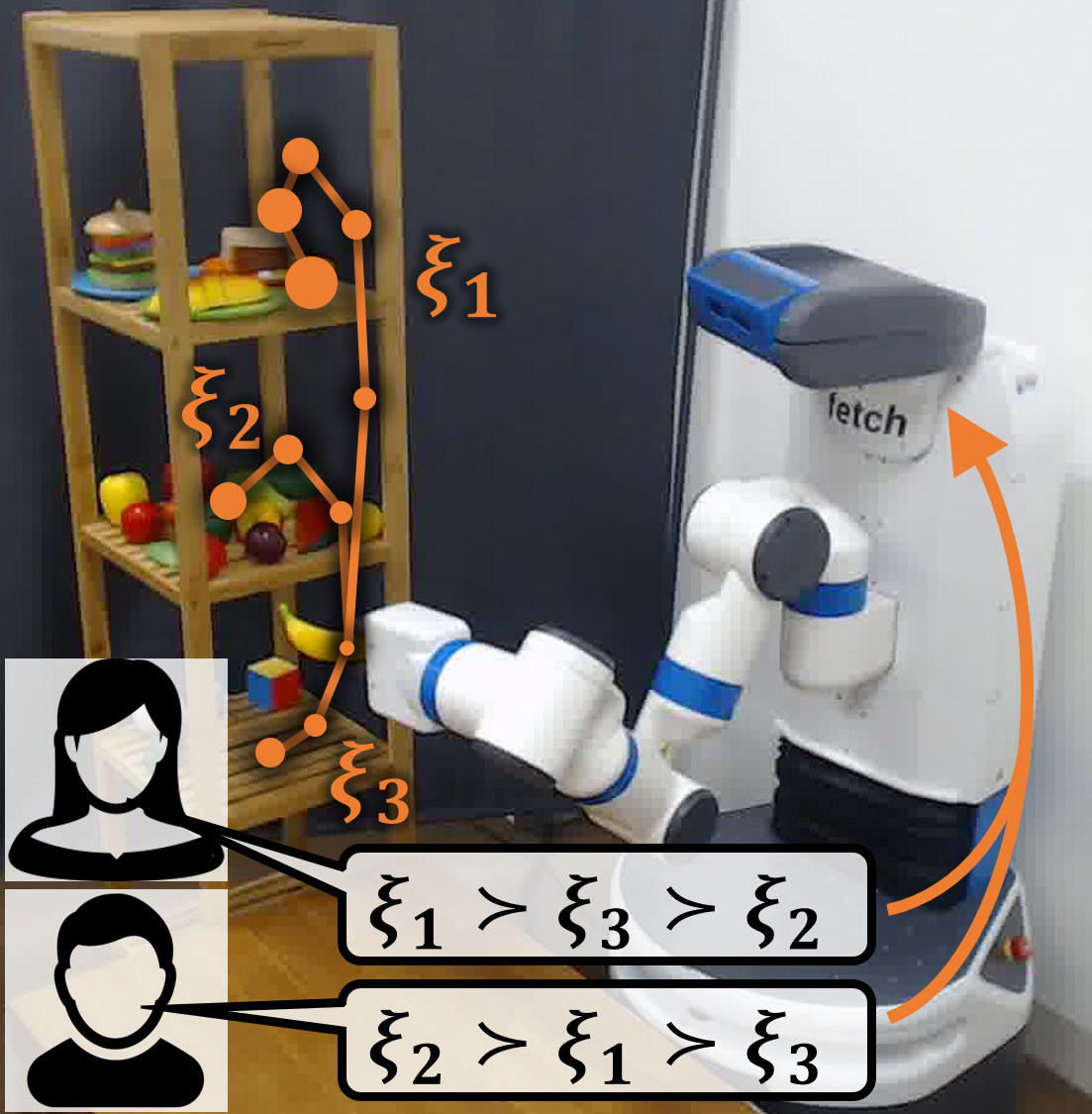}
    \caption{ Figure shows the robot taking a decision based on multiple users' preferences to shelve an item. Source: \citep{myers2021learning}.}
    \label{fig:shelving}
\end{figure}

While working in household environments, a robot has to encounter multiple user preferences for the same task. 
Learning a unimodal reward from data with inconsistent preferences, coming from multiple users, is likely to result in a low-quality policy. To address successful task execution in such situations, \citet{myers2021learning} validate an active query-based \gls{iil} method for learning multimodal reward functions from multiple human preferences. The approach is evaluated with a Fetch robot on the task of learning to shelve an item using multiple users' feedback (Figure~\ref{fig:shelving}). 

\subsubsection{Medical Robots} 

\begin{figure*}[h!]
 \centering
   \subfloat[Suture Tying on a Surgical Da Vinci Research Kit. Source: \citep{laskey2016shiv}]
   {    
   \includegraphics[height=22ex, width=32ex]{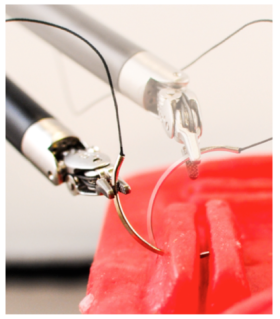}
   }
   \hfill
   \subfloat[Prosthesis Limb Controller for AX-12 Smart Arm. Source: \citep{pilarski2011online} ]
   {
       \includegraphics[height=22ex, width=32ex]{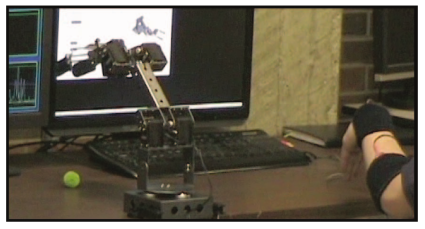}
    }  
 \caption{Assistive robots in medical applications. }
 \label{fig:medical}
\end{figure*}

\paragraph{Surgical Needle Insertion}  
In a medical surgery, a robotic assistant is expected to help the surgeons, for instance, in suture tying. Such tasks require an accurate and precise insertion of the surgical needle, where inaccuracies could lead to an undesirable outcome, such as failed suture or wounded neighboring tissues. 
For such applications, \citet{laskey2016shiv, hoque2021thriftydagger} validate \gls{iil} on a Surgical Da Vinci Research Kit Figure~\ref{fig:medical}(a). 

\paragraph{Myoelectric Prostheses} 
Multi-functional myoelectric prostheses are used to control the movement of robotic appendages. To do so, they monitor electrical signals from muscle tissues, which are produced from the deficient limb.
In these applications, the patient's intent and usage pattern are very important to consider. Therefore, clinical and technical intervention is always required for a patient to improve the device's performance. 
Therefore, to develop artificial intelligent limbs for patient customization, \citet{pilarski2011online} develop successful limb controllers which are initialized using prostheses data and improved using human-delivered signals. The feedback-based learning framework is validated using AX-12 Smart Arm (see, Figure~\ref{fig:medical}(b)) which shows that method can be adapted to varied application settings and patient needs.

\subsubsection{Industrial Robots}
\begin{figure*}[h!]
 \centering
   \subfloat[NIST board insertion]
   {    
   \includegraphics[height=15ex, width=22ex]{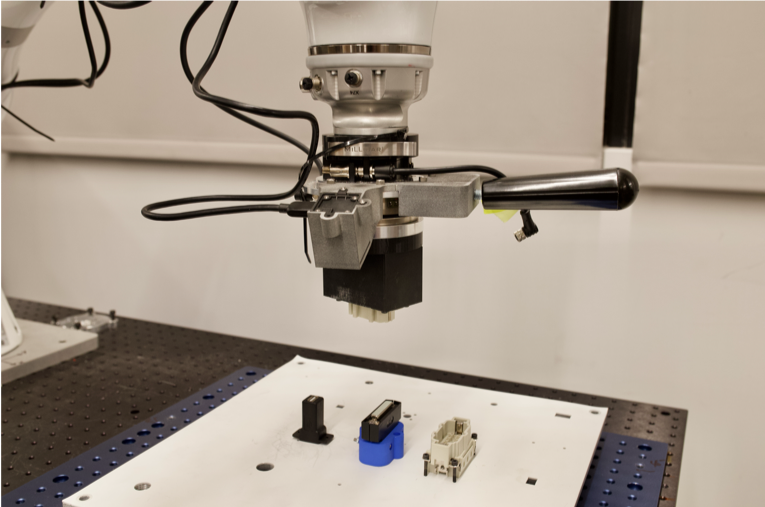}
   }
   \hfill
   \subfloat[Moving HDMI Insertion]
   {
       \includegraphics[height=15ex, width=22ex]{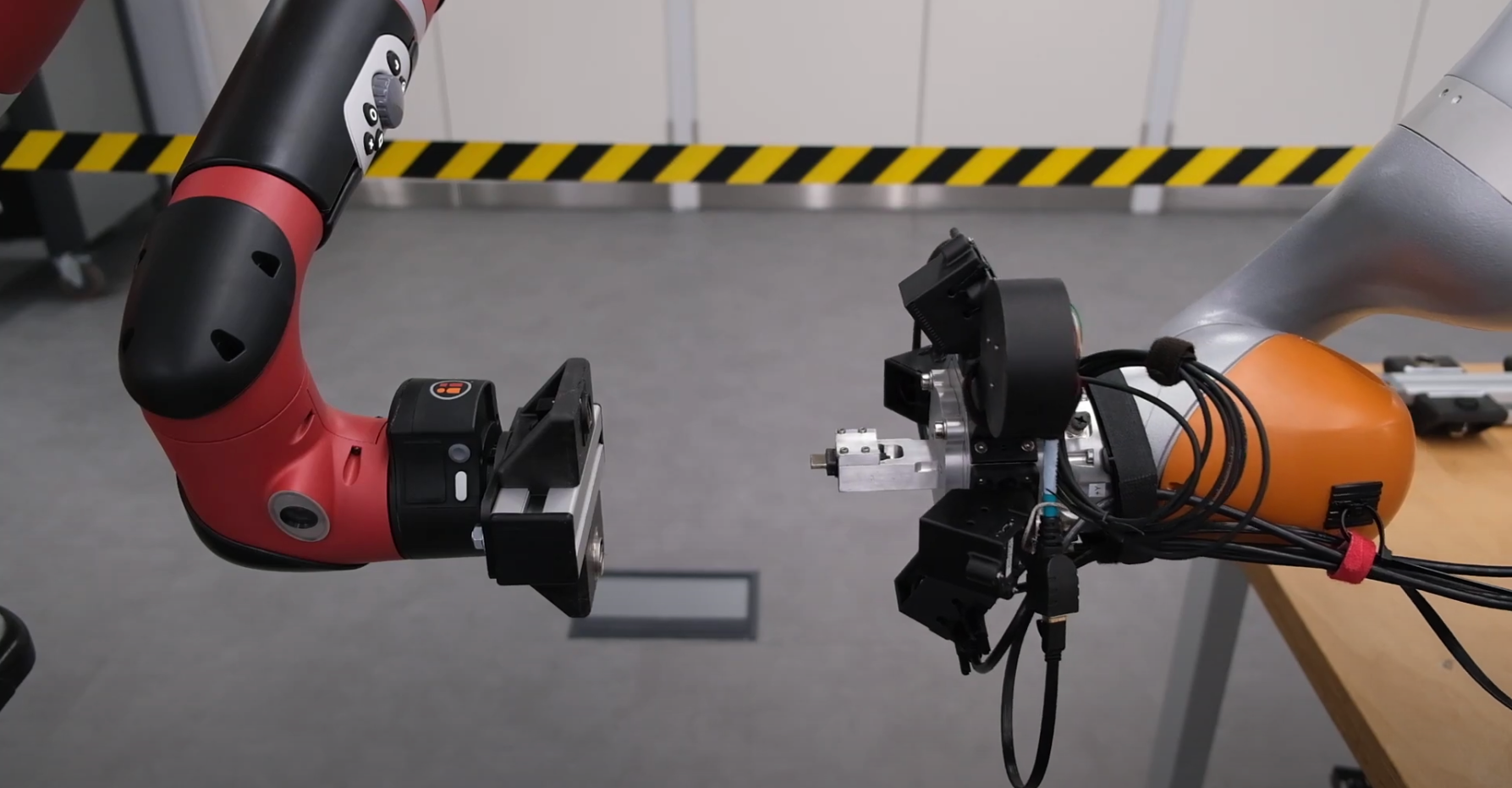}
    }  
     \hfill
   \subfloat[Key-lock insertion]
   {
       \includegraphics[height=15ex, width=22ex]{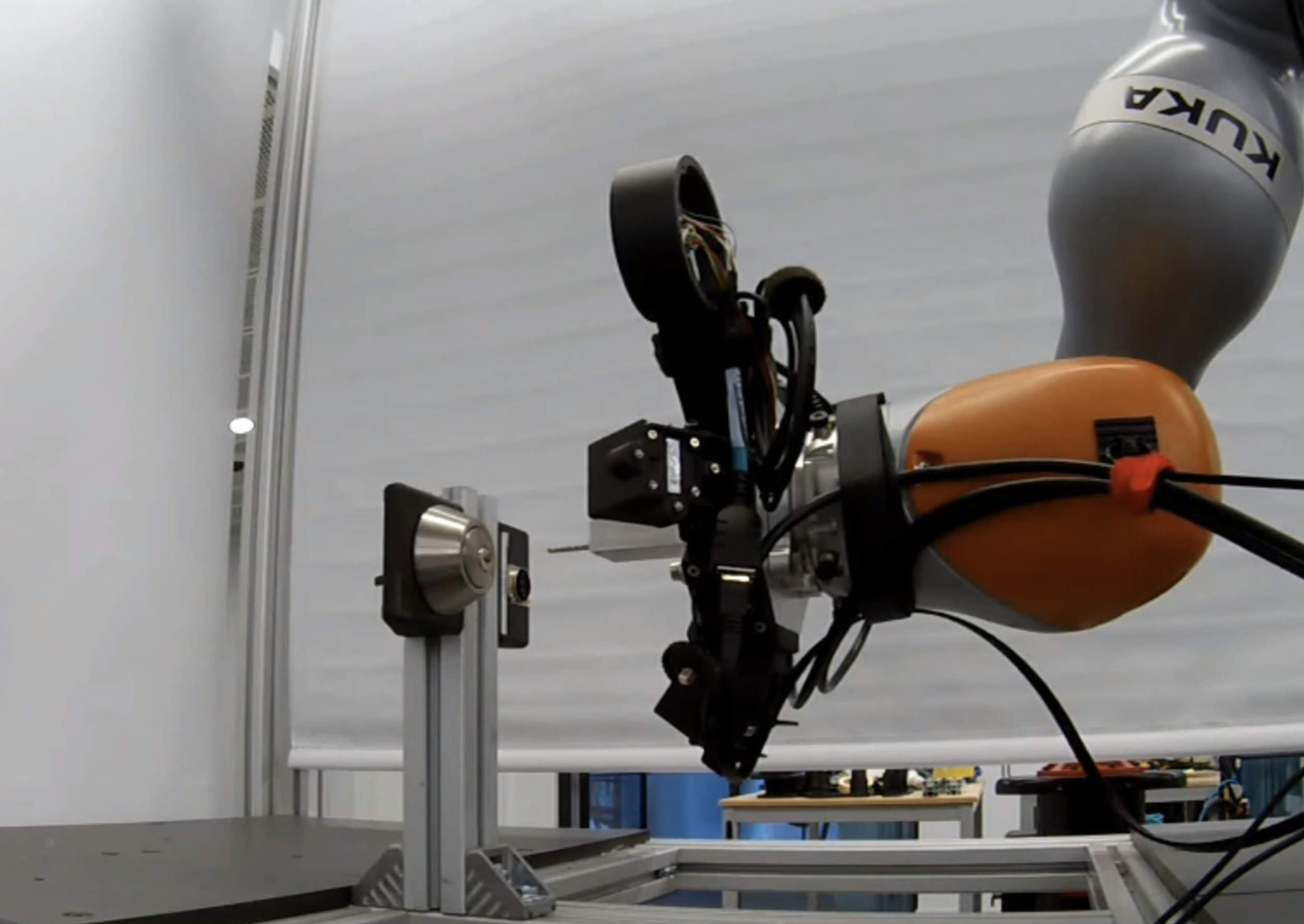}
    }  
 \caption{Robot in Assembly Tasks. Source: \citep{luo2021robust}.}
 \label{fig:assembly-tasks}
\end{figure*}

Industrial applications are composed of precision tasks in constrained environments, which are commonly addressed using handcrafted solutions. Hence, as soon as the object of manipulation and the surroundings vary, a significant amount of time and expertise is needed to adapt the solutions.
\gls{iil} approaches are very useful in such settings, as they allow to learn the task from a few examples and generalize to similar tasks.
\citet{luo2021robust} introduce a framework for learning robust robotic manipulation policies in industrial settings by leveraging demonstrations and human corrections.
In this approach, human demonstrations of the industrial tasks are collected using teleoperation. An initial policy is learned from this data and corrected during run-time by a human teacher. The method is evaluated on three challenging tasks of insertion in static and dynamic settings as shown in Figure~\ref{fig:assembly-tasks}. 

\section{Datasets}\label{sec: datasets}

Datasets are an important resource for developing and testing computational methods. They aim to provide data with the same characteristics that are expected in real environments, but that can be processed offline.
However, it is challenging to use datasets for directly testing \gls{iil} methods, given that, in many cases, they can only be provided by an online source, the human teacher.
For instance, it is common practice to use datasets for policy initialization during experiments, enabling a warm-start which can lead to better policies or shorter training sessions, which is interesting for the human teacher.
However, using datasets for policy initialization is limited for setups and tasks similar or equal to the one used to collect the data, preventing their broad adoption in \gls{iil}. 

\paragraph{RoboTurk}
RoboTurk \citep{mandlekar2018roboturk,mandlekar2019scaling} presents a large dataset of manipulation tasks with robotic arms, whose novelty lies within its crowd-sourced model, i.e., demonstrations from many people were collected and saved in the database.
During the demonstrations, the robots are controlled using a cellphone, which is a widely available interface and it does not compromise the learning performance. The user receives a live stream from the camera of the robot set-up (or its simulation).
Current developments on RoboTurk aim to enable humans to perform corrective interventions also using the mobile phone and video stream interface \citep{mandlekar2020human}, effectively enabling \gls{hil} learning.

\paragraph{RoboNet}
Similar efforts in collecting a large number of demonstrations are performed by \citet{dasari2019robonet}, who presents a collection of demonstrations of manipulation tasks using a diversity of robot arms, multiple view-point cameras, and objects.
The dataset is proposed to be used to learn an initial policy, which can be refined by interactive approaches, leading to improved learned performance.
Furthermore, the variance in the data collection is considered a feature, given the argument that learning methods should be able to generalize across different setups, reducing the limitation on how similar the tasks and applications have to be for benefiting from dataset-based policy initialization.

\section{Benchmarks}\label{sec: benchmarks}

Traditional benchmarks aim to provide an accurate performance metric that allows comparisons \citep{fleming1986not} in a standardized manner \citep{standard2008spec}.
Benchmarks are often constructed by carefully selecting standard data and metric measures for a specific set of applications; otherwise, variations in the results might render analysis insignificant \citep{curnow1976synthetic}.

The necessity of standardization makes sense in traditional computer systems. However, not much work has been done towards the creation of benchmarks for \gls{iil} specifically, since the human teacher is the main source of the data, and its modeling remains challenging. Therefore, given the lack of options, several \gls{iil} papers have evaluated their methods using \gls{rl} benchmarks. Although useful to evaluate important aspects of the learning methods, they are not able to provide reproducibility when it comes to evaluating the human factor of \gls{iil}.
Consequently, in the following of this section, several benchmarks that were not necessarily designed for \gls{iil} are analyzed with the purpose of providing ideas, or starting points, for future work.

\paragraph{OpenAI Gym}
One of the most popular benchmarks for testing \gls{iil} approaches is the OpenAI Gym \citep{brockman2016openai}, which contains Atari games, and simulator-based tasks for classical control and for robotics, simulated using the MuJoCo physics engine \citep{todorov2012mujoco}.
Examples of works tested on OepnAI Gym are \cite{akrour2012april,hoque2021lazydagger,palan2019learning,menda2019ensembledagger,myers2021learning}.

\paragraph{Surreal}
Surreal \citep{fan2018surreal} is a simulated framework for \gls{rl} methods which includes a benchmark for robot arm manipulation tasks.
It provides RGB camera, depth map, and proprioception to be used by the learning methods. Furthermore, teleoperation is supported with $3D$ motion capture devices (\gls{vr} controller), which can be used to provide demonstrations or feedback.
The evaluation is performed through simulation of manual and bi-manual setups of Saywer robots using the MuJoCo \citep{todorov2012mujoco} simulation environment.
Moreover, surreal is used by \citet{lee2020follow}, who perform a state selection step for learning object manipulation tasks from human demonstrations.

\paragraph{Habitat}
Habitat \citep{savva2019habitat} is a simulation-based platform for training embodied agents (virtual robots), which organizes datasets, simulators, and tasks in different layers, aiming to enable transfer learning among different robotic agents.
Habitat 2.0 \citep{szot2021habitat} expands the previous work with \emph{home assistants}, including a $3D$ dataset of apartments with articulated objects for navigation and high-level robot manipulation tasks (e.g., tidy the house, stock groceries, set the table).
The environment and evaluation are all based on Bullet physics simulation \citep{coumans2015bullet}, enabling to perform simulations $2$ orders of magnitude faster than other similar simulations.  
Towards benchmarking, the learning methods are provided with RGB, depth, GPS, and Compass inputs and are evaluated in high-level tasks such as \emph{tidy the house}, \emph{Prepare groceries}, and \emph{Set the table}.
The habitat platform is used by \citet{lin2022efficient}, where graph \glspl{nn} is used to learn manipulation tasks in a \gls{lfd} setup.

\paragraph{Meta-World}
Meta-World \citep{yu2019meta} is a simulation-based benchmark focused on meta-reinforcement learning and multi-task learning.
It provides a broader distribution of the tasks in contrast to what is previously used by the meta-reinforcement learning community. 
It emphasizes generalization to distinctly new tasks, beyond plane parametric variation in a single (or a limited number of) tasks by providing $50$ distinct robotic manipulation tasks (e.g., pick-place, reach, push, open window, open cabinet).
Although all provided tasks are short-horizon (not a long-horizon sequence of tasks) it provides a valuable benchmark and is also used in the robot learning community \citep{sinha2022s4rl}. 
Meta-World tasks are implemented in the MuJoCo physics engine \citep{todorov2012mujoco} and the framework provides an OpenAI Gym environment interface \citep{brockman2016openai}.
Besides benchmarks, several baselines for meta- and multi-task \gls{rl} algorithms are available for comparison.
The provided analysis showed that most of the state-of-the-art approaches for \gls{rl} struggle to learn more than a few tasks at the same time highlighting the difficulty of multi-task learning. 

\paragraph{Robosuite}
Robosuite \citep{zhu2020robosuite} is a simulation-based benchmark based on the MuJoCo physics engine but focused on robot learning.
Its structure is modular, allowing for combining different robotic arms (e.g., Panda, two Sawyer arms), grippers, controllers, and sensors.
It includes complex tasks such as bi-manual peg-in-hole, table wiping, and nut assembly; and it also counts with the addition of community-contributed tasks.
Robosuite is used for imitation learning by \citet{mandlekar2018roboturk}, and in offline \gls{rl} by \citet{sinha2022s4rl}.

\paragraph{RLBench}
RLBench \citep{james2020rlbench} is a learning benchmark for robot-arm manipulation tasks, featuring $100$ hand-crafted tasks.
It provides proprioceptive, RGB, and depth images for the learners.
A CoppeliaSim/V-REP \citep{rohmer2013vrep} simulation environment is used for evaluating learners with a simulated Franka Emika's Panda robot arm.
RLBench provides unique scalability through the use of motion planners to create an arbitrarily large number of synthetic demonstrations, a key enabling characteristic for \gls{rl} and \gls{il} methods.
RLBench is used to test and evaluate the interactive learning method by \citet{chisari2022correct}, who combine both corrective and evaluative feedback from the human teacher to asynchronously train a stochastic policy.

\paragraph{NIST}
Aiming to standardize the application which is used to test real-world robotic systems \citet{kimble2020benchmarking} propose the \gls{nist} benchmark, which is composed of a set of assembly tasks and their instructions.
Instructions are provided to cheaply fabricate the parts of the robotic system, which are then used to reproduce the tasks.
This benchmark has been used by \cite{luo2021robust}, who propose an \gls{iil} method focused on industrial setups.

\paragraph{Conclusion}
The adoption of benchmarks for \gls{iil} has been limited, which is expected to be a direct consequence of the limitations in standardization due to the \emph{human-in-the-loop} factor.
Therefore, the adoption of the benchmarks presented in this section might require the design and implementation of human-centered evaluation metrics (see Section \ref{sec: user experience}) in order pro provide a fair comparison between different methods.
This section briefly covered benchmarks with promising applicability within the \gls{hil} scope for robotics. The interested reader is pointed to \cite{stapelberg2020survey} for a survey on benchmarks for reinforcement learning, and to \cite{zhang2018survey} for a survey on benchmarks for deep learning.

\section{Discussion}
In this Chapter, we highlight the main benchmarks and applications that have been approached through the lenses of \gls{iil}.
The literature shows that \gls{iil} has a wide range of applicability, spanning from household, to medical, to industrial robots. 
We observe that many real-world applications require complex skills that are not easy to demonstrate, either due to the unavailability of a suitable interface or lack of demonstrator expertise. \gls{iil} methods show that they can provide a suitable framework to tackle these problems. 
However, it is still an open challenge to design methods that can handle imperfect feedback without issues. 

Various benchmarks exist that can be used in order to evaluate and compare such algorithms in controlled settings. Nevertheless, due to the presence of the human teacher in the learning loop, exact and reproducible results are difficult to achieve, and still represent an open challenge.
Most benchmarks consist of simulated tasks, as they are the most portable and fastest mean to test new algorithms, ranging from arcade games to realistic robotic simulators. An exception is the \gls{nist} benchmark, which provides a standardized set of easily reproducible table-top tasks to be used for evaluating real-world robots.

\newpage
\chapter{Research Challenges and Opportunities}\label{serc:research challenges}
After discussing the different aspects involved in \gls{iil}, regarding \gls{ml} algorithmic features, ways of interaction, interfaces, human factors, and evaluation considerations, we discuss some of the problems in the domain that represent a challenge and are potentially interesting directions for further research that could make the use of \gls{iil} for transferring the human knowledge to the machines more effectively.
\begin{itemize}
    \item \textit{Successfully combining multiple feedback modalities.} In Chapter \ref{sec:modes}, we presented the different modalities the teachers can use to convey their knowledge to the learning system and discussed the benefits and limitations of each of them, along with possible ways to choose which one is more convenient depending on the available teacher and the characteristics of the problem.
    However, there are situations wherein applying different kinds of feedback can be beneficial, e.g., using absolute corrective demonstrations when the user knows exactly what should be done, relative corrections for tuning the performed actions, and evaluative feedback for those states wherein the teacher is not sure what the right action is, but can still judge whether the agent is doing well or not.
    Providing those kinds of feedback seamlessly, with no specified schedule within the same roll-out of the learning policy, is difficult to do with current methods, and almost all state-of-the-art methods focus on exploiting one kind of feedback.
    Furthermore, combining different datasets of different kinds of interactions, and combining smoothly and without conflicts different update rules depending on the information extracted from each kind of modality, remains an open challenge that has still a long path to follow.

    \item  \textit{Dealing with inconsistencies in the feedback.} Unlike other \gls{ml} approaches, learning with humans in the loop has the specific problem of obtaining noisy data that is not only produced by observation and process noise, but also by the mistakes of the human teachers, which can be conflicting with data obtained in a different moment.
    Detecting feedback mistakes is a difficult problem that impacts safety and learning efficiency.
    For many methods, these mistakes condition the final policy performance.
    Some works have shown that different interaction modalities along with the learning schemes could be more robust to mistakes. Moreover, learning from \gls{mdp} reward functions (\gls{rl}) and human feedback have shown less sensitivity to teachers' mistakes.
    This problem, intrinsic to the human teachers, has been neglected by most state-of-the-art methods, which still work under the assumption of having perfect teachers, and have been evaluated with unrealistic perfect oracles.
    Efforts in this direction are required to obtain reliable learning agents for real situations and users.
    
    \item  \textit{Dealing with inconsistencies in the teaching strategy.} In \gls{iil}, the incremental learning process comes with the advantage that, through iterations, the teachers learn more about the problem at hand and the strategies to solve it.
    New knowledge from the teacher can result in feedback that is different with respect to the past for a specific situation. For instance, an action that used to be rewarded can later stop fitting the current policy; therefore, the teacher may consider punishing it, creating an inconsistency that is not produced by an occasional mistake.
    These kinds of situations are very likely to happen with real teachers. Nevertheless, it is challenging to evaluate learning methods regarding this kind of inconsistency, because it is difficult to replicate a change of strategies with human teachers.
    Therefore, this issue represents an algorithmic challenge for detecting and solving the inconsistencies, but also a challenge from the evaluation procedure perspective.
    Some methods, indirectly deal with these inconsistencies, using a limited dataset of the most recent interactions that allow forgetting old data. However, this approach has the disadvantage of losing valuable and currently valid knowledge provided at the early stages of the learning process.
    
    \item  \textit{Dealing with inconsistencies from multimodal behaviors.} In some applications, like collaborative robots, the human operators could adapt the robot to their personal preferences.
    Then, the learning process collects data from different users who work with the robot but has to adapt to each of the users according to their preferences, while leveraging the knowledge contained in the data of the other users that is not conflicting with the current one's preference.
    The previous example has to do with problems that have multiple valid solutions, that can be demonstrated in different circumstances (even by the same teacher).
    In these cases, considerations should be taken in the model approximator and learning process, in order to capture the multimodality of the solution space.
    A more complex challenge results from being able to consider all the mentioned inconsistencies (in the feedback, the teaching strategy, and the multimodal solutions), and tackling each inconsistent data input with the correct strategy.
    
    \item  \textit{Safety during learning and policy deployment.} As in any \gls{hci} or \gls{hri} system, in \gls{iil} human safety is the main priority every time the system is used.
    The safety of the robot and the surrounding environment is a relevant consideration as well.
    In applications wherein the teaching process involves physical interactions, it is important to develop auxiliary models that support the safety of the system according to factors such as the current performance of the policy, the uncertainty of the policy, task-specific measures and interface restrictions. 
    
    \item  \textit{Data efficiency.} This is a general problem in \gls{ml}, especially when working with \glspl{nn}, since training models based on \gls{nn}s often requires a large amount of data, in particular for high-dimensional data.
    We have observed in the literature that, in general, \gls{iil} methods tend to be more data efficient than other strategies such as conventional \gls{il} or \gls{rl}.
    However, with human users, this problem is more critical, since, to avoid demanding an unfeasible or unpleasant high user workload, the amount of data that can be obtained is limited.
    This is a general problem of \gls{iil}, and it could be indirectly approached when tackling some of the other problems listed in this chapter, or when employing auxiliary costs for training, pre-processing modules, pre-trained models, etc.
    
    \item  \textit{Teaching in high-dimensional spaces.} For human beings, handling many variables at the same time is a very difficult task in general \citep{halford2005many}.
    In tasks wherein the agent has many degrees of freedom or many objects and conditions of the environment to consider, teachers could have a hard time processing what the exact actions or transitions to execute are. 
    Therefore, learning strategies that help to reduce this complexity are desirable.
    Using evaluative feedback methods is a feasible solution in terms of the capabilities of the teachers because the feedback is reduced to one dimension.
    However, evaluative feedback does not guarantee a reduction of the solution space, and the required training time for a teacher to obtain a high-performance policy can be excessively long.
    
    \item  \textit{Realistic simulated teacher.} Evaluating and comparing \gls{iil} methods in realistic scenarios involves several human teachers participating in exhaustive experiments.
    In general, performing experiments with multiple participants interacting with the agents, using different learning methods, and repeating many learning processes, is unfeasible. 
    In order to reduce the load on the participants, partial/preliminary evaluations can be based on experiments with simulated teachers that can be complemented with user studies.
    However, simulating all the human factors and the behaviors different kinds of people have in specific situations requires complex human models that have not been studied in depth.
    Additionally, standardizing such realistic oracles would help to extrapolate the results and insights of previous works.
    Obtaining such results from experiments with perfect and unrealistic oracles bypasses the most difficult and challenging problem of \gls{iil}, which is learning from such a complex teacher.
    
    \item  \textit{Unified subjective measures.} \gls{iil} has been developed mostly by \gls{ml} researchers; however, the domain also falls in the category of \gls{hci} or \gls{hri}.
    Research in \gls{iil} requires standard practices applied by the \gls{hci} or \gls{hri} communities; nevertheless, these considerations are not fully adopted in \gls{iil}.
    Although more and more works are including the analysis of some human factors with user studies, there is still a lack of standard protocols that agree on which the most convenient questionnaires, subjective metrics, and experimental setups are.
    
    \item  \textit{Unified benchmarks.} It is very common in \gls{iil} papers, that the evaluation environment is very specific to each work. There is a lack of common benchmarks to use in \gls{iil} research, as well as agreement on relevant problems for evaluating new methods.
    In the \gls{rl} community, there is more progress in this direction, with a number of different simulated environments available as common test-beds.
    Those resources are useful for the \gls{iil} research community; however, they do not completely fulfill all the needs of this specific field of study, and more development is required in this direction.

\end{itemize}

\newpage
\chapter{Conclusion}\label{sec:conclusions}
In this paper, we survey the most relevant works in the \gls{iil} literature, which have been developed for teaching robots or have potential benefits in robotic applications. 
In recent years, many works have shown the potential of these methods for enabling end-users with non-technical backgrounds to program or adapt the operation of robotic systems.
The incremental component of these learning strategies has a positive impact on the usability of the methods and on the performance level of the obtained policies, with respect to traditional \gls{il}.
The paper provides a structure in the field that facilitates the comprehension of the concepts and problems involved.
This organization can help to speed up the learning curve of the new researchers, but also improve the understanding and perspectives of established \gls{iil} practitioners.

Initially, we mention the problems regarding the ambiguous definitions in the terminology used by different authors and propose one unified terminology, along with the resulting classification of learning schemes that allows specifying what methods that learn from teachers can be considered \gls{iil}.
We group the algorithms according to different important aspects such as the information provided by the teacher during the interaction, the information extracted and modeled during the interaction by the learning agent, the way data obtained is handled for learning, the interfaces used for communicating with teachers and learners, the relation \gls{iil} can have with \gls{rl}, the model representations used for abstracting the obtained knowledge, the considerations that should be taken when having humans in the loop and the ways to evaluate their experience, and few other considerations.
Later on, we present most of the benchmarks that can be used in the experimental design of \gls{iil} methods and the most relevant applications that have been tackled with this kind of strategy.
All these aspects have to be considered when selecting, designing, implementing, or testing a method. 
Finally, we discuss some of the challenges that researchers in this field of study still need to address in order to directly or indirectly improve the learning performance of the agent and the teachers' experience, which can be regarded as the general objectives that the research community aims to achieve.

\newpage
\chapter*{Author Contributions}

\noindent \textbf{Carlos Celemin} designed the structure of the paper, worked on Sections \ref{sec:IntroMotivation}, \ref{sec:IntroTerminology}, \ref{subsub:HumanReinforcements}, \ref{subsec:learning from human preference}, \ref{subsec:AbsoluteCorrections}, \ref{subsub:RelativeCorrections}, \ref{sec:ModesDiscussion}, \ref{sub:on/off-lineLearning}, \ref{sub:DiscussionOnOffpolicy}, Chapters \ref{serc:research challenges}, \ref{sec:conclusions}, sketching the figures, the internal review process, and the final rewrite of the document.

\noindent \textbf{Rodrigo Pérez-Dattari} worked on Sections \ref{sec:background-iil}, \ref{sub:directpolicylearning}, \ref{sub:modelslearned-discussion}, \ref{sub:neuralnetworks}, \ref{sub:on/off-policy-learning-sub}, \ref{sub:onoffpolicy-imitationlearning}, the data management, the internal review process, and the final rewrite of the document.

\noindent \textbf{Eugenio Chisari} worked on Sections \ref{subsub:HumanReinforcements}, \ref{subsec:learning from human preference}, \ref{subsec:AbsoluteCorrections}, \ref{sec:learning_reward}, \ref{sec:transition_models}, \ref{sec:feedback_interpretation}, the internal review process, and the final rewrite of the document.

\noindent \textbf{Giovanni Franzese} worked on Sections \ref{sec:other_surveys}, \ref{sec:LfO}, \ref{sec:learning_reward}, \ref{sec:transition_models}, \ref{sec:confidence}, Chapter \ref{sec:model representations}, sketching the figures, and the internal review process. 

\noindent \textbf{Leandro de Souza Rosa}
worked on Sections \ref{sec: task features}, \ref{sec: datasets}, \ref{sec: benchmarks}, Chapter \ref{section:interfaces}, the data management, and the internal review process.

\noindent \textbf{Ravi Prakash} worked on Sections \ref{sec:objectaffordance}, \ref{sec: applications}, Chapter \ref{sec: user experience}, and the internal review process.

\noindent \textbf{Zlatan Ajanović} worked on Sections \ref{sec:background-dmp}, \ref{sec:objectaffordance}, \ref{sec: benchmarks},
Chapter \ref{sec:reinforcement learning with HIL}, sketching the figures, and the internal review process.

\noindent \textbf{Marta Ferraz} worked on Section \ref{sec:InterfaceDesign}, Chapter \ref{sec: user experience}, and the internal review process.

\noindent \textbf{Abhinav Valada} supervised the development of this project and contributed to the internal review process.

\noindent \textbf{Jens Kober} supervised the development of this project and contributed to the internal review process.
\addcontentsline{toc}{chapter}{\protect{}Author Contributions}


\section*{Acknowledgments}
This research has been funded by the Netherlands Organization for Scientific Research (NWO) project FlexCRAFT, grant number P17-01, by the ERC Stg TERI, project reference \#804907, as well as by the BrainLinks-BrainTools center of the University of Freiburg.


\backmatter
\printglossary[type=\acronymtype]

\printbibliography

\end{document}